\documentclass[final,5p,times,twocolumn]{elsarticle}

\newcommand{\citeauthorandyear}[2][]{\citeauthor{#2} (\citeyear[#1]{#2})}
\usepackage{subcaption}
\usepackage{amsfonts,mathtools}
\usepackage{amssymb}
\usepackage{amsmath}
\usepackage{balance}
\usepackage{rotating}
\usepackage{lscape}
\usepackage[utf8]{inputenc}
\usepackage{multicol,url,color}
\usepackage{multirow}
\usepackage{algorithm}
\usepackage[noend]{algpseudocode}
\usepackage{graphicx}
\usepackage{caption}
\usepackage{subcaption}
\usepackage{amsfonts,mathtools}
\usepackage{amssymb}
\usepackage{amsmath}
\usepackage{balance}
\usepackage{comment}
\usepackage{textcomp}
\usepackage{hyperref}
\usepackage{multirow}
\usepackage{colortbl}
\makeatother
\usepackage{mathtools} 
\usepackage{graphics}
\usepackage{placeins}
\usepackage{array}
 \usepackage{color}

\usepackage{xcolor}

\usepackage{soul,framed} 
\colorlet{shadecolor}{yellow}

\AtBeginDocument{%
  \providecommand\BibTeX{{%
    \normalfont B\kern-0.5em{\scshape i\kern-0.25em b}\kern-0.8em\TeX}}}

\newcommand{\deepwalk}{\emph{DeepWalk}}
\newcommand{\nodetwovec}{\emph{node2vec}}
\newcommand{\linet}{\emph{LINE}}
\newcommand{\verset}{\emph{VERSE}}
\newcommand{\structtwovec}{\emph{struc2vec}}
\newcommand{\hope}{\emph{HOPE}}

\newcommand{\rolx}{\emph{RolX}}
\newcommand{\graphsage}{\emph{GraphSAGE}}

\newcommand*{\myfont}{\fontfamily{lmtt}\selectfont}

\newtheorem{definition}{Definition}[section]

\begin{document}
\begin{frontmatter}
\title{A Comprehensive Analytical Survey on Unsupervised and Semi-Supervised Graph Representation Learning Methods}


\author[inst1]{Md. Khaledur Rahman}
\ead{morahma@iu.edu}
\affiliation[inst1]{organization={Computer Science, Indiana University},
            city={Bloomington},
            postcode={47408}, 
            state={IN},
            country={USA}}
            
\author[inst2]{Ariful Azad\corref{cor1}}
\ead{azad@iu.edu}
\affiliation[inst2]{organization={Intelligent Systems Engineering, Indiana University},
            city={Bloomington},
            postcode={47408}, 
            state={IN},
            country={USA}}
\cortext[cor1]{Corresponding author.}


\begin{abstract}
  Graph representation learning is a fast-growing field where one of the main objectives is to generate meaningful representations of graphs in lower-dimensional spaces. The learned embeddings have been successfully applied to perform various prediction tasks, such as link prediction, node classification, clustering, and visualization. 
  The collective effort of the graph learning community has delivered hundreds of methods, but no single method excels under all evaluation metrics such as prediction accuracy, running time, scalability, etc.  
  This survey aims to evaluate all major classes of graph embedding methods by considering algorithmic variations, parameter selections, scalability, hardware and software platforms, downstream ML tasks, and diverse datasets. 
  We organized graph embedding techniques using a taxonomy that includes methods from manual feature engineering, matrix factorization, shallow neural networks, and deep graph convolutional networks. We evaluated these classes of algorithms for node classification, link prediction, clustering, and visualization tasks using widely used benchmark graphs. We designed our experiments on top of PyTorch Geometric and DGL libraries and run experiments on different multicore CPU and GPU platforms.
  We rigorously scrutinize the performance of embedding methods under various performance metrics and summarize the results. 
  Thus, this paper may serve as a comparative guide to help users select methods that are most suitable for their tasks. 
  
\end{abstract}

\begin{keyword}
Graph Representation Learning \sep Graph Neural Network (GNN)\sep Classification \sep Link Prediction \sep Clustering
\end{keyword}

\end{frontmatter}




\section{Introduction}
A graph or network is a widely-used representation of relation data observed in a variety of applications, such as Online Social Network (OSN) analysis \cite{wasserman1994social}, scientific literature analysis \cite{shiffrin2004mapping}, and biomedical data analysis \cite{pavlopoulos2008survey}c. In a graph, a set of vertices represents entities, such as persons in an OSN, articles in scientific literature, and brain in neurons biomedical data. Similarly, a set of edges represents relationships among entities, such as friendship, citations, and neuron synapses in those networks, respectively. By analyzing graphs, we can extract and infer various structural and feature-based information from graph-structured data that can leverage many graph learning tasks, e.g., detection of communities, recommendation of friends or events, and prediction of online traffic. 

%


Machine learning (ML) methods have been successfully applied to many research areas for solving prediction or recommendation tasks~\cite{mair2000investigation}.
In particular, ML techniques for graphs have enjoyed tremendous success in many downstream tasks, such as node classification, link prediction, and clustering. The central task in graph ML is to learn the representation or embedding of graphs in a lower-dimensional space preserving the intrinsic properties of the original graphs. Then, the lower-dimensional embedding can be used for various prediction tasks. Over the last decade, researchers have developed a plethora of methods to learn the representation of graphs in both unsupervised and semi-supervised fashion. Early graph embedding methods extracted information from graphs by handcrafted feature engineering and then use them to prediction tasks~\cite{henderson2011s,akoglu2010oddball,gallagher2008leveraging}.
Then, various unsupervised methods such as \deepwalk{}~\cite{perozzi2014deepwalk} and {\it node2vec}~\cite{grover2016node2vec} were introduced to propel the graph embedding field to a new level. 
Finally, recent semi-supervised graph embedding techniques can automatically extract a meaningful representation of graphs for the downstream ML tasks~\cite{kipf2016semi,hamilton2017inductive}. 
The diversity of these graph embedding methods often makes it difficult to select the best method given an application and computing platform. 
The aim of this survey is to experimentally evaluate all major classes of graph embedding methods so that application and algorithm developers are able to make informed decisions based on the comparative performance of embedding algorithms.

\begin{table*}[!t]
    \centering
    \begin{tabular}{l|p{14cm}}
    \hline
    {\bf Experimental aspects} & {\bf Coverage of this survey }\\ 
    \hline
     Empirical performance    &  (a) empirical running time, (b) memory requirement, (c) scalability with respect to computational resources and graph size\\
         \hline
        ML tasks & (a) node classification (multiclass and multilabel), (b) link prediction (c) clustering, (d) visualization \\
        \hline
    Parameter sensitivity    & (a) random walk length,  (b) embedding dimension, (c) sampling rate, (d) convergence\\
    \hline
    Algorithm coverage & (a) feature-engineering based, (b) matrix-factorization based, (c) random-walk-based embedding, (d) GNNs \\
    \hline
    Hardware platforms & (a) Intel, AMD, and ARM CPUs (b) NVIDIA GPUs \\
    \hline
    Software platforms & (a) PyTorch Geometric (PyG) (b) Deep Graph Library (DGL) \\
    \hline
    Dataset coverage & (a) homogeneous and heterogeneous networks, (b) scale-free networks, (c) small and large networks\\
    \hline
    \end{tabular}
    \caption{The left column shows various experimental aspects related to graph representation learning and the right column shows a comprehensive set of tasks covered in this survey. }
    \label{tab:coverage}
\end{table*}

In addition to algorithmic diversity, evaluations of graph embedding methods are significantly influenced by the structure and size of graphs, software and hardware platforms where evaluations are performed, and various implementation details.
Most real-world graphs are sparse with irregular sparsity structures such as scale-free graphs~\cite{barabasi2003scale}. The irregular and often unpredictable sparsity of graphs makes it difficult to
predict the performance of an algorithm on a graph without an experimental evaluation. 
In addition to the property of irregular structure, graphs can be very large; e.g., the Facebook network has around 2 billion vertices and over a trillion edges \cite{lerer2019pytorch}. 
Many existing methods may fail to process such large graphs because of memory and scalability limitations. 
Parallel graph embedding methods such as VERSE~\cite{tsitsulin2018verse}, Force2Vec~\cite{rahman2020force2vec} and parallelized DeepWalk~\cite{perozzi2014deepwalk} attempted to address the scalability problem by using multi-core processors and GPUs.  
A handful of methods~\cite{lerer2019pytorch,ahmed2013distributed} employed distributed-memory parallelism to process graphs with hundreds of millions vertices.
Thus, to understand the performance (runtime, memory usage, scalability, etc.) of graph embedding methods, we need to run an extensive evaluation with a diverse collection of both small and large graphs on modern parallel computing systems. 
This survey aims to serve this evaluation purpose.
Through our experimental evaluations, we shed light on efficiency, effectiveness, and scalability for graph representation learning.





Several survey papers in the literature summarized existing methods, discussing applications, and in some cases, reporting their performance measures for different benchmark graphs \cite{cai2018comprehensive,goyal2018graph,hamilton2017representation,wu2019comprehensive,cui2018survey}. However, these surveys do not provide in-depth analyses of running time, scalability, and memory consumption, which are the bottlenecks for large scale graph embedding. We cover all these issues in this survey in addition to the adverse effect of the unconscious negative sampling approach used by several methods upon downstream prediction tasks. Thus, this paper provides performance-centric guidance about existing graph embedding methods. We also discuss several future research directions that may advance this rapidly evolving field.

\subsection{Our Contributions}
One of the main contributions of this this survey is to provide a comprehensive review of various experimental aspects related to unsupervised and semi-supervised graph embedding methods.
Table~\ref{tab:coverage} provides a summary of various experimental aspects surveyed this paper. We cover empirical performances (runtime, memory usage, scalability) on various hardware and software platforms for diverse algorithms and datasets. 
We also compare the performance of these algorithms in node classification, link prediction, clustering, and visualization tasks.  
Additionally, we survey the sensitivity of various methods with respect to their parameters. 
Hence, this paper complements other published surveys that extensively discussed existing graph embedding methods and applications. 
Below, we summarize our key contributions in this survey:

\begin{itemize}
    \item \textbf{Taxonomy of the embedding methods:} We provide a taxonomy of existing graph representation learning methods and discuss the relation in the respective category. Our discussion focuses on the categories of the corresponding methods. We briefly outline how some methods differ from others in the same category.
    
    \item \textbf{In depth running time and memory usage analysis:} We provide an enhanced discussion about the runtime and memory consumption which are mostly ignored in the existing survey papers. This may outline a new insight to develop scalable methods which will run fast and consume less memory.
    
    \item \textbf{Parameter sensitivity and negative sampling:} We show parameter sensitivity for several methods, such as the effect of choosing various dimensional embeddings, number of walks in random-walk-based method, etc. We also analyze the effect of choosing different numbers of negative samples. This analysis is significant since the results of prediction may significantly drop if the fraction of false negatives is high in the set of negative samples.
    
    \item \textbf{Comprehensive experimental evaluation:} We rigorously evaluate all embeddings and show comparative performance among several methods for several prediction tasks such as link prediction, multi-label classification, and graph reconstruction. We also compare methods based on their modularity score for clustering and visualizations.
    
    \item \textbf{Challenges and limitations:} We discuss some challenges that are commonly faced in the graph embedding domain. Then, we provide a list of limitations present in the existing graph embedding methods.
    
\end{itemize}

The rest of the survey is organized as follows: In Section \ref{sec:preli}, we provide some preliminary definitions and notations that are used throughout the paper. We formally define the graph embedding problem in Section \ref{sec:problem}. We discuss the taxonomy of existing graph embedding methods in Section \ref{sec:methodologies}. We discuss benchmark datasets and experimental analyses in Section \ref{sec:experimentalcomp}. Finally, we discuss the challenges and limitations of the existing methods in Section \ref{sec:limitations}.

\section{Preliminaries}
\label{sec:preli}
In this section, we define several terminologies that are used throughout the rest of the paper. For the sake of simplicity, we assume that our input graph is undirected and unweighted.

We represent a graph by $G = (V, E)$, where, $V$ is the set of vertices, $E$ is the set of edges and each $z_i$ of a set $Z$ represents 1-dimensional embedding of $i^{th}$ vertex. We represent neighbors and the degree of vertex $u$ in $G$ by $N(u)$ and $deg(u)$, respectively. An adjacency matrix $A$ of graph $G$ is defined as follows:
\[
    A_{ij}= 
\begin{cases}
    1,& \text{if vertex } i \text{ and vertex } j \text{ are connected by an edge} \\
    0,              & \text{otherwise}
\end{cases}
\]

\begin{definition}
(One-hot encoded representation) Given a graph $G$ with $|V| = n$, one-hot encoded representation of $V$ is denoted by a $n \times n$ dimensional matrix $\mathcal{V}$ such that $\mathcal{V}_{ij} = 1$, when $i = j$, otherwise $\mathcal{V}_{ij} = 0$ and $i^{th}$ row is a $1 \times n$ dimensional binary vector representation for $i^{th}$ vertex.
\end{definition}

\begin{definition}
(Node embedding) Given a graph $G$ with $|V|=n$ and $|E|=m$, a node embedding is a mapping function $f : u \rightarrow  {z_u \in {\rm I\!R}^d} \; \forall u \in  V$ such that $d\ll n$ and the function $f$ preserves some similarity measure defined on graph $G$.
Here, $d$ is the dimension of the embedding. 
\end{definition}

\begin{definition}
(Similarity function) We define a similarity function over $Z$ to be any pairwise function $\sigma: z_u \times z_v \rightarrow [0, 1]$. This similarity function is symmetric, i.e., for any $z_u$ and $z_v$, $\sigma(z_u, z_v) = \sigma(z_v, z_u)$. 
\end{definition}

\begin{definition}
(Identity matrix) We define a matrix $I$ to be an indentity when all of its diagonal entries are $1$'s and all other entries in the matrix are $0$'s, i.e., $\forall i$ $I_{ii} = 1$, and $I_{ij}=0$, when $i \neq j$. 
\end{definition}

\begin{definition}
(Degree matrix) We represent the degree matrix $D$ of a graph $G$ by filling all diagonal entries by degree of corresponding vertex and all other entries as zeros, i.e., $\forall i\; D_{ii} = deg(i)$, and $D_{ij}=0$, when $i \neq j$.
\end{definition}

We represent a differentiable non-linear function by $\phi$ which has different gradients at different points. The graph Laplacian matrix \cite{merris1994laplacian} is represented by $L$ which is obtained from the degree matrix and adjacency matrix of the graph as follows: $L = D - A$.

As ``graph" is also referred to as ``network" in the literature \cite{barabasi2016network}, we use these two terms interchangeably throughout the rest of our discussion. Similar to this, we also use other corresponding terminologies interchangeably like vertex vs. node, edge vs. link, and graph embedding vs. graph representation learning. 

\subsection{Problem Definition}
\label{sec:problem}
\begin{figure}[!htb]
    \centering
    \includegraphics[width=0.99\linewidth]{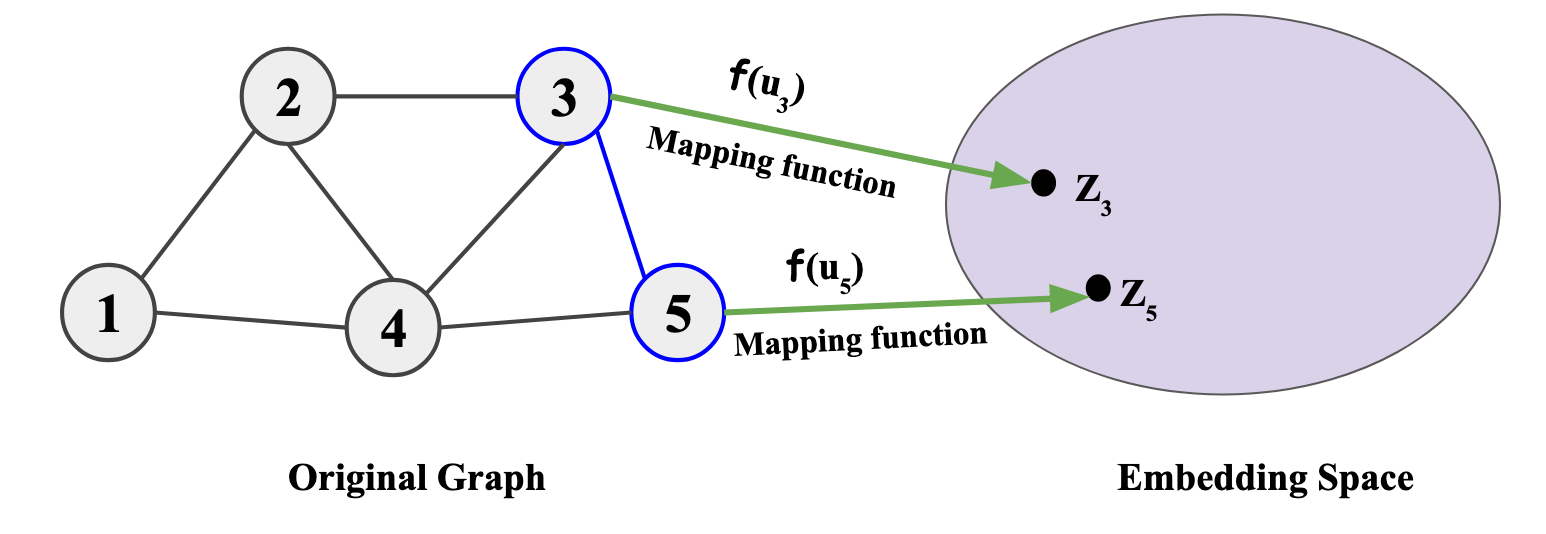}
    \caption{A conceptual figure to show that a mapping function preserves the notion of similarity of the original graph in embedding space. $f$ is the mapping function and $z_i$ is the representation of $i^{th}$ vertex in the embedding space.}
    \label{fig:definition}
\end{figure}
We can formally define the problem as follows: Given a graph $G$, we map it onto a $d$-dimensional space $\mathcal{Z} \in {\rm I\!R}^{n \times d}$ such that the intrinsic properties of the original graph are preserved as much as possible. In other words, if we pick $|N(u)|$ number of vertices for vertex $u$ based on the similarity function $\sigma$ in the embedding space, then we should get all the neighbors of the original graph. Precisely, this represents the embedding of nodes in a network. In Fig. \ref{fig:definition}, we show pictorially that the embedding of vertices $3$ and $5$ preserves some notion of similarity in the embedding space as shown by $z_3$ and $z_5$, respectively. So, if we find similarities among all other vertices with respect to vertex 3 in the embedding space, then the similarity between $z_3$ and $z_5$ will be higher than the similarity between $z_3$ and $z_1$. A mapping or encoding function $f$ is applied to generate such embedding of vertices.

Node embedding can be further extended to edge representation or even full representation of the graph. Simply, we can create an edge embedding of a graph by concatenating the embedding of two incident vertices of an edge. Similarly, a full representation of the graph can be deduced by taking the average contributions of all vertices for a given dimension. We describe several other kinds of edge embedding in Section \ref{sec:linkpred}.

\section{Methodologies}
\label{sec:methodologies}
\begin{figure}[!ht]
    \centering
    \includegraphics[width=0.98\linewidth]{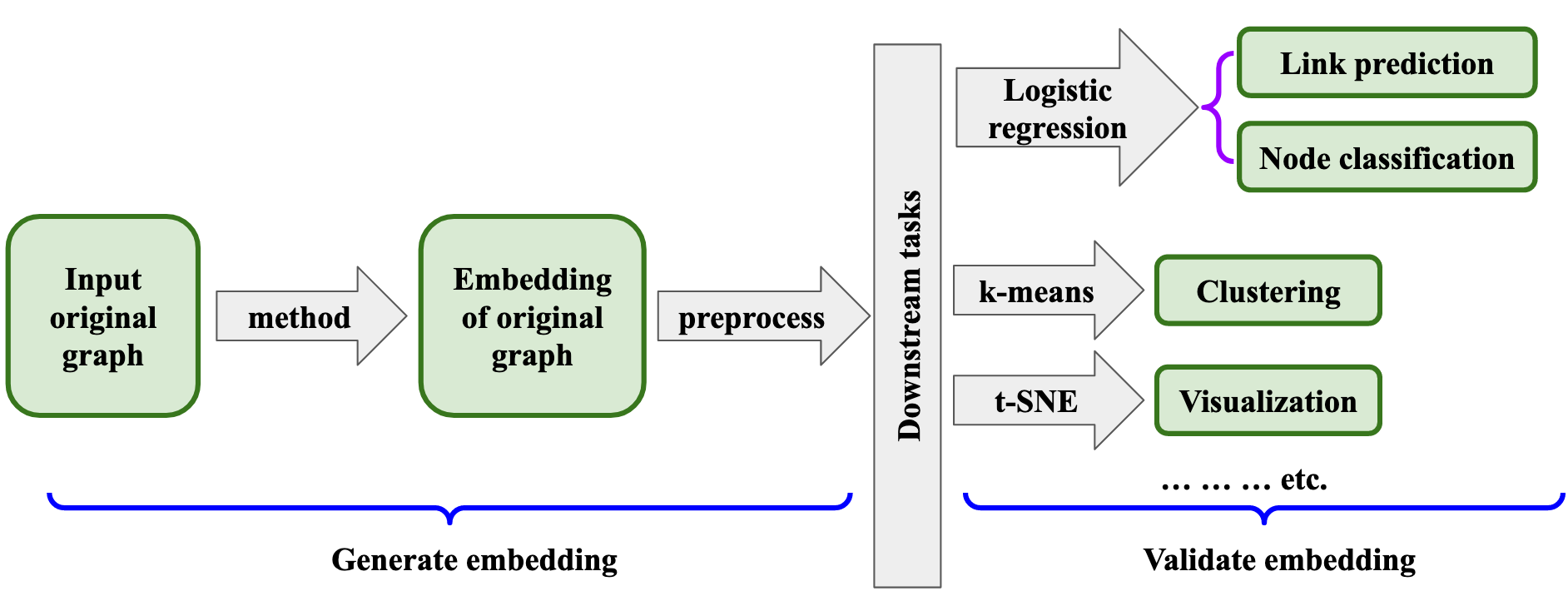}
    \caption{A general framework that most of the existing methods follow.}
    \label{fig:generalsteps}
\end{figure}

Most of the existing methods follow a general framework to generate and validate their graph embedding. We can summarize this precisely by Fig. \ref{fig:generalsteps}. In the embedding generation step, a method is employed, and in the validation step, several prediction tasks are carried out using the learned embedding. Sometimes, embedding is pre-processed for performing latter tasks. For example, a portion of a dataset is used for training and the remaining portion of the dataset is used for testing. Generally, logistic regression is used for link prediction or node classification tasks, with the K-means algorithm being applied to cluster the graph, and the t-SNE \cite{maaten2008visualizing} dimensionality reduction tool being used to visualize the embedding data.

\begin{figure}[!ht]
    \centering
    \includegraphics[width=0.95\linewidth]{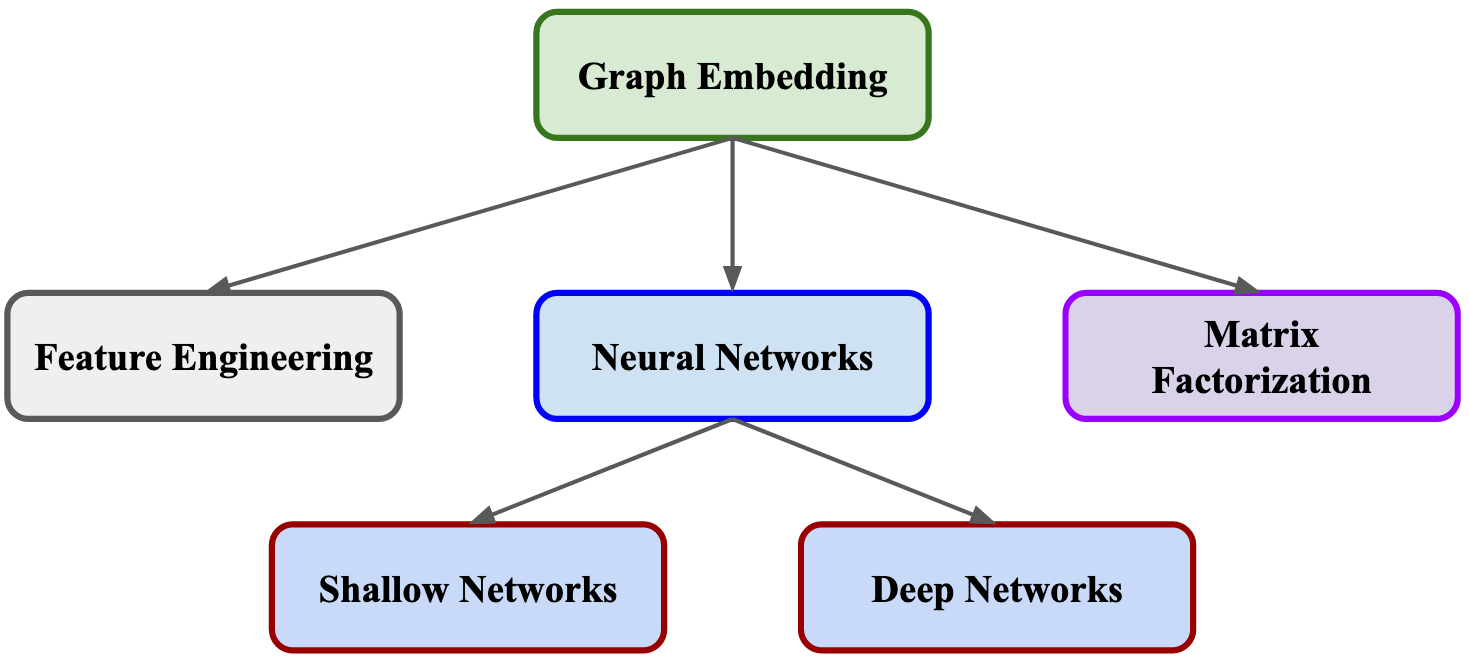}
    \caption{A taxonomy of graph embedding methods.}
    \label{fig:taxonomy}
\end{figure}

The methods for generating graph embedding can be classified into several categories based on the procedures, objective functions, and types of optimizations used. We propose a taxonomy of these categories as shown in Fig. \ref{fig:taxonomy}. The early methods (during 2008 to 2011) of graph embedding were based upon the feature engineering of graphs. Then, with the increasing popularity of neural networks (shallow as well as deep), some methods were introduced (e.g., \deepwalk{} in 2014) and evolved which borrowed the idea from language modeling. At the same time, some matrix factorization methods were introduced and evolved alongside the shallow networks-based methods. In 2017, the paper on graph convolutional neural networks \cite{kipf2016semi} was published and gained much popularity. In the following, we describe each of these categories elaborately.

\subsection{Feature Engineering}
Early methods on graph mining tasks were mostly based on supervised (hand-crafted) feature engineering \cite{henderson2011s,akoglu2010oddball,gallagher2008leveraging}. Those hand-crafted features are designed with respect to some common intuitions that are supposed to infer some meaningful information from graphs. For example, \emph{ReFeX} by \citeauthorandyear{henderson2011s} extracts a predetermined set of features such as the degree of a vertex (in/out-degree in case of directed graph and weights in case of a weighted graph), number of edges in the egonet (number of incoming/outgoing edges in case of a directed graph) and recursive features by summing up or averaging two types of previous features. \citeauthorandyear{gallagher2008leveraging}, on the other hand, focuses on extracting features such as the average degree of egonet, the number of incident links, betweenness centrality, and clustering coefficient. These extracted features are then arranged in a vector format with their corresponding label and fed to a standard machine learning classifier for making predictions. The prediction tasks based on the extracted features are conducted in the following way: First feature vectors with corresponding labels are partitioned into two sets, e.g., $T\%$ of the samples are used for training and the rest of the datasets are used for testing. Then, logistic regression or any other standard machine learning method can be trained to learn the parameters of the model using the training dataset. Finally, the prediction task is performed using the test dataset based on the trained model. These types of methods do not always perform well on graph mining tasks because a predefined set of features is not always enough to capture the latent characteristics of the graph. Sometimes, they are found to be difficult to generalize across different types of graphs due to the highly irregular structure. Thus, more advanced methods have been evolved over time which is briefly discussed in the following sections.
\subsection{Matrix Factorization}

Matrix Factorization is an effective technique to decompose a matrix into two lower-dimensional rectangular matrices \cite{koren2009matrix}. This technique has been successfully applied to recommender systems \cite{koren2009matrix}, data compression \cite{yuan2005projective}, and spectral clustering \cite{ding2005equivalence}. It was first applied to graph factorization in large scale by \citeauthorandyear{ahmed2013distributed}. 
Matrix $\mathcal{A}$ of dimension $M \times N$ can be decomposed into two lower-dimensional rectangular matrices $\mathcal{B}$ and $\mathcal{C}$ having dimensions $M \times R$ and $R \times N$, respectively. Generally, $R$ is much less than $M$ and $N$. In the case of factorizing an adjacency matrix, we have $M=N$ and $\mathcal{C=B'}$, where $\mathcal{B}'$ is a transpose matrix of $\mathcal{B}$. The loss function for this problem is defined by the following Equation:

\begin{equation}
    \label{eqn:mfloss}
    f(A,B,\lambda) = \frac{1}{2}\sum_{(i,j)\in E} (A_{ij} - B_i.B_j')^2 + \frac{\lambda}{2}\sum_i\parallel B_i \parallel^2 
\end{equation}

In Equation \ref{eqn:mfloss}, $\lambda$ is a regularization parameter. To optimize the loss function, we need to calculate the gradient of Equation \ref{eqn:mfloss}. Doing so, we get the following:
\begin{equation}
    \label{eqn:mfgradient}
    \frac{\partial f(A,B,\lambda)}{\partial B_i} = -\sum_{j\in N(i)} (A_{ij} - B_i.B_j')B_j + \lambda B_i
\end{equation}

Now we can iteratively optimize the loss function by updating $\mathcal{B}$ using Stochastic Gradient Descent (SGD)~\cite{recht2011hogwild}. \citeauthorandyear{ahmed2013distributed} implemented this method in distributed systems to optimize the loss function in both synchronous and asynchronous ways. 

\citeauthorandyear{luo2011cauchy} propose a graph embedding technique that uses Cauchy distribution\footnote{\url{https://en.wikipedia.org/wiki/Cauchy_distribution}} to preserve the local topology of the graph in the embedding space~\cite{luo2011cauchy}. Cauchy graph embedding optimizes the following objective function:

\begin{equation}
\label{eqn:cauchy}
    \begin{split}
        \max_z \sum_{uv}\frac{w_{uv}}{(z_u-z_v)^2 + \gamma^2},\\
        s.t. \sum_u z_u^2 = 1, \sum_u z_u = 0.
    \end{split}
\end{equation}
Here, $\gamma$ is the scaling parameter, $w_{uv}$ is the edge weight between vertices $u$ and $v$, and $(z_u-z_v)^2$ represents the distance between vertices $u$ and $v$ in the embedding space. Thus the whole term indicates a similarity distribution that we want to maximize. Note that $z_u=0$ will maximize the objective function but collapse all vertices to a single one, which is not expected i.e., we do not want to maximize similarity where two vertices are located far away in terms of graph-theoretic distance. Thus, the authors add an extra term $z_u^2=1$ in the constraints which prevent the embedding of all the vertices from collapsing into the origin. 

\emph{GraRep} method, introduced by \citeauthorandyear{cao2015grarep}, first constructs a transition probability matrix from an adjacent matrix to generate the embedding of the graph. The transition probability of a vertex $u$ to another vertex $v$ can be obtained from the adjacent matrix by normalizing the $u^{th}$ row dividing by $deg(u)$ and then taking $(u,v)$ entry of the normalized matrix. In particular, they denote the probability of transition from vertex $u$ to $v$ after $k$ steps as $p_k(v|u) = A_{u,v}^k$, where $A^k$ is a $k^{th}$-order transition probability matrix. The reason for constructing such a matrix is to capture high-order similarity among vertices that have been found effective in random walk-based methods \cite{perozzi2014deepwalk}. Then they deduce the loss function using the following equation:

\begin{equation}
    \label{eqn:grareploss}
    L_k(u,v) = A_{u,v}^k.\log(\sigma(u.v)) +\frac{\lambda}{n}\sum_{u'}A_{u',v}.\log(\sigma(-u.v))
\end{equation}

In Equation \ref{eqn:grareploss}, $\sigma(u.v)$ generally represents a sigmoid function defined as $\sigma(u.v)=\frac{1}{1+e^{-u.v}}$. Authors of \emph{GraRep} mostly follow the work of \citeauthorandyear{levy2014neural} in neural word embedding.
They calculate the gradient of Equation \ref{eqn:grareploss} and formulate the problem as a matrix factorization where each entry of the original matrix is computed in the following way:

\begin{equation}
    \label{eqn:grareploss2}
    x = u.v = \log \frac{A_{u,v}^k}{\sum_{u'}A_{u',v}^k} - \log(\frac{\lambda}{n})
\end{equation}

\citeauthorandyear{cao2015grarep} create a matrix $\mathcal{X}$ for all vertices using Equation \ref{eqn:grareploss2} and replace all negative values by 0's. Then, they factorize $\mathcal{X}$ into $U\Sigma V'$ using Singualar Value Decomposition (SVD) \cite{golub1971singular}, where $\Sigma$ represents the singular values in a diagonal matrix. Finally, they take $U(\Sigma)^{\frac{1}{2}}$ as the representation of the graph. 

HOPE~\cite{ou2016asymmetric} is another matrix factorization-based method that preserves asymmetric transitivity of the graph by creating a higher-order proximity matrix similar to \emph{GraRep}. A generalized proximity matrix $S$ is constructed using a similarity metric and an adjacent matrix of the graph. Then, the objective function becomes to minimize $\parallel S - U_s.U_t\parallel^2_F$, where $U_s$ and $U_t$ are two embeddings for the source and the target, respectively. They formulate the proximity matrix $S$ from the local and global proximity as follows:

\begin{equation}
    S = M_g^{-1}.M_l
\end{equation}
Here, $M_g$ represents the global proximity matrix and $M_l$ represents the local proximity matrix. Generally, the global proximity has a recursive definition that can iterate over the whole graph to find global transitivity, whereas the local proximity has no recursive definition and tends to find local transitivity from neighborhoods. Katz Index \cite{katz1953new} measures the relative influence of a node within a network, which can be used as a proximity measure. Thus, $S$ can be defined by its Katz Index as follows:

\begin{equation}
\label{eqn:hope}
\begin{split}
    S & = \sum^\infty_{l=1}\beta.A^l = \beta.A.S + \beta.A\\
    S & = (I-\beta.A)^{-1}.\beta.A
\end{split}
\end{equation}
In Equation \ref{eqn:hope}, $A$ is the adjacency matrix, $\beta$ is a decay parameter, $M_g = I-\beta.A$ where $I$ is the identity matrix, and $M_l = \beta.A$. Specifically, authors incorporated the Katz Index \cite{katz1953new} and the Rooted PageRank to construct global proximity matrices, and Common Neighbors and Adamic-Adar~\cite{adamic2003friends} to construct local proximity matrices. They also choose SVD to generate the final embedding.

Other techniques use a similar formulation, such as structure-preserving embedding~\cite{shaw2009structure}, which maps graphs onto euclidean space which is also effective for visualization. In summary, SVD and graph laplacian have a strong influence on matrix factorization-based methods. Technically, these types of models have higher computation and memory consumption costs.

\subsection{Neural Networks}
Neural networks have been widely used to solve prediction tasks in various research domains. In a neural networks-based method, there are three layers stacked in cascaded style, namely, (i) Input layer, (ii) Hidden layer(s), and (iii) Output layer. Generally, one hot-encoded vector of a vertex or input feature is fed to the network as input. The output layer contains target vertices or label(s) that are also one-hot encoded vectors. There can be multiple hidden layers in a neural network-based model. Hence, we can divide the neural network-based methods into two categories: (1) Shallow networks having one hidden layer and (2) Deep networks having more than one hidden layer. We describe each of these categories below:

\begin{figure}[!ht]
    \centering
    \includegraphics[width=0.37\linewidth]{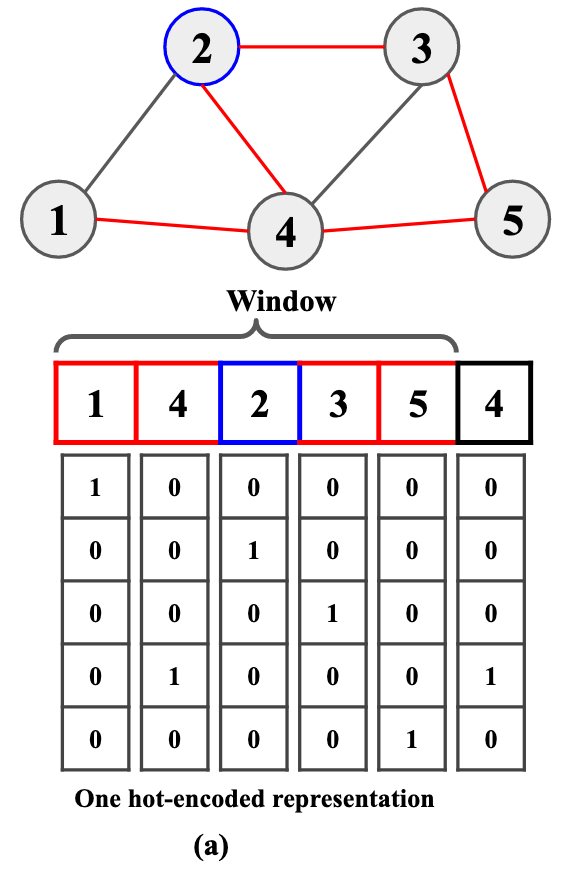}\hspace{0.20cm}
    \includegraphics[width=0.58\linewidth]{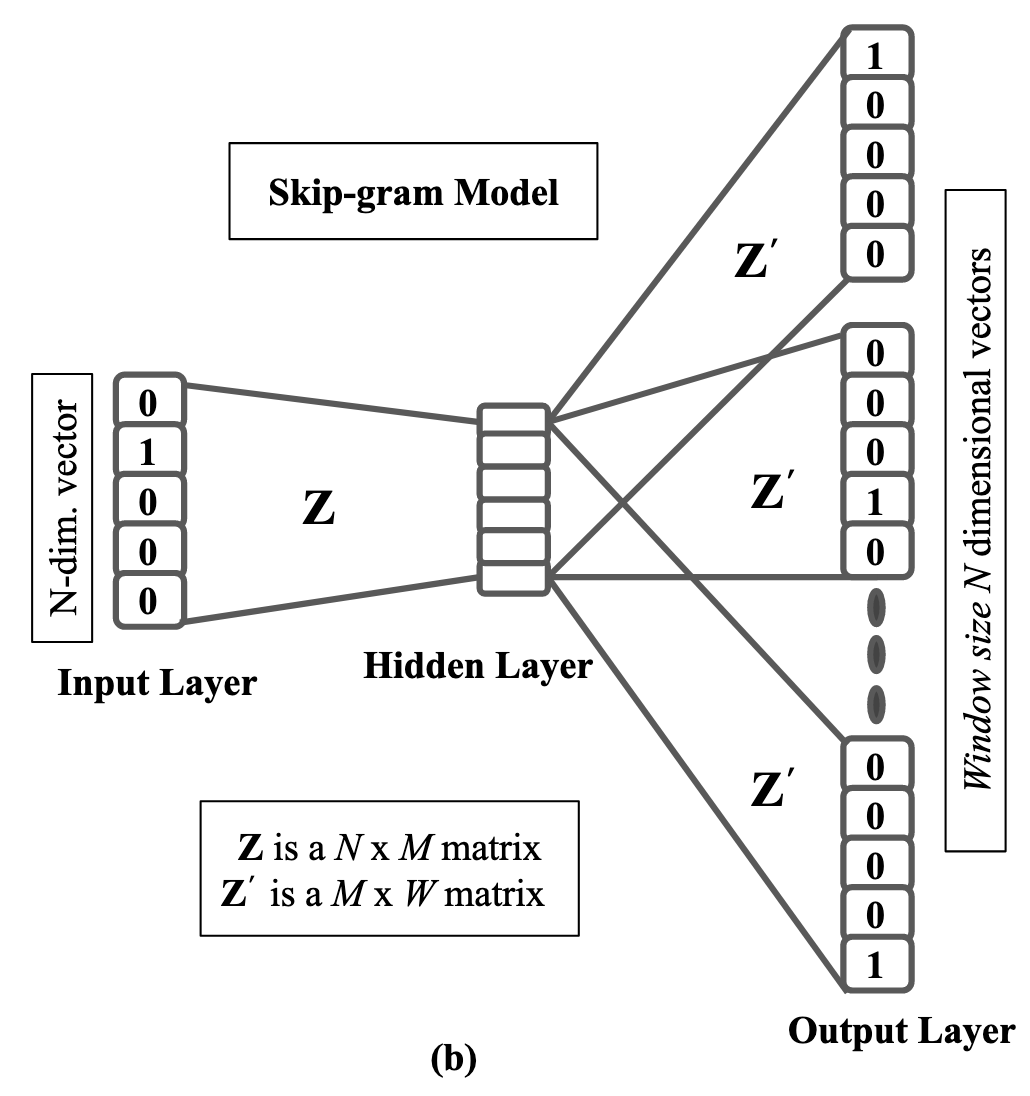}
    \caption{(a) An example graph with one-hot encoded representation and sliding window. (b) Skip-gram model used in random-walk based methods such as \deepwalk{}~\cite{perozzi2014deepwalk} and \nodetwovec{}~\cite{grover2016node2vec}}
    \label{fig:skipgram}
\end{figure}

\subsubsection{Shallow Networks}
In this type of model, a random walk is performed on a graph and a path of a set of vertices is sampled for each random walk. Now, considering each vertex in the path as a word in a sentence, we can model this problem in a similar way to that of word representation in the NLP domain \cite{mikolov2013distributed,mikolov2013efficient}. The basic task in word representation is to find the representation of words in vector space that preserves both syntactic and semantic meanings. This type of representation also becomes helpful for downstream analysis. In the graph embedding problem, we want to find the representation of vertices that preserves structural and neighborhood relationships in a graph. A popular model in the word embedding domain is \emph{word2vec} that uses the skip-gram model \cite{mikolov2013distributed} for optimizing the prediction of source context words from a target word. A pioneering graph embedding method, called \deepwalk{} \cite{perozzi2014deepwalk}, infers the analogy between word embedding and graph embedding, and uses the skip-gram model for training. In this model, target vertices are fed to the input layer and the weights of the network are updated to learn the semantics of the graph based on the source context vertices.

Suppose a random walk has the path $v_1, v_2, v_3, \ldots v_l$ such that $1 \le k \le l$, where $v_k$ is the target vertex and all others are context vertices. To compute likelihood, a fixed window size of $w$ slides over the path. Then, the likelihood of the target vertex based on the skip-gram model would be $P((v_{k-w}, \ldots, v_{k-1}, v_{k+1}, \ldots, v_{k+w})|\sigma(v_k))$, where $\sigma$ represents a similarity function. The skip-gram model assumes that context vertices are conditionally independent. Hence, using negative log-likelihood, we can minimize the following objective function:

\begin{equation}
\label{eqn:objskipgram}
    -\log \prod_{i=k-w}^{k+w}P(v_i|\sigma(v_k)
\end{equation}

The probability $P(v_i|\sigma(v_k)$ in Equation \ref{eqn:objskipgram} is calculated by a normalized softmax function which can defined as follows:
\begin{equation}
\label{eqn:softmax}
P(v_i|\sigma(v_k) = \frac{e^{z_k.z_i}}{\sum_{i'}e^{z_k.z_i'}}
\end{equation}
Note that the normalizing factor $\sum_{i'}e^{z_k.z_i'}$ is computationally expensive which is asymptotically $O(n)$. A more efficient way is to use hierarchical softmax~\cite{morin2005hierarchical} which takes $O(\log n)$ time to approximate Equation \ref{eqn:softmax}.

In Fig. \ref{fig:skipgram} (a), we have shown a toy graph with its representation. Let us sample a random walk from the graph as $1\rightarrow 4 \rightarrow 2 \rightarrow 3 \rightarrow 5 \rightarrow 4$ and window size is $2$. In Fig. \ref{fig:skipgram} (b), we have shown the skip-gram model for the sampled random walk where $2$ is the target vertex and $\{1,4, 3, 5\}$ are context vertices in the current window as shown in Fig.\ref{fig:skipgram} (a). The input vertex is a $N$ dimensional one-hot-encoded vector. The hidden layer is a $M$ dimensional vector representing activation units. $Z$ and $Z'$ are two weight matrices that represent the embedding of the network and are learned by optimizing the objective function. A popular choice for optimizing the objective function is to use Stochastic Gradient Descent \cite{bottou2010large}. Thus the runtime for \deepwalk{} is $O(dn\log n)$, where $d$ is the dimension of embedding for a graph with $n$ vertices.

\begin{figure}
    \centering
    \fbox{\includegraphics[width=0.45\linewidth, height=2.2cm]{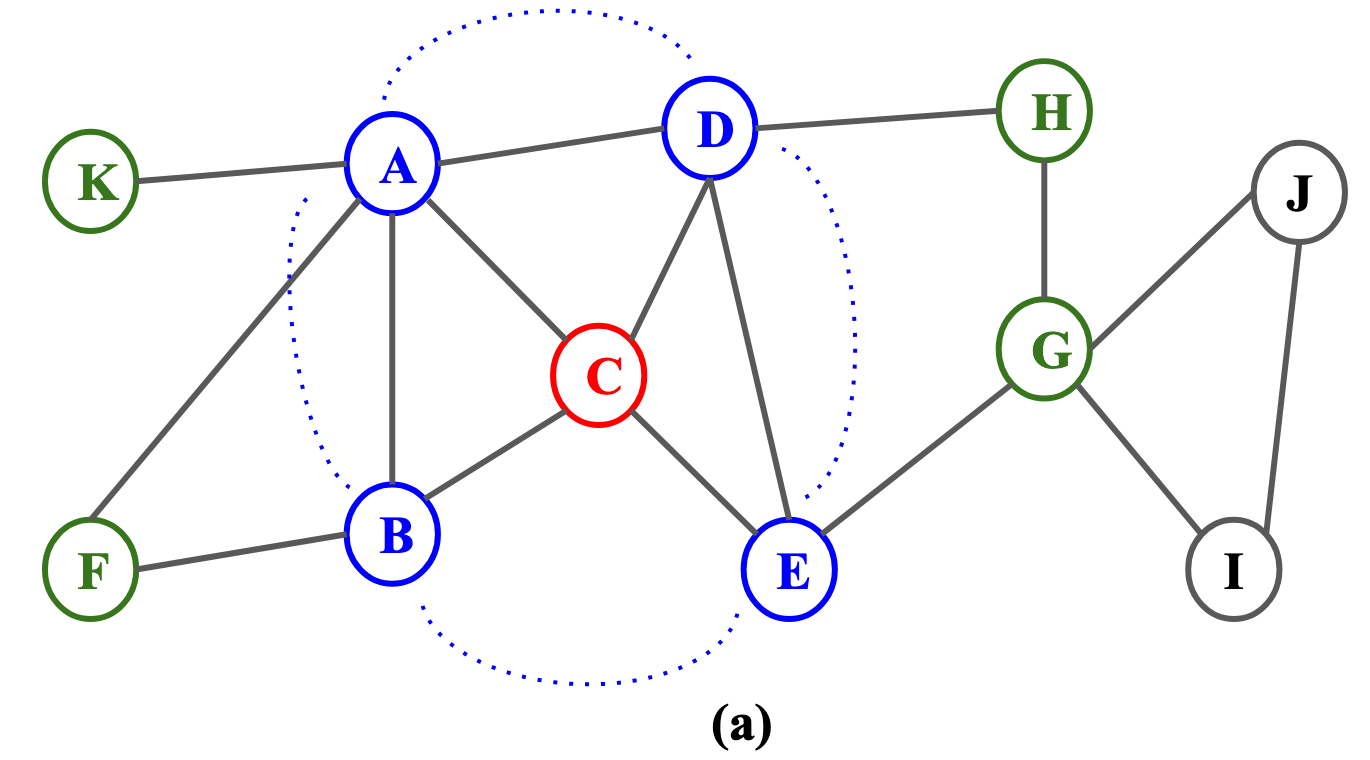}}
    \fbox{\includegraphics[width=0.45\linewidth, height=2.2cm]{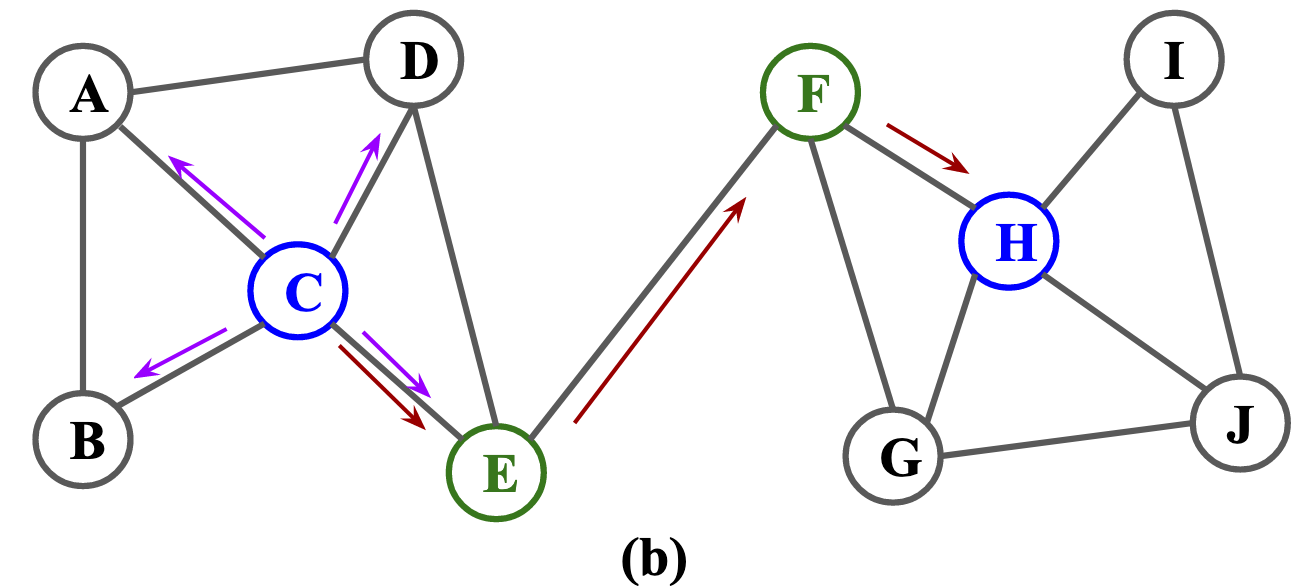}}
    \caption{(a) First order proximity vertices of $C$ are $A,B,D$, and $E$. Second order proximity vertices of $C$ are $K, F, G$, and $H$. (b) A DFS traversal walk $C\rightarrow E\rightarrow F \rightarrow H$ contributes to homophily characteristics discovery in the graph whereas a BFS traversal from $C$ to $\{A,B, D, E\}$ contributes to structural discovery.}
    \label{fig:networkstructures}
\end{figure}

After the reputation of \deepwalk{}, many other methods have been introduced in the literature that is based upon a similar optimization model. \citeauthorandyear{tang2015line} introduces \linet{} that optimizes an objective function based on first order and second order proximities. Instead of taking a random walk-based path sampling, \linet{} finds similarity considering all neighbors that are in the 1-hop distance as the first order proximity and 2-hop distance as the second-order proximity. For unweighted graph, first-order proximity is optimized by following the reduced KL-divergence Equation \ref{eqn:1stline} whereas second-order proximity optimization follows Equation \ref{eqn:objskipgram}:

\begin{equation}
    \label{eqn:1stline}
    O_1 = -\sum_{(i,j)\in E} \log p_1(v_i,v_j)
\end{equation}
where, $p_1(v_i, v_j) = \frac{1}{1+e^{-z_i.z_j}}$. In Fig. \ref{fig:networkstructures} (a), we represent the first order proximity vertices (which are $A, B, E$, and $D$) of $C$ by blue circles which are in $1$-hop distance from $C$. Similarly, second order proximity vertices of $C$ are $K, F, G$, and $H$ (shown by green circles in Fig. \ref{fig:networkstructures} (a)). It further improves running time by using negative sampling instead of hierarchical softmax. Thus the overall runtime for \linet{} becomes $O(sdn)$, where $s$ is the number of negative samples and generally, $s < \log n$.  

\citeauthorandyear{grover2016node2vec} propose \nodetwovec{} which is another popular work that follows similar optimization function to the \deepwalk{} and negative sampling like \linet{}. It samples biased random walks from the network. Basically, it introduces two controlling parameters which can simultaneously discover homophily and structural equivalence in the network. A homophily relation determines the similarity among vertices that are close to one another or belong to the same clusters or community. On the other hand, structural equivalence characterizes vertices having a similar role in the graph. Authors of \nodetwovec{} state that Breadth-First Search (BFS) based random walks are likely to contribute to structural equivalence discovery in the graph whereas Depth First Search (DFS) based random walks are likely to contribute to homophily discovery in the network. Though it is laborious to find optimal values for each controlling parameter, this method introduces some notion of universality in the tandem discovery of different latent characteristics. The running time for this method is the same as \linet{}. In Fig. \ref{fig:networkstructures}(b), we show two types of walks represented by colored arrows where crimson-colored arrows to direct DFS search and violet-colored arrows direct BFS search. It is easy to infer from Fig. \ref{fig:networkstructures} (b) that $C$ is structurally equivalent to $H$ which is a hub in the network. Similarly, $E$ and $F$ are structurally equivalent which represent bridges in the network.

\citeauthorandyear{tsitsulin2018verse} introduce \verset{}, a more recent work based on random walk, which can instantiate the embedding using several similarity functions such as personalized PageRank \cite{page1999pagerank}, adjacency similarity~\cite{tsitsulin2018verse} and simrank~\cite{jeh2002simrank}. Authors optimize the objective function using cross-entropy as it is proportional to KL-divergence. They also claim that the stationary distribution of random walk eventually converges to personalized PageRank. \citeauthorandyear{rahman2020force2vec} introduce a parallel force-directed graph embedding method, called Force2Vec, that can efficiently utilize the multi-core architecture and effectively generate embedding of graphs.

Most of the previously discussed shallow networks work better for homophily prediction. \citeauthorandyear{ribeiro2017struc2vec} propose \structtwovec{} to the solve \emph{structural equivalence} problem in networks. To begin, it creates a multi-layer weighted graph from the original graph where each layer contains an equal number of vertices and edges to the original graph. Weights in each layer are calculated as inversely proportional to structural distance as follows: If there is an edge between vertices $u$ and $v$ in $G$, then weight $w(u,v)=e^{-f(u,v)}$, where $f(u,v)$ is the structural distance between $u$ and $v$ which is determined by the recursive Equation \ref{eqn:struc2vec}.

\begin{equation}
    \label{eqn:struc2vec}
    f_k(u,v) = f_{k-1}(u,v) + g(D_k(u),D_k(v))
\end{equation}
Here, $k$ represents $k$-hop neighborhoods, $f_{-1} = 0$, $D_k(u)$ represents the ordered degree sequence of all vertices that are $k$-hop distant from $u$, and $g$ measures the distance between degree sequences using Dynamic Time Warping (DTW) \cite{rakthanmanon2013addressing}. There are directed edges between two consecutive layers that connect corresponding vertices $u$'s with an edge weight determined by the logarithmic number of edges incident to $u$ having a larger weight than the average weight of the complete graph in that respective layer. Then, \structtwovec{} samples random-walks from the multi-layered graph based on the normalized weight as the probability. The rest of the procedures are similar to \deepwalk{}, as described previously. \citeauthorandyear{ribeiro2017struc2vec} apply \structtwovec{} to three novel airport datasets and empirically show that their method achieves better performance over previously existing methods.

Some other works use the random-walk based model for graph embedding such as \emph{GEMSEC}~\cite{rozemberczki2019gemsec} which generates embeddings that are more useful for evaluating clusters and \emph{SINE}~\cite{zhang2018sine} which generates embeddings that are more effective for incomplete graphs. In summary, the skip-gram model, introduced in natural language modeling, has a huge influence on shallow network-based graph embedding techniques.

\subsubsection{Deep Networks}
\begin{figure}[!ht]
    \centering
    \includegraphics[width=0.9\linewidth]{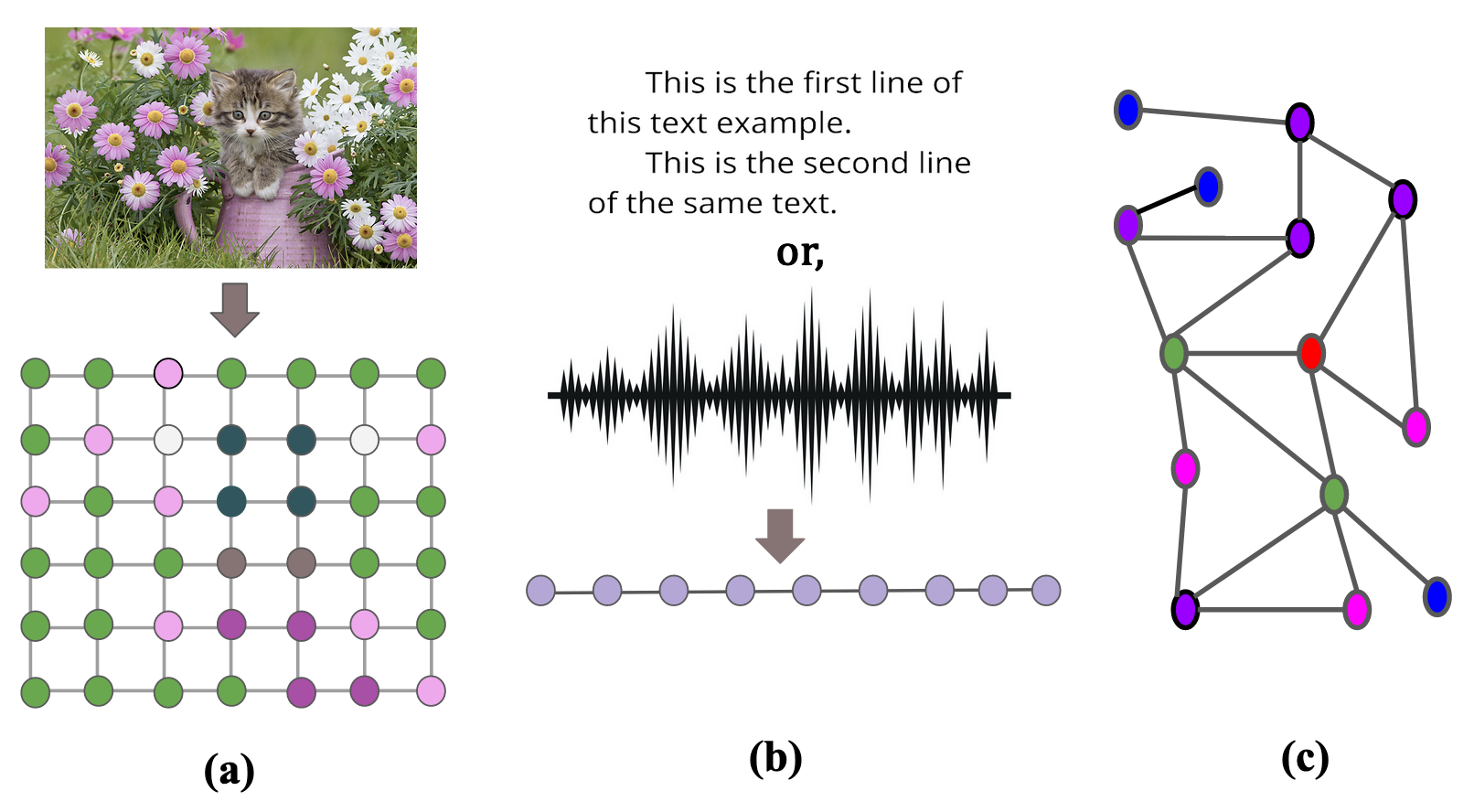}
    \caption{(a) An image can be converted to grid graph. (b) Text or speech signal can be converted to line graph. (c) An example of graph having irregular structure such as different vertices have different degrees.}
    \label{fig:deepgraph}
\end{figure}
Deep learning \cite{lecun2015deep} has become a very powerful framework which has been successfully applied in many research domains such as image classification \cite{litjens2017survey,rahman2019evaluating}, bioinformatics \cite{min2017deep}, speech recognition \cite{amodei2016deep}, etc. The idea of a convolutional operator \cite{lawrence1997face} has made deep learning more flexible to use. It can extract features from structured input by itself which lessens the tiresome handcrafted feature extraction and automates the classification process. In deep convolutional neural networks, generally, a filter\footnote{A smaller two-dimensional grid and the dimension of a filter is less than the dimension of the image.} slides over the image in each convolutional layer, performs matrix multiplication, and pools information from it as features. This convolutional operator works well for regularly structured data as it can extract spatial features by matrix multiplication and pooling. For example, we show an image in Fig. \ref{fig:deepgraph} (a) where it can be represented as a grid graph. Then a filter slides over it from left to right and then top to bottom to capture spatial information from the image. Similarly, texts or speech signals (see Fig. \ref{fig:deepgraph}(b)) can be represented as a line graph which is easy to design a convolutional operator for. However, it is very challenging to design a convolutional operator for graphs because of their highly irregular structure. In Fig. \ref{fig:deepgraph} (c), we show such a graph where different vertices have different degrees. 

Sometimes, preliminary knowledge becomes available for a graph, such as users' profiles, images in social networks or, gene expressions profiles in biological networks. We may also want to use such information that may carry meaning about the graph. Unfortunately, other approaches, like matrix-factorization or shallow networks, can not deal with such information. In addition to this, embedding for dynamic graphs is also essential as new nodes (i.e., users) are coming to social networks daily, and we may not want to generate embeddings for new nodes by running the whole procedure again. This is termed as \emph{transductive} problem settings i.e., a graph embedding model can generalize the network so that it can generate embeddings for new incoming nodes. On the other hand, an \emph{inductive} setting can generalize embeddings for completely new graphs of similar type e.g., embedding of Protein-Protein Interaction (PPI) networks will be similar, and as a result, knowledge of generating embedding for PPI networks can be applied to future PPI networks. Previously discussed graph embedding methods lag behind these characteristics. \emph{Planetoid} covers different settings with the help of random-walk-based methods discussed previously, but its performance is not impressive~\cite{yang2016revisiting}. Thus, efforts have been made to develop methods that can apply convolutional operators on graphs using preliminary knowledge of the graph while also supporting \emph{transductive} and \emph{inductive} settings. In the following, we briefly discuss some existing graph convolutional network-based methods.

There are several works on graph neural networks \cite{gori2005new,scarselli2008graph,gallicchio2010graph}. \citeauthorandyear{kipf2016semi} were the first to introduce the idea of Graph Convolutional Network (\emph{GCN}). They designed spectral convolutional operators to capture structural information from neighbors. They define spectral convolution as a multiplication of a signal $x\in {\rm I\!R}^n$ with a filter $g_w = diag(w)$ parameterized by $w\in {\rm I\!R}^n$ in the Fourier domain and further approximate it as the following:
\begin{equation}
\label{eqn:convgraph}
    g_w \times x  = U\Sigma U^Tx \approx w(I - D^{-\frac{1}{2}}AD^{-\frac{1}{2}})x
\end{equation}

Here, $U$ is the matrix of the eigenvectors of the normalized graph Laplacian, $\Sigma$ represents a diagonal matrix of eigenvalues, and $I-D^{\frac{-1}{2}}AD^{\frac{-1}{2}}$ is the symmetric normalized form of the graph laplacian~\cite{van2011graphs}. The approximation term is derived by \citeauthorandyear{kipf2016semi} from \cite{hammond2011wavelets}. They generalize Equation \ref{eqn:convgraph} as the output of a convolution layer in GCN by setting this normalized form to $\hat{A}$. Their layer-wise propagation function becomes the following:
\begin{equation}
\label{eqn:convlayer}
    H^{l+1} = \phi(\hat{A}H^lW^l)
\end{equation}
Here, $H^l$ is the activation of $l^{th}$ hidden layer and $H^0 = X$, where $X$ is the input feature of dimension $n \times f$. The $\phi$ is a non-linear activation function in Equation \ref{eqn:convlayer} and a popular choice, also used in GCN, are Rectified Linear Units (ReLU)~\cite{nair2010rectified}. $\hat{A}$ is the adjacency matrix of dimension $n \times n$ and $W^l$ is the weight matrix of the $l^{th}$ layer having dimension $f \times d$. Thus, our final output embedding $\mathcal{Z}$ comes from $H^{l+1}$ whose dimension is $n \times d$. Sometimes, a bias term is added, but to keep it simple we skip any bias in Equation \ref{eqn:convlayer}. In GCN, authors in particular show results varying across several layers in the network. They define a softmax activation function in the output layer, compute cross-entropy loss, and update weights using a batch gradient descent approach. In the semi-supervised settings, authors train their model using a subset of vertices of the graph and then test performance based on the remaining subset of vertices. 

\begin{figure}[!ht]
    \centering
    \includegraphics[width=0.9\linewidth]{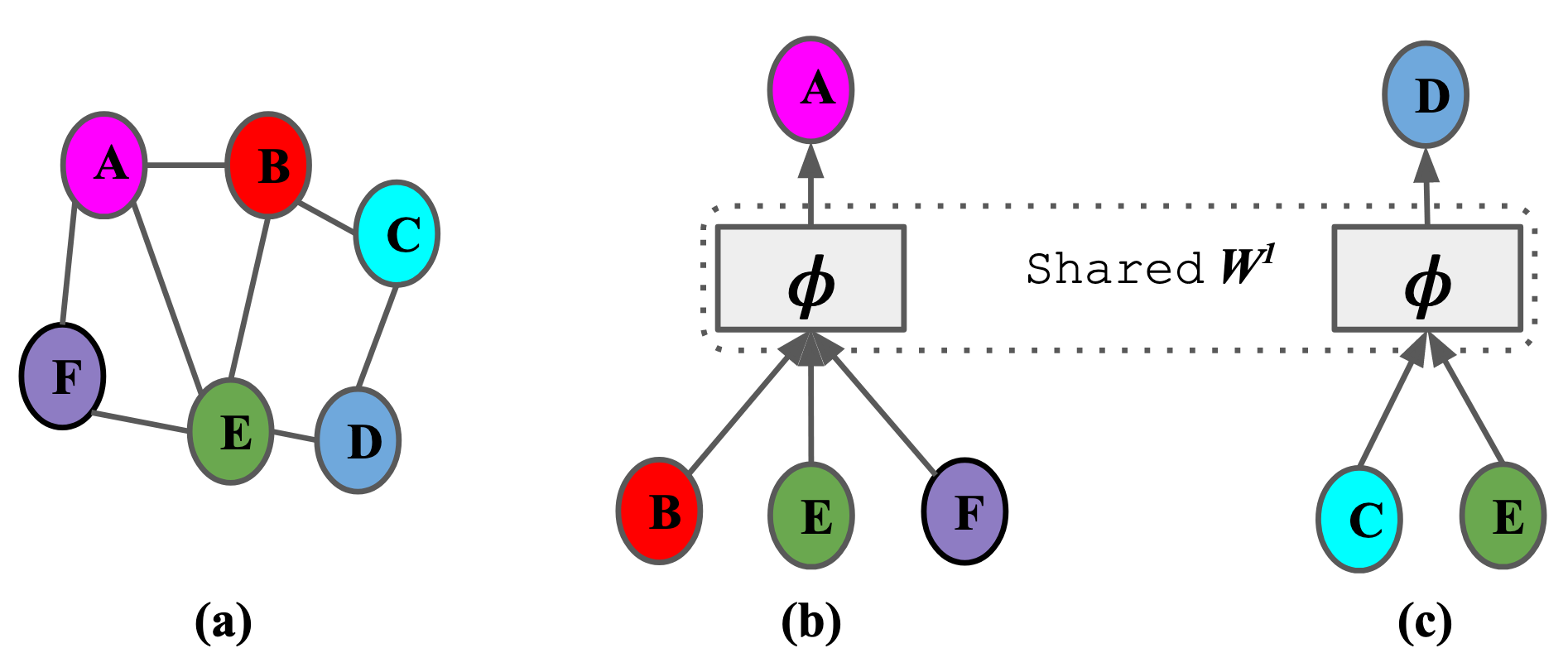}
    \caption{(a) An example graph. (b) Contributions from the neighbors of vertex $A$ are being aggregated towards it through an activation function. (c) Contributions from the neighbors of vertex $D$ are being aggregated towards it through an activation function. Note that the weight matrix $W^l$ is shared in the $1^{th}$ layer of the graph convolutional networks.}
    \label{fig:gcnexample}
\end{figure}

In Equation \ref{eqn:convlayer}, notice that when we multiply $H$ by $W$ and then by $\hat{A}$, we sum-up contributions from neighbors. In Fig. \ref{fig:gcnexample}, we summarize this procedure. When we compute the activation of a layer by Equation \ref{eqn:convlayer}, we take contributions from neighbors e.g., activation of vertex $A$ takes contributions from its neighbors $B$, $E$, and $F$ (see Fig. \ref{fig:gcnexample}(b)). Similarly, the activation of vertex $D$ considers contributions from its neighbors $C$, and $E$ (see Fig. \ref{fig:gcnexample}(c)). Note that the weight of a layer $W$ is shared among all vertices while computing their activations. 

\graphsage{} is an extension of GCN that carefully explores some neighborhood aggregation procedures~\cite{hamilton2017inductive}. Authors of \graphsage{} represent the convolutional mechanism in the following way:
\begin{equation}
\label{eqn:convgraphsage}
\begin{split}
H^{l+1}_{N(u)} & = AGGREGATE_{l+1}(\{H^l_{v},\forall v\in N(u)\}) \\
H^{l+1}_u & = \phi\big(W^l.CONCAT(H^{l}_u, H^{l+1}_{N(u))}\big)
\end{split}
\end{equation}

\begin{figure}[!ht]
    \centering
    \includegraphics[width=0.9\linewidth]{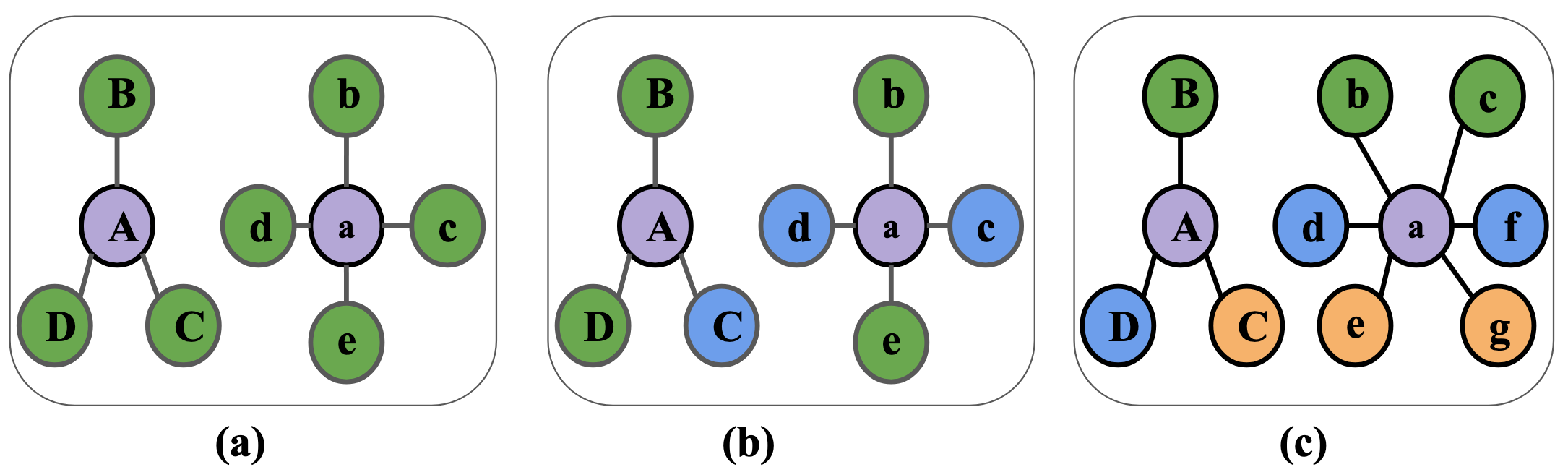}
    \caption{Examples where aggregation function fails to distinguish graph structures (same color indicates same contribution). (a) Mean and Max-pooling aggregation fail. (b) Max-pooling aggregation fails. (c) Mean and Max-pooling aggregation fail.}
    \label{fig:gintest}
\end{figure}

\citeauthorandyear{hamilton2017inductive} use three aggregation functions: (1) Mean aggregator - it computes contributions from all neighbors of a node and then takes the arithmetic mean (2) Max-pool aggregator - it takes the maximum of all contributions from neighbors, and (3) Long Short-Term Memory (LSTM) - it is a variant of vanilla Recurrent Neural Network (RNN) that is normally useful for sequential or time-series data. Mean and Max-pool aggregators are order-invariant whereas LSTM is order dependent~\cite{hochreiter1997long}. In \graphsage{}, authors show superior performance over previous methods in both \emph{transductive} and \emph{inductive} settings. Graph Attention Networks \emph{GAT}~\cite{velivckovic2017graph} is another popular work that shows better performance over previous methods. It takes weighted contributions from neighbors in an aggregation function. Recent work by \cite{xu2018powerful} generalizes characteristics of graph neural network and proposes a Graph Isomorphism Network (GIN) that is as powerful as the Weisfeiler-Lehman approach \cite{shervashidze2011weisfeiler} for graph isomorphism testing. Authors claim that an aggregation function should be \emph{injective} and must not map two different neighborhoods to the same representation. They propose a Multi-Layer Perceptrons (MLP) based convolution mechanism and summation as an aggregation function. Notably, MLP is a universal approximator \cite{hornik1989multilayer} and thus, decoupled from the summation-based aggregation function, it can capture different graph structures very well. Authors also show the limitations of several aggregation functions used in \graphsage{} which can be summarized by Fig. \ref{fig:gintest} (the same color represents the same contribution). Between two graphs in Fig. \ref{fig:gintest} (a), vertices $A$ and $a$, both get the same contributions from neighbors using Mean or Max-pooling aggregation functions, though graphs have different structures. Similarly, Max-pooling aggregation fails for the case of Fig. \ref{fig:gintest} (b) as both $A$ and $a$ will get the same maximum contribution from one of its neighbors. For the same reason, both Mean and Max-pooling fail for the scenario shown in Fig. \ref{fig:gintest} (c). They suggest that summation is an \emph{injective} aggregation function that can distinguish such structures among different graphs.

Earlier methods of graph convolution networks run more slowly and consume a significant amount of memory even for a very small graph. Thus, researchers have made efforts to improve runtime for training to effectively operate on large graphs. Of course, there does not exist any such method that gains maximum performance, consumes minimum memory, and runs in optimal time. There is always a trade-off between sacrificing one option and gaining better performance in another option. \emph{Fast-GCN}~\cite{chen2018fastgcn} is such a method that sacrifices performance in terms of accuracy measures but runs significantly faster than previous methods. Sometimes, contributions from neighbors take a long time and consume more memory, in the case of densely connected or scale-free graphs. Thus, traditional batch processing does not help. They assume that the vertices of a graph have an independent and identical sampling distribution and evaluate the activation of a layer through a Monte Carlo approximation \cite{karp1989monte}. Their batch update procedure runs faster and shows comparable performance. It is common practice to shuffle the mini-batch while training a model using SGD, as it leads to quicker convergence. However, random shuffling of the mini-batch does not help while working with graphs if the sequential ordering of vertices is already grouped based on some properties. Thus, it is a good idea to cluster the graph before mini-batch training. \emph{Cluster-GCN}~\cite{chiang2019cluster} is such an approach that clusters the graph first and then takes a subset of vertices as a mini-batch. It intentionally removes all inter-cluster edges from the graph and can train each batch independently. Authors of \emph{Cluster-GCN} show that their approach converges faster than other methods and can train very large graphs within a reasonable time. \citeauthorandyear{zeng2019accurate} propose a similar technique where they sample a subgraph from $G$ using the Frontier Sampling approach \cite{ribeiro2010estimating} and then use the subgraph as a mini-batch for training. Authors empirically show that their highly optimized parallel implementation achieves a significant speed-up over \graphsage{} without sacrificing accuracy.

The embedding generation of a graph is basically treated as an encoding scheme to reduce the dimensions to vector space. Thus, researchers also apply an autoencoder~\cite{hinton2006reducing} based technique to solve this particular problem in this area. There are also some other methods that are based on Recurrent Neural Networks (RNN)~\cite{rumelhart1986learning} or its variants. We mention some of these methods in the following Section \ref{sec:othertools}.

\begin{table*}[!htb]
\centering
\begin{tabular}{|l|c|c|c|l|}
\hline
\textbf{Tool Name} & \textbf{Code} & \textbf{Venue} & \textbf{Runtime}                          & \multicolumn{1}{c|}{\textbf{Link}}                \\ \hline
\textcolor{red}{ReFeX}~\cite{henderson2011s}             & Python          & KDD'11            &                  $dm+d^2n$                         & \url{https://github.com/randomsurfer/refex}             \\ \hline
\textcolor{red}{RolX}~\cite{henderson2012rolx}               & Python          & KDD'12            &                 $dm+drn$                          & \url{https://github.com/benedekrozemberczki/RolX }      \\ \hline
\textcolor{violet}{GF}~\cite{ahmed2013distributed}                 & Python          & WWW'13            &               $dm$                            & \url{https://github.com/thunlp/OpenNE   }               \\ \hline
\textcolor{violet}{GraRep}~\cite{cao2015grarep}             & MATLAB          & CIKM'15           &                  $dn^3$                         & \url{https://github.com/ShelsonCao/GraRep}     \\ \hline
\textcolor{violet}{HOPE}~\cite{ou2016asymmetric}               & MATLAB          & KDD'16            &                $d^2m$                           & \url{https://github.com/ZW-ZHANG/HOPE}                  \\ \hline
\textcolor{violet}{GRAPHWAVE}~\cite{donnat2018learning}          & Python          & KDD'18            &               $pdm$                            & \url{https://github.com/snap-stanford/graphwave  }      \\ \hline
\textcolor{orange}{DeepWalk}~\cite{perozzi2014deepwalk}           & Python          & KDD'14            & $cdn\log n$                                    & \url{https://github.com/phanein/deepwalk }              \\ \hline
\textcolor{orange}{LINE}~\cite{tang2015line}              & C++          & WWW'15            & $dsn$                                       & \url{https://github.com/tangjianpku/LINE}               \\ \hline
\textcolor{orange}{node2vec}~\cite{grover2016node2vec}           & Python          & KDD'16            & $cdsn $                                      & \url{https://github.com/aditya-grover/node2vec }        \\ \hline
\textcolor{orange}{struc2vec}~\cite{ribeiro2017struc2vec}          & Python          & KDD'17            &             $ldn^3$                              & \url{https://github.com/leoribeiro/struc2vec }          \\ \hline
\textcolor{orange}{metapath2vec}~\cite{dong2017metapath2vec}          & Python          & KDD'17            &             $cdsn $                               & \url{https://github.com/apple2373/metapath2vec}          \\ \hline
\textcolor{orange}{HARP}~\cite{chen2018harp}               & Python          & AAAI'18           &                $cdn\log n$                           & \url{https://github.com/GTmac/HARP}                     \\ \hline
\textcolor{orange}{VERSE}~\cite{tsitsulin2018verse}              & C++          & WWW'18            & $dsn $                                      & \url{https://github.com/xgfs/verse}                     \\ \hline
\textcolor{orange}{SINE}~\cite{zhang2018sine}               & C          & ICDM'18           &                     $dsm$                      & \url{https://github.com/daokunzhang/SINE}      \\ \hline
\textcolor{orange}{GEMSEC}~\cite{rozemberczki2019gemsec}            & Python          & ASONAM'19         &             $cdsn $                              & \url{https://github.com/benedekrozemberczki/GEMSEC}     \\
\hline
\textcolor{orange}{Force2Vec}~\cite{rahman2020force2vec}            & C/C++          & ICDM'20         &             $dsn $                              & \url{https://github.com/HipGraph/Force2Vec} 
\\
\hline
\textcolor{blue}{GCN}~\cite{kipf2016semi}                & Python          & ICLR'17           & $ldm $                                      & \url{https://github.com/tkipf/gcn }                     \\ \hline
\textcolor{blue}{GraphSAGE}~\cite{hamilton2017inductive}          & Python          & NIPS'17           & $r^ld^2n$ & \url{https://github.com/williamleif/GraphSAGE }         \\ \hline
\textcolor{blue}{FastGCN}~\cite{chen2018fastgcn}           & Python          & ICLR'18           & $rld^2n$                  & \url{https://github.com/matenure/FastGCN}               \\ \hline
\textcolor{blue}{GAT}~\cite{velivckovic2017graph}                & Python          & ICLR'18           & $ldm$                                      & \url{https://github.com/PetarV-/GAT}                    \\ \hline
\textcolor{blue}{Cluster-GCN}~\cite{chiang2019cluster}       & Python          & KDD'19            & $ldm$                                       & \url{https://github.com/zhengjingwei/cluster\_GCN} \\ \hline
\textcolor{blue}{GIN}~\cite{xu2018powerful}                & Python          & ICLR'19           & $ldm $                                      & \url{https://github.com/weihua916/powerful-gnns}        \\ \hline
\textcolor{blue}{GraphSAINT}~\cite{zeng2020graphsaint}                & Python          & ICLR'20           & $ldm $                                      & \url{https://github.com/GraphSAINT/GraphSAINT}        \\ \hline
\end{tabular}
\caption{A list of popular graph representation learning methods are tabulated with corresponding programming languages, their publication years and venues, theoretical running time and publicly available repositories (\textcolor{red}{red}, \textcolor{violet}{violet}, \textcolor{orange}{orange} and \textcolor{blue}{blue} colors represent feature engineering, matrix factorization, shallow neural network and graph neural networks based methods, respectively). $n$ - number of vertices, $m$ - number of edges, $r$ - number of roles, $d$ - embedding dimension or size of representation, $s$ - number of negative samples, $p$ - order of Chebyshev polynomial approximation, $c$ - a constant factor which is a product of number of walks, walk length, and window size for Skip-gram model, $l$ - number of layer in graph convolutional networks, $r$ - number of sampled neighbors per vertex.}
\label{tab:listoftools}
\end{table*}

\subsection{Other Methods}
\label{sec:othertools}
There are some social networks where each edge may have either a positive weight or negative weight, indicating trust or distrust between two users in the network. General-purpose network embedding methods can not solve this problem efficiently. Thus, efforts have been made to analyze signed networks \cite{wang2017attributed,islam2017distributed}. \emph{Regular equivalence} is a relaxation of \emph{structural equivalence} where the embeddings of two vertices will be similar if they have neighbors which are also similar. \citeauthorandyear{tu2018deep} propose a Deep Recursive Network Embedding (\emph{DRNE}) method which uses LSTM to solve this problem and predicts different centrality measures\footnote{\url{https://en.wikipedia.org/wiki/Centrality}} very well. Some other methods also use variants of RNN for generating graph embeddings \cite{yu2018learning} or artificial graph generation~\cite{you2018graphrnn}. Apart from this, autoencoder based methods have been applied to graph embedding~\cite{kipf2016variational,pan2018adversarially} or molecular graph (such as SMILES\footnote{``The Simplified Molecular-Input Line-Entry System (SMILES) is a specification in the form of a line notation for describing the structure of chemical species using short ASCII strings". (Wikipedia)}) reconstruction problems~\cite{simonovsky2018graphvae,kusner2017grammar,dai2018syntax}. We see few methods in the literature that introduce a common framework to generate an embedding for large graphs using a multi-level approach where other unsupervised methods such as \deepwalk{} or \linet{} can be used for underlying computations~\cite{liang2018mile,chen2018harp}. There are some other methods that are developed for a specific purpose or to solve a specific problem. For example, \emph{edge2vec}\footnote{\url{https://github.com/RoyZhengGao/edge2vec}} has been proposed for three biomedical prediction tasks \cite{gao2018edge2vec}, namely, (i) biomedical entity classification, (ii) compound-gene bioactivity prediction, and (iii) biomedical information retrieval. Authors empirically show that \emph{edge2vec} outperforms general-purpose graph embedding methods for their heterogeneous biomedical dataset. \emph{OhmNet} has been proposed for multi-cellular function prediction \cite{zitnik2017predicting} that uses \nodetwovec{} as an underlying model.
\emph{BioENV}\footnote{\url{https://github.com/xiangyue9607/BioNEV}} has been developed recently by combining several general-purpose existing methods to solve biomedical link prediction task~\cite{yue2019graph}. Another recent method is \emph{MeSHHeading2vec} that converts Medical Subject Headings (MeSH) tree structure into a relationship network and applies five existing graph embedding methods to perform several downstream analyses tasks \cite{guo2020meshheading2vec}. We provide a shortlist of popular general-purpose methods for graph embedding in Table \ref{tab:listoftools}.

\section{Experiment Setup}
\label{sec:experimentalcomp}
\subsection{Experimental Coverage}
The primary objective of this survey is to compare and contrast the performance of unsupervised and semi-supervised graph embedding methods via experiments with a diverse collection of graphs.
We aim to compare the running time, memory utilization, and scalability of graph embedding algorithms. 
We also compare the performance of these algorithms in node classification, link prediction, clustering, and visualization tasks.  
Additionally, we survey the sensitivity of various methods with respect to their parameters. 
Table~\ref{tab:exp_coverage} shows the experimental converge considered in this survey. 
By considering aspect of systems, performance with ML tasks, and parameter sensitivity, this survey will help users select methods based on the requirements of their respective applications. 

\begin{table*}[!htb]
    \centering
    \begin{tabular}{l|l}
    \hline
    {\bf Experiment type} & {\bf Experiment details}\\ 
    \hline
     Time and memory    &  running time (CPU and GPU), memory requirement\\
     Scalability    &  scaling with processors, scaling with graphs\\
         \hline
        Classification & node classification (multiclass and multilabel), link prediction \\
        Interpretation & clustering, visualization \\
        \hline
    Algorithm-specific parameters    & number of layers, random walk length \\
    Algorithm-independent parameters    & embedding dimension, sampling, convergence\\
    \hline
    \end{tabular}
    \caption{The experimental coverage in this survey. Horizontal lines separate different aspects of experiments, from top to bottom: complexity and scalability, performance in downstream tasks, and parameter sensitivity. }
    \label{tab:exp_coverage}
\end{table*}



\begin{table*}[!htb]
\centering
\begin{tabular}{c|c|c|c|c}
\hline
\textbf{Graphs} & \textbf{Vertices} & \textbf{Edges} & \textbf{\#Labels} & \textbf{Avg. Degree} \\ \hline
Cora            & 2,708              & 5,429           & 7                 & 3.89                 \\ 
Citeseer        & 3,327              & 4,732           & 6                 & 2.736                \\ 

Pubmed  &   19,717	&   44,338	&   3	&   4.49  \\ 
Flickr          & 89,250	    &   899,756	&   7	&   20.16              \\ \hline
BlogCatalog	&   10312	&   333983	&   39	&   64.77 \\ 
Youtube         & 1,138,499           & 2,990,443        & 47                & 5.253                \\ \hline
\end{tabular}
\caption{Datasets used for different experiments. Graphs are available at \url{https://linqs.soe.ucsc.edu/data} and \url{https://sparse.tamu.edu/} [Last accessed: Dec. 8, 2021].}
\label{tab:dataset}
\end{table*}

\begin{table*}[!htb]
\centering

\begin{tabular}{c|c|c|c|c|c|}
\cline{2-6}
\textbf{}                                          & \textbf{Property} & \textbf{\begin{tabular}[c]{@{}c@{}}Intel\\ Skylake  8160\end{tabular}}   & \textbf{\begin{tabular}[c]{@{}c@{}}AMD\\ EPYC 7551\end{tabular}} & \textbf{\begin{tabular}[c]{@{}c@{}}ARM \\ ThunderX CN8890\end{tabular}} & \textbf{\begin{tabular}[c]{@{}c@{}}Intel Skylake 7900X \\ + Nvidia Titan RTX\end{tabular}} \\ \hline
\multicolumn{1}{|c|}{\multirow{4}{*}{\parbox[t]{2mm}{\multirow{1}{*}{\rotatebox[origin=c]{90}{Core}}}}}        & Clock             & 2.10 GHz                                                           & 2 GHz                                                            & 1.9 GHz                                                         & 3.3 GHz (1.35 GHz)\\[0.18ex] 
\multicolumn{1}{|c|}{}                             & L1 cache          & 32 KB   &                              32 KB                        & 32 KB                                                    &       32 KB   \\[0.18ex] 
\multicolumn{1}{|c|}{}                             & L2 cache          & 1 MB   &                              512 KB                        & $\times$ & 1 MB (6 MB) \\[0.18ex] 
\multicolumn{1}{|c|}{}                             & LLC          & 32 MB                                                                & 8 MB                                                            & 16 MB                                                                & 13.75 MB \\[0.18ex] \hline
\multicolumn{1}{|c|}{\multirow{3}{*}{\parbox[t]{2mm}{\multirow{1}{*}{\rotatebox[origin=c]{90}{\vspace{-2cm} Node}}}}} & Sockets           & 2                                                                  & 2                                                                & 1                                                     &         1 \\ [0.18ex]
\multicolumn{1}{|c|}{}                             & Cores/soc.        & 24                                                                 & 32                                                    & 48                                      &            10 (576 TC)            \\ [0.18ex]
\multicolumn{1}{|c|}{}                             & Memory            & 256 GB                                                              & 128 GB                                                            & 64 GB                                                            & 64 GB (24 GB)\\ \hline
\multicolumn{1}{|c|}{\multirow{2}{*}{\parbox[t]{2mm}{\multirow{1}{*}{\rotatebox[origin=c]{90}{Env.}}}}}  & Compiler          & gcc 10.1.0                                                         & gcc 5.4.0                                                        & gcc 7.5.0                                                       & gcc 7.5.0 (CUDA 11.1) \\ 
\multicolumn{1}{|c|}{}                             & Flags             & \begin{tabular}[c]{@{}c@{}}O3, mavx512f, \\ mavx512dq\end{tabular} & \begin{tabular}[c]{@{}c@{}}O3, mavx,\\ mfma\end{tabular}         & \begin{tabular}[c]{@{}c@{}}O3, asimd,\\ armv8-a\end{tabular}        & \begin{tabular}[c]{@{}c@{}}O3, mavx512f, \\ mavx512dq\end{tabular}\\ \hline
\end{tabular}
\caption{Hardware configurations of different CPU and GPU architecture for our experiments. Numbers in the parenthesis represent GPU specification. Titan RTX has 4608 CUDA cores (single precision) and 576 Tensor Cores (TC).}
\label{tab:hardware}
\end{table*}

\subsection{Computing Platform Coverage}
The choice of a computing platform does not influence the performance of algorithms in various ML tasks nor it impacts their  parameter sensitivity.
Hence, for these two classes of experiments, we used an Intel CPU with 48 cores and 256GB memory.
For running time and scalability experiments, we used state-of-the-art Intel, AMD, and ARM CPUs and an NVIDIA GPU. 
Our experimental results on different platforms will help users understand the expected performance of different methods on different CPUs and GPUs.
We briefly provide the architecture details of different server machines in Table \ref{tab:hardware}.



\subsection{Coverage of Methods and Setup}
\label{sec:expsetup}
Over the last few years, researchers have developed hundreds of new methods for graph representation learning. 
A representative subset of available methods is shows in Table \ref{tab:listoftools} where we selected methods from different categories (Fig.~\ref{fig:taxonomy}).
For each method, we report its computational complexity and a link to the publicly available software. 
For comparative experiments, we select a subset of methods based on their popularity and citations. More specifically, we choose RolX from the feature engineering based methods, HOPE from the matrix factorization based methods, a set of methods $\{$DeepWalk, LINE, struc2vec, VERSE, Force2Vec$\}$ from shallow embedding methods, and a set of methods $\{$GCN, GraphSAGE, FastGCN, GAT, Cluster-GCN, GraphSAINT$\}$ from the GNN methods.

We briefly discuss each of these methods below with their hyper-parameter set-up. We use default values for other parameters when not explicitly mentioned. 


\begin{itemize}
     \item \textbf{\rolx{}\footnote{\url{https://github.com/benedekrozemberczki/RolX}}~\cite{henderson2012rolx}:} We set all parameters to the same as described in the paper. Specifically, we set 5 as the number of binarization bins, 32 as the batch size, 0.9 as the pruning cut-off, 128 as the dimension of the output embedding, and 3 as the depth of recursion. 
     
    \item \textbf{\hope{}\footnote{\url{https://github.com/palash1992/GEM}}~\cite{goyal2018graph}:} The original version is implemented in MATLAB. However, a more generalized and Python implementation is available in the GEM \cite{goyal2018graph} package. We set $\beta$ to 0.01, as shown in Equation \ref{eqn:hope}. We also set the output embedding dimension to 128 to match with other methods. We explicitly mention whenever we use different parameters other than these. 
    
    \item \textbf{\deepwalk{}\footnote{\url{https://github.com/phanein/deepwalk}}~\cite{perozzi2014deepwalk}:} This is a pioneering work for the graph embedding problem which generates random walks for each vertex in the graph and then formulates the problem in a similar way as \emph{word2vec}. In fact, \deepwalk{} uses the \emph{word2vec} framework directly in its programming model. Recently, authors have updated their repository with a new version of \emph{word2vec}. To conduct the experiment, we set several of its hyper-parameters as follows: walk length to $80$, number of walks per vertex to $10$, size of dimension to $128$, and number of workers to $48$. 
    
    \item \textbf{\linet{}\footnote{\url{https://github.com/tangjianpku/LINE}}~\cite{tang2015line}:} In this work, authors use a KL-divergence based optimization function where \emph{first-order} and \emph{second-order} proximity of vertices are considered. They assume that neighbors of neighbors can capture a similar kind of inherent information. This method works for directed/undirected graphs as well as for weighted graphs. We set several of its hyper-parameters as follows: size of dimension $=128$, order of proximity $=2$, number of negative samples $=5$, total number of training samples $=10000$ millions, starting value of learning rate $=0.025$ and number of threads $=48$.
    
     \item \textbf{\structtwovec{}\footnote{\url{https://github.com/leoribeiro/struc2vec}}~\cite{ribeiro2017struc2vec}:} This method generates an embedding based on the structural equivalence of the graph. It uses a hierarchy to measure vertex similarity and constructs a multilayer graph. In the following step, it generates random walks from the multilayer graph and generates an embedding using a similar optimization model to \deepwalk{}. To conduct experiments using \structtwovec{}, we set the number of walks to 20, walk length to 80, window size to 5, size of embedding dimension to 128, layers to 6 and use all options available for running time optimization.
     
    \item \textbf{\verset{}\footnote{\url{https://github.com/xgfs/verse}}~\cite{tsitsulin2018verse}:} This method embeds graphs based on several similarity measures of vertices. The full version of \verset{} has higher running time and memory consumption. Thus, authors also provide a negative sampling based technique which has comparatively lower running time and memory consumption. To run \verset{}, we set several of its hyper-parameters as follows: size of dimension $=128$, number of negative samples $=5$ and number of threads $=48$. 
    
    \item \textbf{Force2Vec\footnote{\url{https://github.com/HipGraph/Force2Vec}}~\cite{rahman2020force2vec,rahman2022force2vecj}:} This method embeds graphs force-directed graph layout method. This method use negative sampling and effective parallelization techniques that significantly boost-up the runtime performance. There are three versions of the algorithms where either sigmoid or $t$-distribution can be used as a similarity function. To run Force2Vec, we set several of its hyper-parameters as follows: size of dimension $=128$, number of negative samples $=5$ and number of threads $=48$.
    
    \item \textbf{\emph{GCN}\footnote{\url{https://github.com/tkipf/gcn}}~\cite{kipf2016semi}}, \textbf{\graphsage{}\footnote{\url{https://github.com/williamleif/graphsage-simple}}~\cite{hamilton2017inductive} and \textbf{\emph{FastGCN}\footnote{\url{https://github.com/matenure/FastGCN}}~\cite{chen2018fastgcn}}:} For all experimental datasets, we use 1000 vertices for testing, 500 vertices for validation and the rest of the vertices for training. To increase the usability of \graphsage{}, authors implemented this method in Tensorflow and PyTorch. For all of our experiments, we report results based on the PyTorch implementation of \graphsage{} using the \emph{Mean} aggregation function.
    
    \item \textbf{\emph{GAT}~\cite{velivckovic2017graph}}, \textbf{ClusterGCN~\cite{chiang2019cluster}} and \textbf{\emph{GraphSAINT}~\cite{zeng2020graphsaint}}: For all experimental datasets, we use 1000 vertices for testing, 500 vertices for validation and the rest of the vertices for training. We use the implementations available in PyG framework.
\end{itemize}

\subsection{Datasets Coverage}
To conduct experiments, we choose both homogeneous and heterogeneous benchmark networks that are commonly used to assess the performance of state-of-the-art methods. We briefly discuss the properties of the datasets as follows.
\subsubsection{Homogeneous Networks}
We define a network to be homogeneous when each node and each edge is assigned to a single label. Thus, this type of networks contains only structural information which is more local to a vertex or an edge. In the following, we describe three such homogeneous networks which have been widely used for benchmarking in the literature \cite{lu2003link,velivckovic2017graph,klicperadiffusion2019,namata2012query}.

\begin{itemize}
    

    \item \textbf{Cora:} This dataset consists of 2708 machine learning papers which are classified into one of following seven topic classes: Case Based, Genetic Algorithms, Neural Networks, Probabilistic Methods, Reinforcement Learning, Rule Learning, and Theory. This set of papers is selected in such a way that they have at least one one-way mutual citation. There are 5429 such citations and we term these as edges in the graph. 
    
    \item \textbf{Citeseer:} This dataset consists of 3312 scientific papers that are classified into one of the following six classes: Agents, Artificial Intelligence, Database, Human Computer Interaction, Machine Learning, and Information Retrieval. This set of papers is selected in a similar way to Cora. After processing the dataset, there are 4732 citations which we term as edges in the network. 
    
    \item \textbf{Pubmed:} This is another widely used diabetes dataset which has 19717 scientific papers classified into one of three diabetes type: Diabetes Mellitus Experimental, Diabetes Mellitus Type 1, Diagebes Mellitus Type 2. There are 44338 edges in the graph which represent links or citations between a pair of papers. We process this dataset to make a consistent format with the above two networks.
    
\end{itemize}

Software implementations of different methods often process graphs in an edgelist, CSV \cite{epasto2019single}, or CSR \cite{tsitsulin2018verse} format.
We preprocess the aforementioned networks in different formats required by various software.  
As each vertex in a homogeneous network has one label, the classification problem solved on such networks is a multi-class classification problem. All these networks are available on the UCSC\footnote{\url{https://linqs.soe.ucsc.edu/data}} website.

\subsubsection{Heterogeneous Networks}
We define a network to be heterogeneous when its vertices can have more than one label. The vertices are likely to contain global information about the network. Predictions in such network are harder than in homogeneous networks. According to \citeauthorandyear{cai2018comprehensive}, the main sources of such networks are community based questions answering sites \cite{fang2016community}, multimedia networks \cite{chang2015heterogeneous,zhang2016learning} and knowledge graphs \cite{bollacker2008freebase}. In the following, we describe two such heterogeneous networks which have been used for benchmarking performance in the literature \cite{perozzi2014deepwalk,grover2016node2vec,tang2009relational}.

\begin{itemize}
    \item \textbf{BlogCatalog:} This dataset contains bloggers of the BlogCatalog\footnote{https://www.blogcatalog.com/} website as vertices and friendship between two bloggers as edges. Each blogger submits a blog to the website and mentions some metadata about the blog. Based on this, bloggers can label their blogs in a prespecified set of categories i.e., each blogger can choose multiple categories. This dataset has 10312 vertices, 333983 edges and 39 categories. According to \citeauthorandyear{tang2009relational}, each blogger includes his/her own blog under 1.6 categories on average.
    
    \item \textbf{Youtube:} This is a social network dataset curated from the popular video sharing website Youtube\footnote{\url{https://www.youtube.com/}}. Each vertex in the network represents a user and each edge represents friendship between two users. According to \citeauthorandyear{perozzi2014deepwalk}, the labels of vertices indicate groups of viewers who enjoy common type of videos. This is a relatively large dataset which can be used for scalability purpose. It has 1138499 vertices, 2990443 edges and 47 labels.
\end{itemize}

We also preprocess these datasets so that we can run all methods in their respective input formats. 
As each vertex in a heterogeneous network can have more than one label, the classification problem solved on such networks is a multi-label classification problem. All these datasets are available on the ASU\footnote{\url{http://socialcomputing.asu.edu/pages/datasets}} website. Besides these, we have also used three air-traffic datasets (namely, brazilian air-traffic, european air-traffic and US air-traffice) introduced in \structtwovec{} which are mainly used for structural equivalence discovery. Properties of these datasets are given in Table \ref{tab:structuraldata}.
\begin{table}[!h]
    \centering
    \begin{tabular}{c|c|c|c}
        Air-traffic & \# vertices & \# edges & Diameter \\ \hline
        Brazilian & 131 & 1038 & 5 \\
        European & 399 & 5995 & 5 \\
        USA & 1190 & 13599 & 8 \\ \hline
    \end{tabular}
    \caption{Air-traffic datasets of three regional airports which are mostly used for structural identity.}
    \label{tab:structuraldata}
\end{table}

\subsection{Performance Metrics}
A key focus of this survey is to understand the practical performance such as runtime, memory utilization, and scalability of various graph embedding and GNN methods on different graphs and hardware platforms.  
We also measure the accuracy of different methods when performing various machine learning tasks.
We use $F1$-micro and $F1$-macro scores for the multi-label classification task, accuracy for the binary prediction task, and the modularity score to compare the results of clustering. We give a brief description of these measures below:

\begin{itemize}
    \item \textbf{Accuracy:} When the predicted class is positive (negative) and the ground truth class is also positive (negative), it is termed as a true positive (negative). But when the predicted class is positive (negative) whereas ground truth class is negative (positive) then we term this as a false positive (negative). We represent an absolute number of true positive, false positive, true negative and false negative as TP, FP, TN and FN, respectively. In any type of binary classification or link prediction, we define accuracy as the following:
    \begin{equation}
        Accuracy = \frac{TP+TN}{TP+TN+FP+FN}
    \end{equation}

    \item \textbf{$F1$-micro:} For the multi-class classification task, the $F1$-micro score aggregates the contributions of all classes to calculate the average value of final $F1$ score. In terms of precision-recall we can define this for a set of classes $C$ as follows: 
    \begin{multline}
        P = \frac{\sum_{c\in C}TP_c}{\sum_{c\in C}(TP_c+FP_c)}, R=\frac{\sum_{c\in C}TP_c}{\sum_{c\in C}(TP_c+FN_c)},\\ F1\text{-micro}=\frac{2*P*R}{P+R}
    \end{multline}
    
    \item \textbf{$F1$-macro:} Unlike the $F1$-micro score, the $F1$-macro score will compute the score independently for each class and finally take their average $F1$ score. In terms of precision-recall, we can define this for a set of classes $C$ as follows: 
    \begin{multline}
        P = \frac{TP}{(TP+FP)}, R=\frac{TP}{(TP+FN)},\\ F1\text{-macro}=\frac{1}{|C|}\sum_{c\in C}\frac{2*P_c*R_c}{P_c+R_c}
    \end{multline}
    
    \item \textbf{Modularity:} The modularity score is a well-known measure to evaluate the effectiveness of any graph clustering technique. It computes the fraction of the edges that are within a given cluster minus the expected fraction, if edges are distributed randomly \cite{newman2004finding}. We can compute this score by using following Equation:
    \begin{equation}
    \label{eqn:modularity}
        \frac{1}{2m}\sum_{ij}\big[A_{ij} - \frac{k_ik_j}{2m}\big]\delta(c_i, c_j)
    \end{equation}
    Here, $A$ is the adjacency matrix of the graph, $m$ is the number of edges, $k_i$ is the degree of the vertex $v_i$. The membership of vertex $v_i$ belonging to a cluster is represented by $c_i$, and  $\delta(c_i, c_j) = 1$, if $i$ and $j$ are in the same cluster; otherwise, $\delta(c_i, c_j) = 0$.
    
\end{itemize}

\section{Runtime and Scalability}




We measure the average time of several runs under the same experimental setup for all methods in the corresponding server machine. Since no other programs run on those servers, we get almost consistent time for all runs. Thus, the standard deviation of runtimes is insignificant that has been ignored in the rest of our discussion. 

\subsection{Training Time for Unsupervised Embedding Algorithms}

\begin{table*}[!htb]
\centering
\begin{tabular}{|c|c|c|c|c|c|c|}
\hline
\multirow{2}{*}{\textbf{Methods}} & \multicolumn{6}{c|}{\textbf{Runtime in seconds}}                              \\ \cline{2-7} 
                                  & \textbf{Cora}     & \textbf{Citeseer} & \textbf{Blogcatalog} & \textbf{Pubmed}   & \textbf{Flickr7}   & \textbf{Youtube}   \\ \hline
RolX                              & {\cellcolor[rgb]{0.851,0.918,0.827}}1.10     & $\times$        & $\times$           & 25.74    & 349.84    & 53,877.42 \\ \hline
HOPE                              & 3.09     & 3.24     & 25.02       & 140.36   & $\times$         & $\times$         \\ \hline
DeepWalk                          & 73.95    & 87.24    & 332.30      & 591.48   & 1,989.57  & 28,049.53 \\ \hline
struc2vec                        & 274.52   & 367.05   & 1,237.27    & 2,199.16 & 8,176.75  & $\times$         \\ \hline
LINE                              & 1,913.29 & 1,889.01 & 11,495.60   & 1,897.50 & 10,259.28 & 12,843.44 \\ \hline
VERSE                             & 46.99    & 55.94    & 158.04      & 300.35   & 1,297.92  & 15,798.77 \\ \hline
HARP                              & 8.40     & 5.75     & 779.68           & 98.61    & 1,772.73  & 10,854.49 \\ \hline
Force2Vec                         & 1.43     & {\cellcolor[rgb]{0.851,0.918,0.827}}1.71     & {\cellcolor[rgb]{0.851,0.918,0.827}}9.81        & {\cellcolor[rgb]{0.851,0.918,0.827}}10.49    & {\cellcolor[rgb]{0.851,0.918,0.827}}53.99     & {\cellcolor[rgb]{0.851,0.918,0.827}}662.81    \\ \hline
\end{tabular}
\caption{Runtime (in seconds) of different unsupervised methods for different benchmark graphs on the Intel server. We run all methods using the same computing resources. For each graph, the fastest runtime is shown in a green shade. `$\times$' indicates that the method failed to run or it went out of available memory.}
\label{tab:runtimegraphsintel}
\end{table*}

To conduct experiments for runtime, we run each of the methods with their parameters discussed in Section \ref{sec:expsetup}. If a method has a parallel implementation, we use the maximum number of cores available in each processor. 
Table \ref{tab:runtimegraphsintel} reports the runtime (in seconds) of different graph-embedding algorithms on the Intel Skylake server  for both homogeneous and heterogeneous networks. 
We observe that the Force2Vec algorithm runs the fastest for most of the graphs. This runtime is expected for Force2Vec due to the simplicity in its underlying model, efficient shared memory-based parallelism, and effective memory utilization. 
One reason behind the superior performance of Force2Vec on the Intel processor is Force2Vec's ability to utilize Single Instruction Multiple Data (SIMD) vectorization units.
The RolX method shows better results for the small graph Cora; however, it shows relatively high runtime for large graphs. We also observe that popular DeepWalk and LINE methods 
may take several hours for some of our large-scale graphs. Their high running times are expected from their higher computational complexity as shown in Table~\ref{tab:listoftools}.
We can see in Table \ref{tab:runtimegraphsintel} that \structtwovec{} takes higher runtime than other methods due to its higher computational complexity. It creates a multi-layer graph from the original graph which increases its runtime significantly. We could not run this method on Intel server for the Youtube graph due to the higher memory consumption.

\begin{table*}[!htb]
\centering
\begin{tabular}{|c|c|c|c|c|c|c|} 
\hline
\multirow{2}{*}{\textbf{Methods}} & \multicolumn{6}{c|}{\textbf{Runtime in seconds}}                             \\ 
\cline{2-7}
                         & \textbf{Cora}     & \textbf{Citeseer} & \textbf{Blogcatalog} & \textbf{Pubmed}   & \textbf{Flickr7}   & \textbf{Youtube}  \\ 
\hline
HOPE                     & 7.21     & 11.12    & 165.19      & 985.43   & $\times$         & $\times$        \\ 
\hline
DeepWalk                 & 83.10    & 90.69    & 351.63      & 626.02   & 3,506.17  & $\times$        \\ 
\hline
struct2vec               & 891.08   & 1,045.26 & 2,817.38    & 8,589.17 & 32,169.06 & $\times$        \\ 
\hline
LINE                     & 1,437.14 & 1,295.16 & 1,621.78    & 1,327.12 & 2,300.31  & $\times$        \\ 
\hline
VERSE                    & 36.63    & 45.07    & {\cellcolor[rgb]{0.851,0.918,0.827}}146.15      & 265.04   & 1,676.94  & $\times$        \\ 
\hline
HARP                     & 26.25    & 27.27    & 892.32      & 128.60   & 1,295.83  & $\times$       \\ 
\hline
Force2Vec                & {\cellcolor[rgb]{0.851,0.918,0.827}}6.64     & {\cellcolor[rgb]{0.851,0.918,0.827}}6.87     & 253.52      & {\cellcolor[rgb]{0.851,0.918,0.827}}50.68    & {\cellcolor[rgb]{0.851,0.918,0.827}}455.93    & $\times$        \\
\hline
\end{tabular}
\caption{Runtime (in seconds) of different unsupervised methods for different benchmark graphs on the ARM server. All methods use the same computing resources. For each graph, the fastest runtime is shown in a green shade. `$\times$' indicates that the method failed to run because it went out of available memory.}
\label{tab:runtimegraphsarm}
\end{table*}

Table \ref{tab:runtimegraphsarm} reports the runtime of all methods on the ARM server. 
Since the ARM ThunderX processor has 64GB memory (as opposed to 256GB in the Intel server), all methods failed to process the Youtube graph on the ARM server.
The relative performance of different methods on the ARM processor is similar to that of the Intel processor (Table \ref{tab:runtimegraphsintel}), except for the Blogcatalog graph where VERSE runs faster than Force2Vec. 
The average degree of nodes in Blogcatalog is comparatively high where VERSE's sampling strategy has significant benefits, and Force2Vec does not support SIMD vectorization with load-balancing for the ARM processor. These two reasons may have contributed to the observed performance for Blogcatalog.


\begin{table*}[!htb]
\centering
\begin{tabular}{|c|c|c|c|c|c|c|} 
\hline
\multirow{2}{*}{\textbf{Methods}} & \multicolumn{6}{c|}{\textbf{Runtime in seconds}}                         \\ 
\cline{2-7}
                         & \textbf{Cora}   & \textbf{Citeseer} & \textbf{Blogcatalog} & \textbf{Pubmed}   & \textbf{Flickr7} & \textbf{Youtube}  \\ 
\hline
HOPE                     & {\cellcolor[rgb]{0.851,0.918,0.827}}3.50   & {\cellcolor[rgb]{0.851,0.918,0.827}}6.33     & {\cellcolor[rgb]{0.851,0.918,0.827}}59.49       & 314.40   & $\times$       & $\times$       \\ 
\hline
DeepWalk                 & 88.80  & 128.38   & 599.26      & 1,036.80 & $\times$       & 57,646.77       \\ 
\hline
struct2vec               & 276.99	&   342.49	&   2,652.36	&   3,891.30        & $\times$       & 1,534,844.69        \\ 
\hline
VERSE                    & 111.53 & 129.84   & 376.14      & 722.87   & $\times$       & 42,649.80        \\ 
\hline
HARP                     & 11.41  & 8.53     & 1,086.59    & 130.41   & $\times$       & 17,365.54        \\ 
\hline
Force2Vec                & 15.16  & 16.83    & 233.17      & {\cellcolor[rgb]{0.851,0.918,0.827}}149.39   & $\times$       & {\cellcolor[rgb]{0.851,0.918,0.827}}5,640.13       \\
\hline
\end{tabular}
\caption{Runtime (in seconds) of different unsupervised methods for different benchmark graphs on the AMD server.All methods use the same computing resources. For each graph, the fastest runtime is shown in a green shade. `$\times$' indicates that the method failed to run or it went out of available memory.}
\label{tab:runtimegraphsamd}
\end{table*}

In Table \ref{tab:runtimegraphsamd}, we report the runtime of different methods using the AMD Epyc server. 
When running smaller graphs such as Cora and Citeseer on the 64-core AMD processor, parallel methods such as Force2Vec and VERSE do not have enough work to keep all cores busy.
Consequently, the overheads of thread creation and scheduling overshadow the benefit of parallel executions, making HOPE the fastest method for small graph on the AMD processor. 
By contrast, VERSE and Force2Vec outperform other methods for large graphs such Pubmed and Youtube. 

{\bf Summary:} For small graphs, all graph embedding methods run in a reasonable time on all processors considered. 
However, highly parallel algorithms such as Force2Vec and multilevel methods such as HARP are needed to generate embeddings of large graphs such as Pubmed, Flickr7, and Youtube.
Computationally expensive methods such as RolX and struc2vec are impractical for large graphs (for example, struc2vec takes more than two weeks to generate the embedding for the Youtube graph on the AMD processor).

\subsection{Scalability of Unsupervised Embedding Algorithms}
\begin{figure*}[!ht]
    \centering
    \includegraphics[width=0.29\linewidth,height=3.5cm]{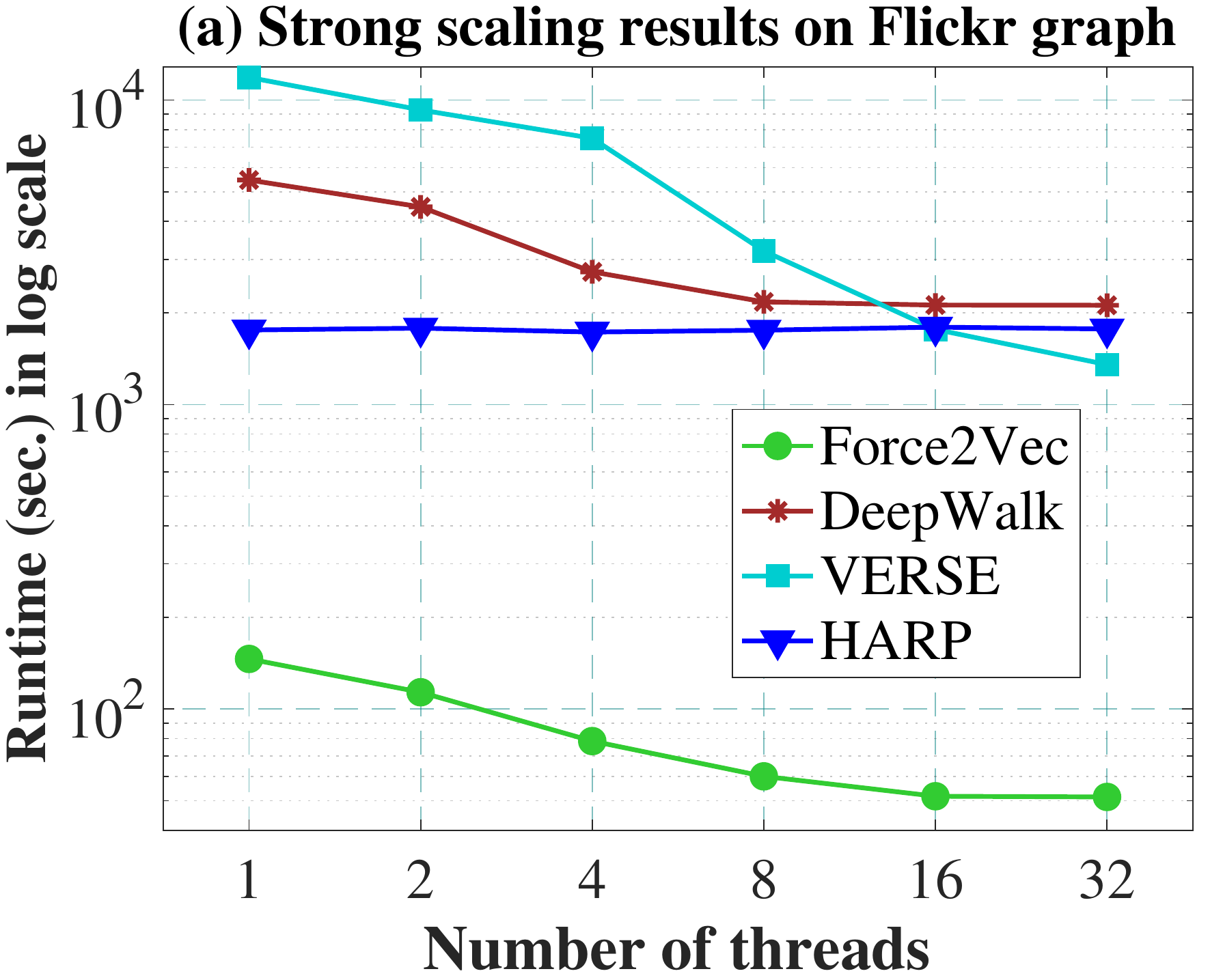}
    \includegraphics[width=0.29\linewidth,height=3.5cm]{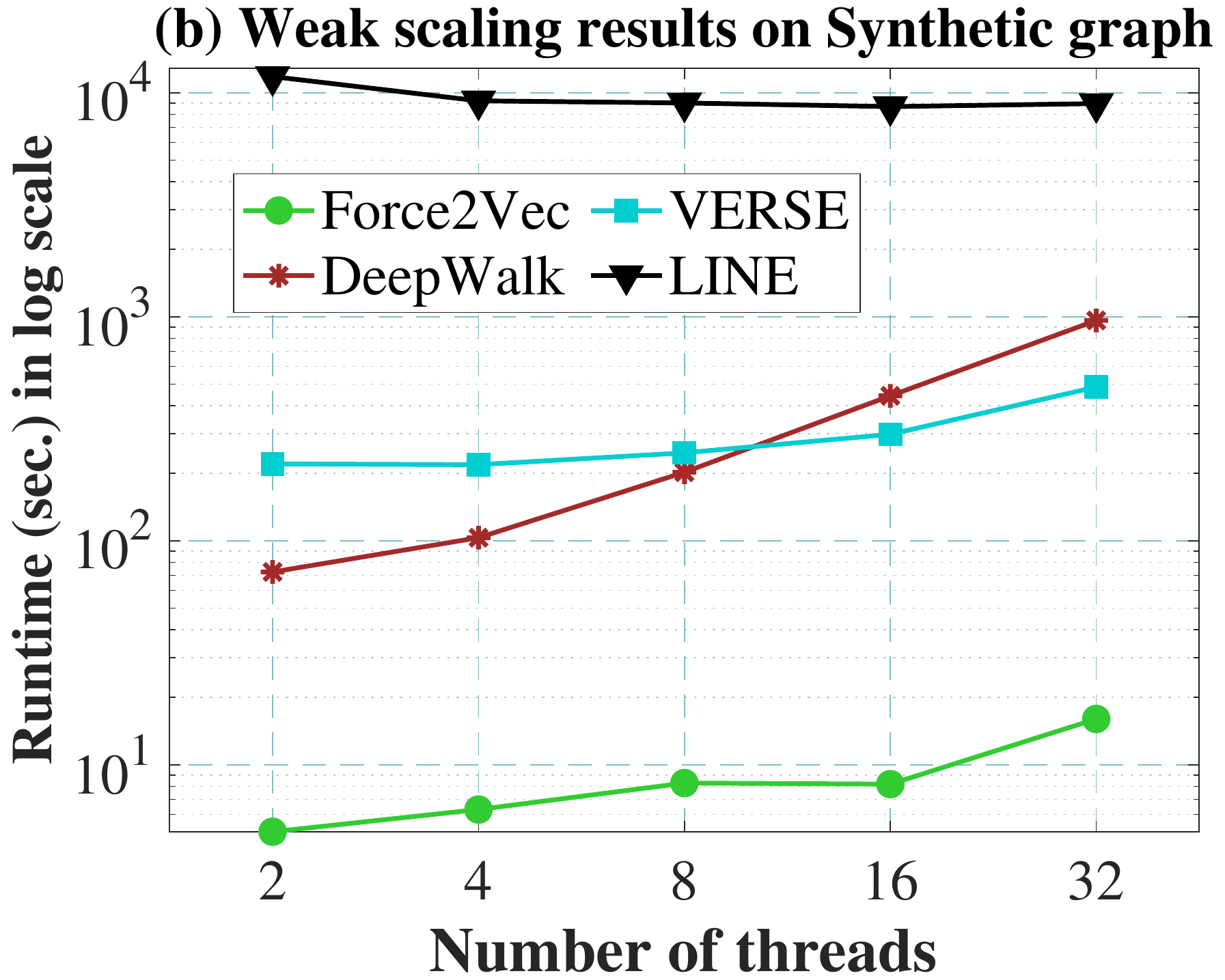}
    \includegraphics[width=0.29\linewidth,height=3.5cm]{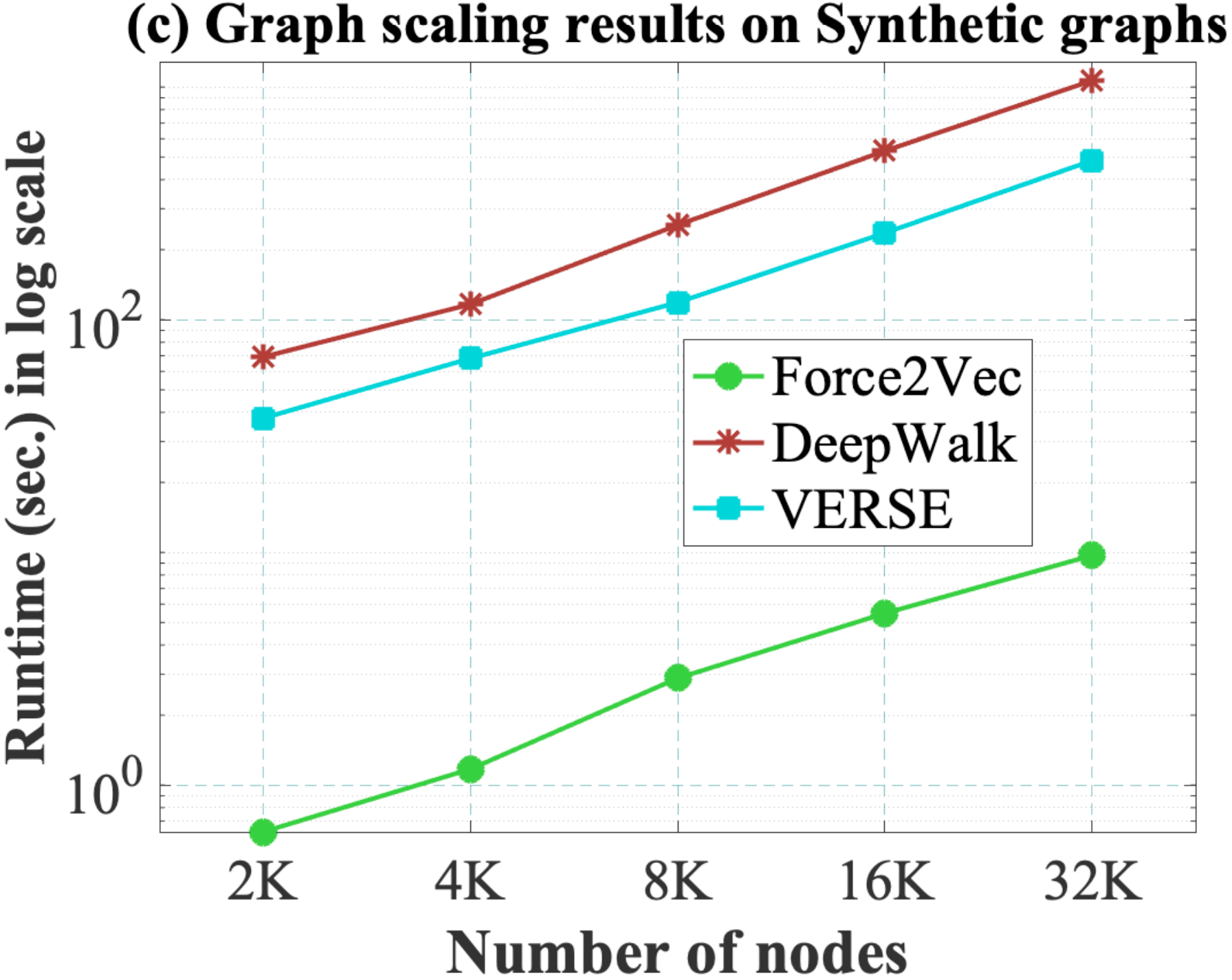}
    \caption{(a) Strong scaling results using the Flickr graph. (b) Weak scaling results using synthetic graphs. (c) Graph scaling results using synthetic graphs. Experiments were run on the Intel server.}
    \label{fig:scalingresults}
\end{figure*}
We analyze three types of scalability, namely, (i) strong scaling, where the number of threads are increased for a fixed input graph, (ii) weak scaling, where the size of the graph and the number of threads are increased while keeping the work per thread constant, and (iii) graph scaling where the size of the graph is increased for a fixed number of threads.
Strong scaling and weak scaling studies are commonly used by high-performance researchers to demonstrate the performance of algorithms with increasing computing resources. 
For our analyses, we use a real-world graph and several synthetic graphs. Some of the unsupervised methods such as DeepWalk, VERSE, LINE, HARP, and Force2Vec have parallel implementations for multicore processors. Thus, we pick these methods for the scalability study. We show strong scaling, weak scaling, and graph scaling results in Figs. \ref{fig:scalingresults} (a), \ref{fig:scalingresults} (b), and \ref{fig:scalingresults} (c), respectively.   

Fig. \ref{fig:scalingresults} (a) shows the strong scaling results for the Flickr dataset where the $x$-axis denotes different number of threads and the $y$-axis denotes the runtime in $\log$-scale. We see that the runtime of Force2Vec and VERSE decrease almost linearly with the increase of the number of threads. The better scalability of Force2Vec and VERSE make them suitable to run experiments on modern multicore processors. 
By contrast, HARP and DeepWalk exhibit limited scalability as they do not run significantly faster even when they use 32 threads. 

To conduct a weak scaling experiment, we create five synthetic networks using a benchmark dataset generator by \citeauthorandyear{lancichinetti2008benchmark}. The number of vertices in these five networks are 2K, 4K, 8K, 16K and 32K, respectively. Then, we run all methods using 2, 4, 8, 16, and 32 threads for graphs having 2K, 4K, 8K, 16K, and 32K vertices, respectively. In an idead weak scaling scenario, the runtime should remain the same for different numbers of threads as the problem size assigned per thread does not increase. In Fig. \ref{fig:scalingresults} (b), we observe that LINE, VERSE, and Force2Vec keep the runtime stable, whereas DeepWalk shows poor weak scaling performance.

In Fig. \ref{fig:scalingresults} (c), we show the graph scaling results for Force2Vec, VERSE and DeepWalk. We ran all methods for a fixed 48 threads using different graphs. We increase the number of nodes by a factor of 2 starting from a graph having 2K nodes generated similar to weak scaling \citeauthorandyear{lancichinetti2008benchmark}. All these methods show similar graph scaling results which is almost linear in the log-scale. 

{\bf Summary.}
Our experiments suggest that Force2Vec, LINE and VERSE show desirable strong and weak scaling behavior. 
Hence, these methods can exploit parallel processors to run faster (relative to their sequential runtime) and process large-scale graphs quickly.


\subsection{Training time for GNN methods}
In the recent years, GNNs emerge as a popular option for graph representation learning, especially when nodes are partially labeled to facilitate semi-supervised predictions. 
In this section, we analyze the training time of some popular GNN methods.
\begin{table*}[!htb]
\centering
\begin{tabular}{c|ccc|ccc} 
\hline
\multicolumn{1}{l|}{} & \multicolumn{3}{c|}{\textbf{Category 1}}                                & \multicolumn{3}{c}{\textbf{Category 2}}                                     \\ 
\hline
\textbf{Graphs}       & \textbf{GCN} & \textbf{GraphSAGE} & \textbf{FastGCN} & \textbf{GAT} & \textbf{ClusterGCN} & \textbf{GraphSAINT}  \\ 
\hline
Cora                  & 1.85         & 59.93              & 4.94             & 15.66        & 205.51              & 262.20               \\
Citeseer              & 2.01         & 216.35             & 3.61             & 67.90        & 228.49              & 220.3                \\
Pubmed                & 24.85        & 771.56             & 3.39             & 88.54        & 260.78              & 661.60               \\
\hline
\end{tabular}
\caption{Training and validation time (in seconds) for different semi-supervised methods on the Intel server. The same hardware configurations were used in all experiments.}
\label{tab:runtime_gnns_cpu}
\end{table*}
We measure the training time of GCN, GraphSAGE and FastGCN using their original source codes which are publicly available (Category 1). To measure the runtime of GAT, ClusterGCN and GraphSAINT, we use source codes available in the PyTorch Geometric (PyG) framework (Category 2) \cite{feylenssen2019}. We ran all these methods on the Intel Skylake server and reported their training time in Table \ref{tab:runtime_gnns_cpu}. We observe that GCN is the fastest GNN method for smaller graphs, but FastGCN runs faster than other methods for bigger graphs. 
The benefit of FastGCN stems from a sampling approach that is used to accelerate the training process while keeping the accuracy competitive to GCN and GraphSAGE. 
ClusterGCN and GraphSAINT apply a clustering technique and a sampling technique, respectively, as a pre-processing step before starting the training procedure. 
These expensive reprocessing steps make ClusterGCN and GraphSAINT slower than some of their peers. 

{\bf Summary.} GCN and FastGCN are generally the fastest GNN methods. They have comparatively simpler computational overheads compared to other GNN methods.

\begin{table*}[!htb]
\centering
\arrayrulecolor{black}
\begin{tabular}{!{\color{black}\vrule}c!{\color{black}\vrule}c!{\color{black}\vrule}c!{\color{black}\vrule}c!{\color{black}\vrule}c!{\color{black}\vrule}c!{\color{black}\vrule}c!{\color{black}\vrule}c!{\color{black}\vrule}c!{\color{black}\vrule}} 
\arrayrulecolor{black}\cline{1-1}\arrayrulecolor{black}\cline{2-9}
\multirow{3}{*}{\textbf{Graphs}} & \multicolumn{4}{c!{\color{black}\vrule}}{\textbf{GPU}}                                                          & \multicolumn{4}{c!{\color{black}\vrule}}{\textbf{CPU}}                                                           \\ 
\cline{2-9}
                                 & \multicolumn{2}{c!{\color{black}\vrule}}{\textbf{GAT}} & \multicolumn{2}{c!{\color{black}\vrule}}{\textbf{GCN}} & \multicolumn{2}{c!{\color{black}\vrule}}{\textbf{GAT}} & \multicolumn{2}{c!{\color{black}\vrule}}{\textbf{GCN}}  \\ 
\cline{2-9}
                                 & \textbf{PyG} & \textbf{DGL}                            & \textbf{PyG} & \textbf{DGL}                            & \textbf{PyG} & \textbf{DGL}                            & \textbf{PyG} & \textbf{DGL}                             \\ 
\arrayrulecolor{black}\cline{1-1}\arrayrulecolor{black}\cline{2-9}

Cora                             & {\cellcolor[rgb]{0.851,0.918,0.827}}1.158        & 1.630                                   & {\cellcolor[rgb]{0.851,0.918,0.827}}0.680        & 1.358                                   & {\cellcolor[rgb]{0.851,0.918,0.827}}4.762        & 8.161                                   & 4.340        & {\cellcolor[rgb]{0.851,0.918,0.827}}2.799                                    \\ 
\hline
Citeseer                         & {\cellcolor[rgb]{0.851,0.918,0.827}}1.220        & 1.980                                   & {\cellcolor[rgb]{0.851,0.918,0.827}}0.685        & 1.021                                   & {\cellcolor[rgb]{0.851,0.918,0.827}}6.890        & 14.772                                  & 4.713        & {\cellcolor[rgb]{0.851,0.918,0.827}}3.700                                    \\
\hline
Pubmed                           & {\cellcolor[rgb]{0.851,0.918,0.827}}1.267        & 2.668                                   & {\cellcolor[rgb]{0.851,0.918,0.827}}0.826        & 1.300                                   & 43.951       & {\cellcolor[rgb]{0.851,0.918,0.827}}40.873                                  & 63.101       & {\cellcolor[rgb]{0.851,0.918,0.827}}4.590                                    \\ 
\hline
\end{tabular}
\arrayrulecolor{black}
\caption{Training and validation time (in seconds) of different semi-supervised methods in PyG and DGL frameworks with respect to GPU and CPU. We run all methods using same computing resources and experimental setup. Better runtime of each category is shown in bold face for each dataset.}
\label{tab:runtime_cpu_gpu}
\end{table*}

\subsection{Impact of Graph Learning Frameworks}
Over the last few years, several frameworks have been develop to simplify the development and testing of GNN and graph embedding methods.  
Among them, PyG \cite{feylenssen2019} and Deep Graph Library (DGL) \cite{wang2019dgl} have become popular. PyG is built on top of PyTorch whereas DGL supports PyTorch, MXNet, and Tensorflow backends. 
Here, we analyze the runtime performance of PyG and DGL frameworks using three benchmark datasets. For PyG and DGL experiments, we use GNN models with identical hyper-parameters.

Table \ref{tab:runtime_cpu_gpu} shows the training and validation time of two popular GNN models, GAT and GCN, using the PyG and DGL frameworks. We ran the same experiments on an Intel CPU and an NVIDIA GPU (see Table~\ref{tab:hardware}). 
We observed that PyG runs GNN models faster than DGL on our GPU. However, there is no clear winner when their performances are compared on CPUs. For example, GCN implemented with DGL ran faster than its implementation with PyG. By contrast, PyG's CPU implementation of GAT runs faster for most graphs considered in this paper. 
Both of these frameworks are under active development and their performance will continue to improve over time.

{\bf Summary.} At the time of this survey, PyG ran faster than DGL on our GPU, but their CPU performances were mixed. Since the performance gap is not significant, users can use wither frameworks based on their preferences.  


\subsection{Memory Consumption Analysis}
\begin{figure*}[!ht]
    \centering
    \includegraphics[width=0.42\linewidth,height=4cm]{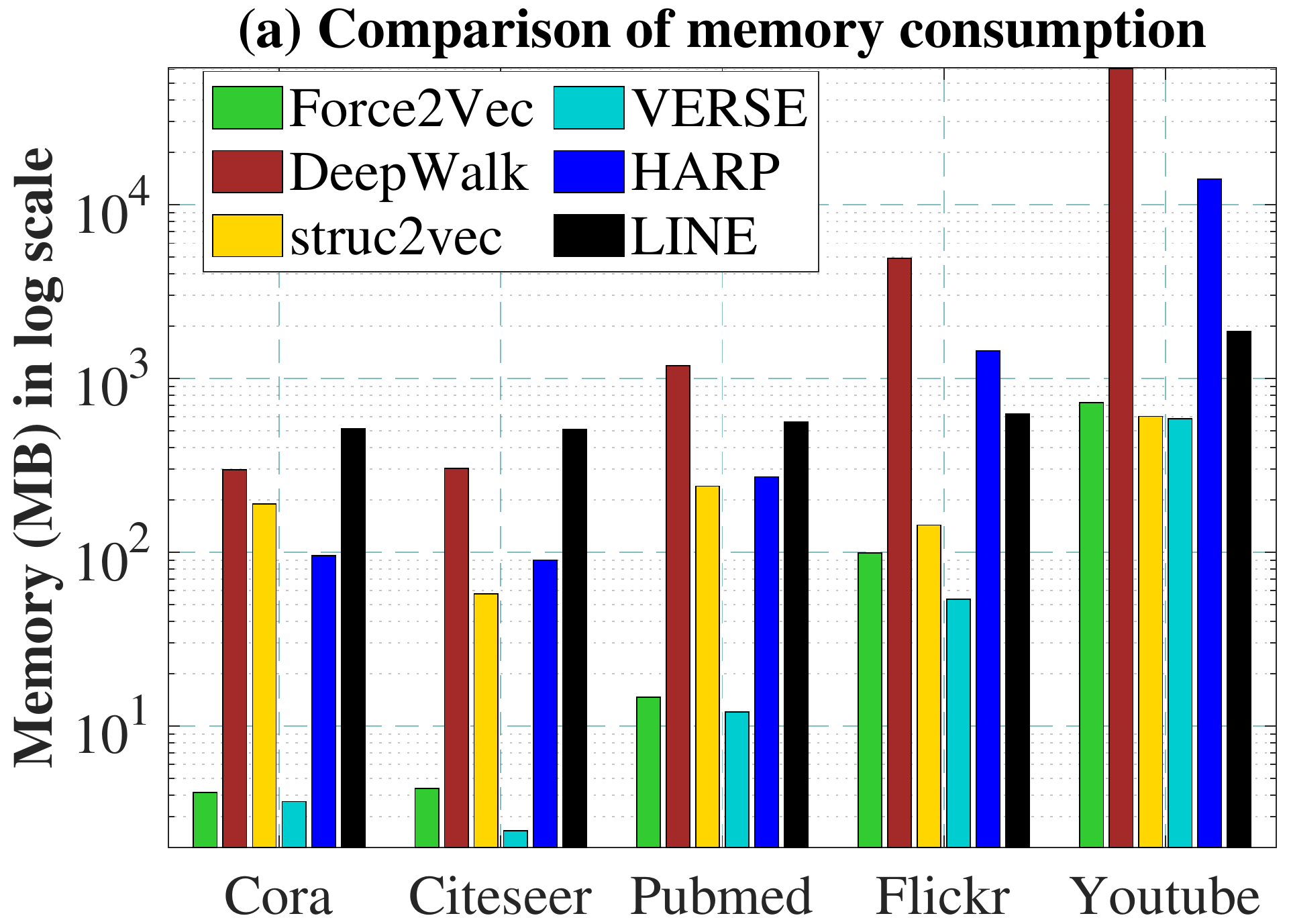}
    \includegraphics[width=0.22\linewidth,height=4cm]{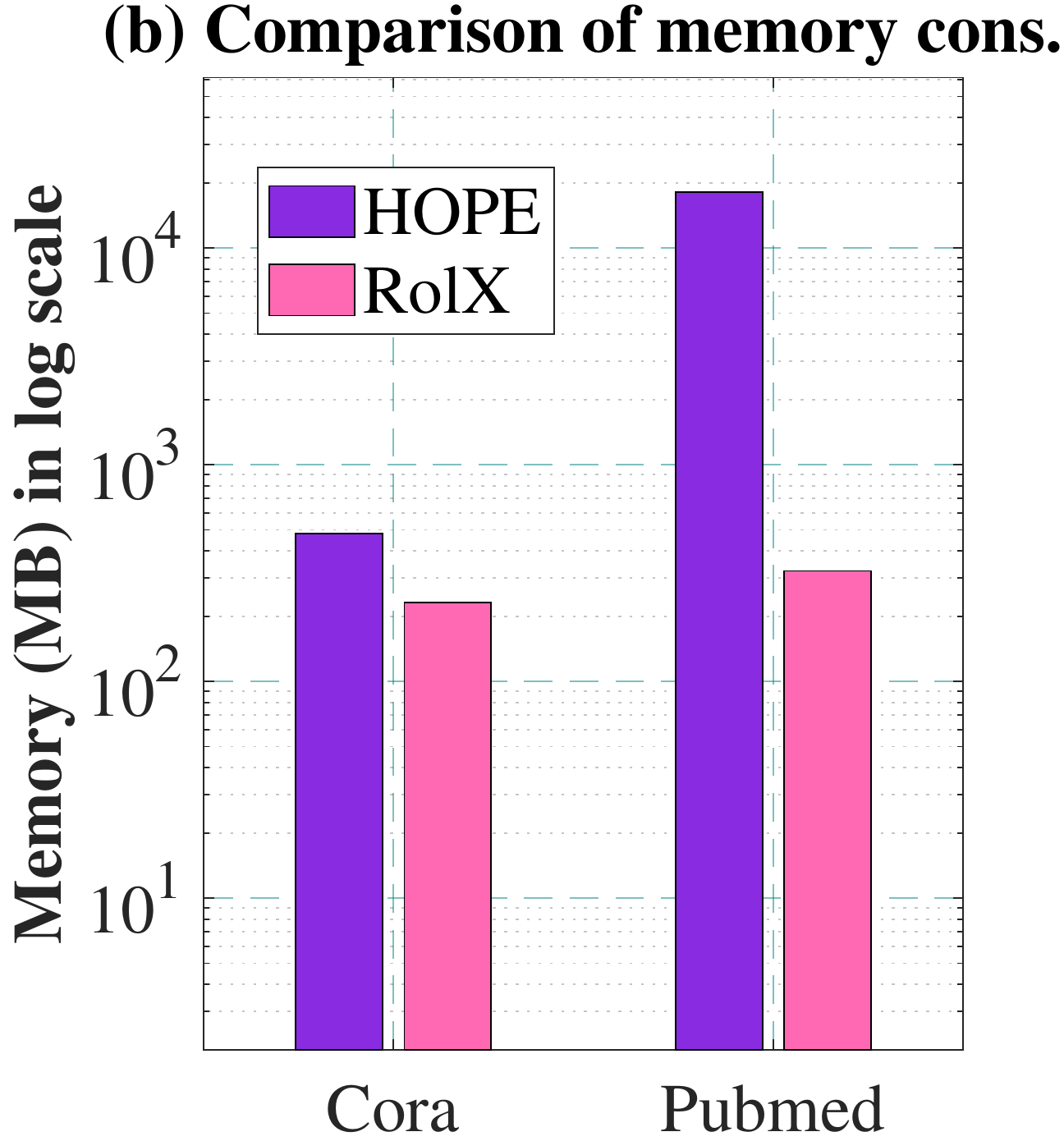}
    \includegraphics[width=0.32\linewidth,height=4cm]{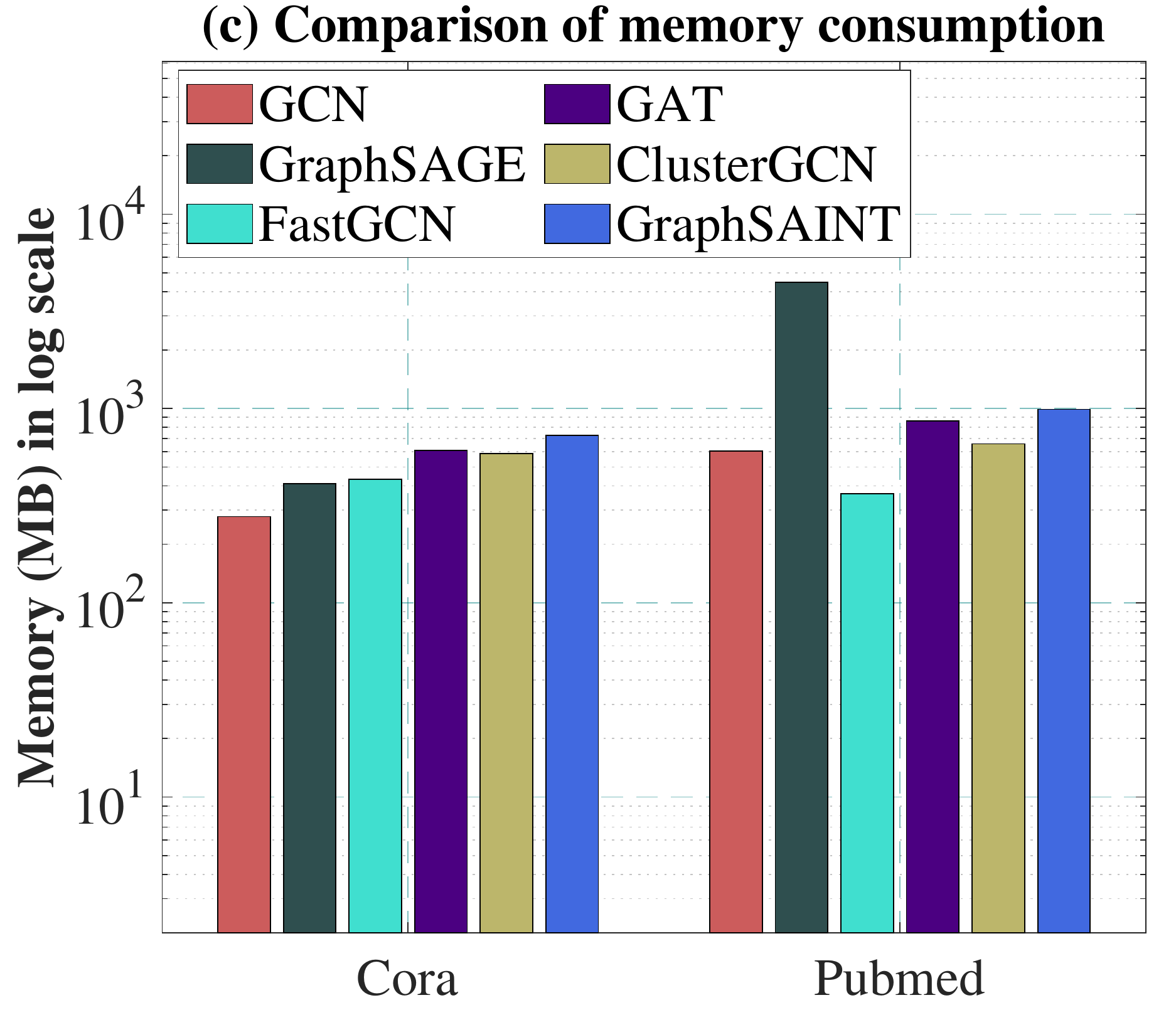}
    \caption{(a) Memory consumption by unsupervised graph embedding methods for different datasets. (b) Memory consumption by matrix factorization and feature engineering based methods for two benchmark datasets. (c) Memory consumption by GNN methods for two benchmark datasets.}
    \label{fig:abcmemoryconsumption}
\end{figure*}
Memory requirement is an important property of graph embedding and GNN algorithms, and it often determines whether a method can successfully run on various GPUs and CPUs.
We analyze memory consumption of a method using a Python package called {\myfont memory-profiler} and report the measurement in Megabytes in Fig. \ref{fig:abcmemoryconsumption}. Among unsupervised graph embedding algorithms in Fig. \ref{fig:abcmemoryconsumption} (a), VERSE consumes the least memory. 
The memory profile of Force2Vec is also similar to VERSE.
Both of these methods only allocate necessary data structures for the input sparse graph and output embedding. 
Random-walk based methods, on the other hand, consume more memory because they need to store a set of random walks generated in the preprocessing step. For example, DeepWalk samples a set of random walks for each vertex and stores those in arrays before passing them to the \emph{word2vec} model. 
DeepWalk stores random walks in disk when the generated walks do not fit in memory, which can be significantly more expensive.
For the Pubmed graph in Fig. \ref{fig:abcmemoryconsumption} (a), VERSE and Force2Vec consume around 12MB and 14.63MB of memory, respectively whereas struc2vec, LINE and DeepWalk consume 240MB, 560MB and 1182MB, respectively. Similarly, for Flickr, VERSE consumes around 50MB of memory whereas DeepWalk consumes more than 4GB. In Fig. \ref{fig:abcmemoryconsumption} (b), we show the memory consumption by HOPE and RolX which are chosen as the representative methods from the matrix factorization and feature engineering based methods, respectively. We see that HOPE consumes more than 17GB of memory for the Pubmed dataset which has only 19K vertices.
The high memory requirements of matrix factorization methods arise from their use 
of dense decomposition techniques (e.g., singular value decomposition). 

Fig. \ref{fig:abcmemoryconsumption} reports the memory consumption of various GNN methods. 
We observe that GNNs consume a significant amount of memory even for small graphs such as Cora. 
Most GNNs store weight matrices, biases and hidden representations in all intermediate layers, which resulted in their high memory demands.
In particular, GraphSAGE consumes more memory for the Pubmed dataset than GCN and FastGCN because  GraphSAGE concatenates hidden vector in successive layers whereas other methods generate fixed length hidden vector for each node.  

{\bf Summary.}
Among all graph embedding methods, VERSE and Force2Vec are more memory efficient than others. 
GNN methods typically require more memory than unsupervised graph embedding methods. 


\subsection{Applications}
\label{sec:application}

\begin{figure*}[!htb]
    \centering
    \includegraphics[width=0.32\linewidth,height=4.2cm]{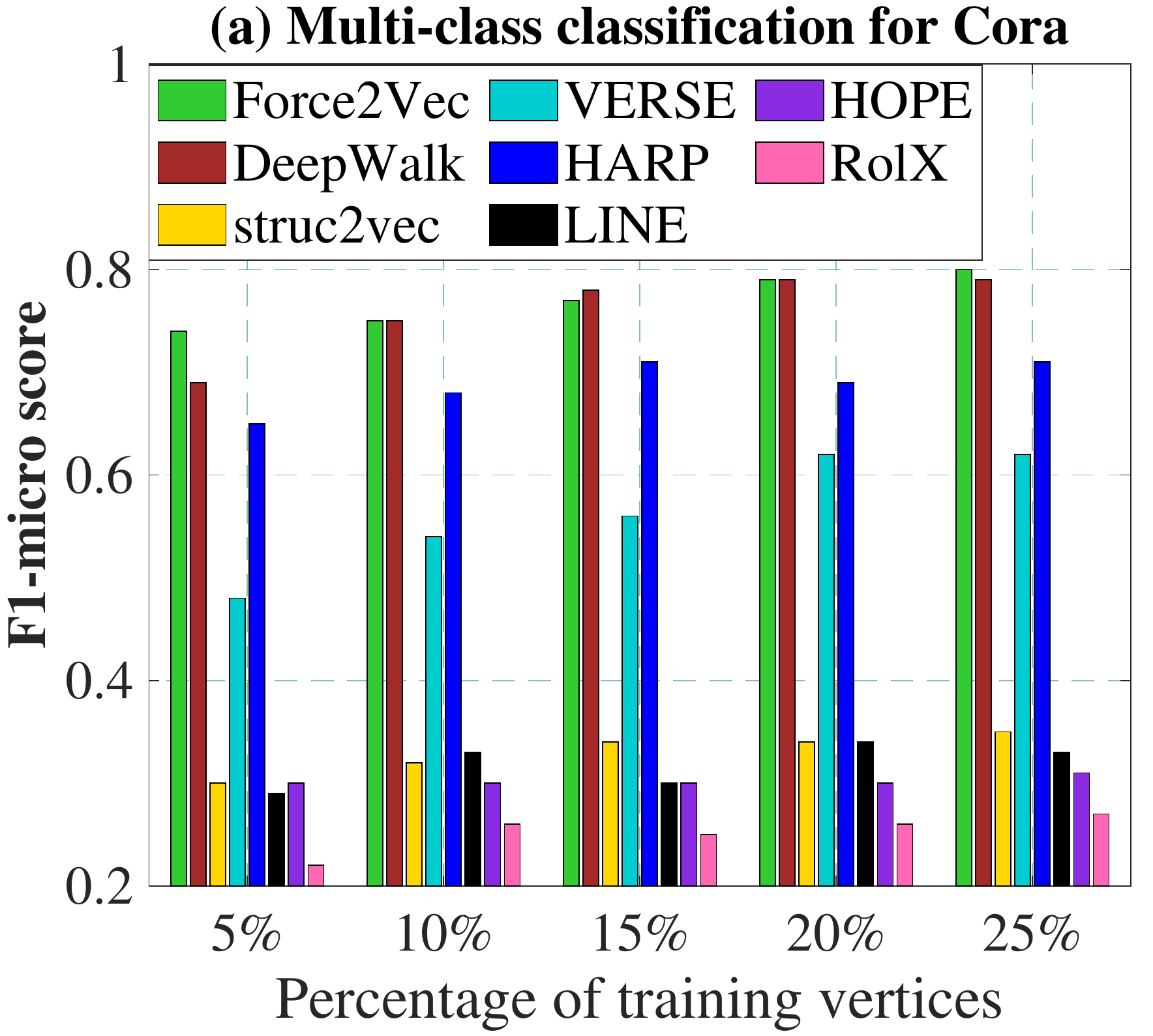}
    \includegraphics[width=0.32\linewidth,height=4.2cm]{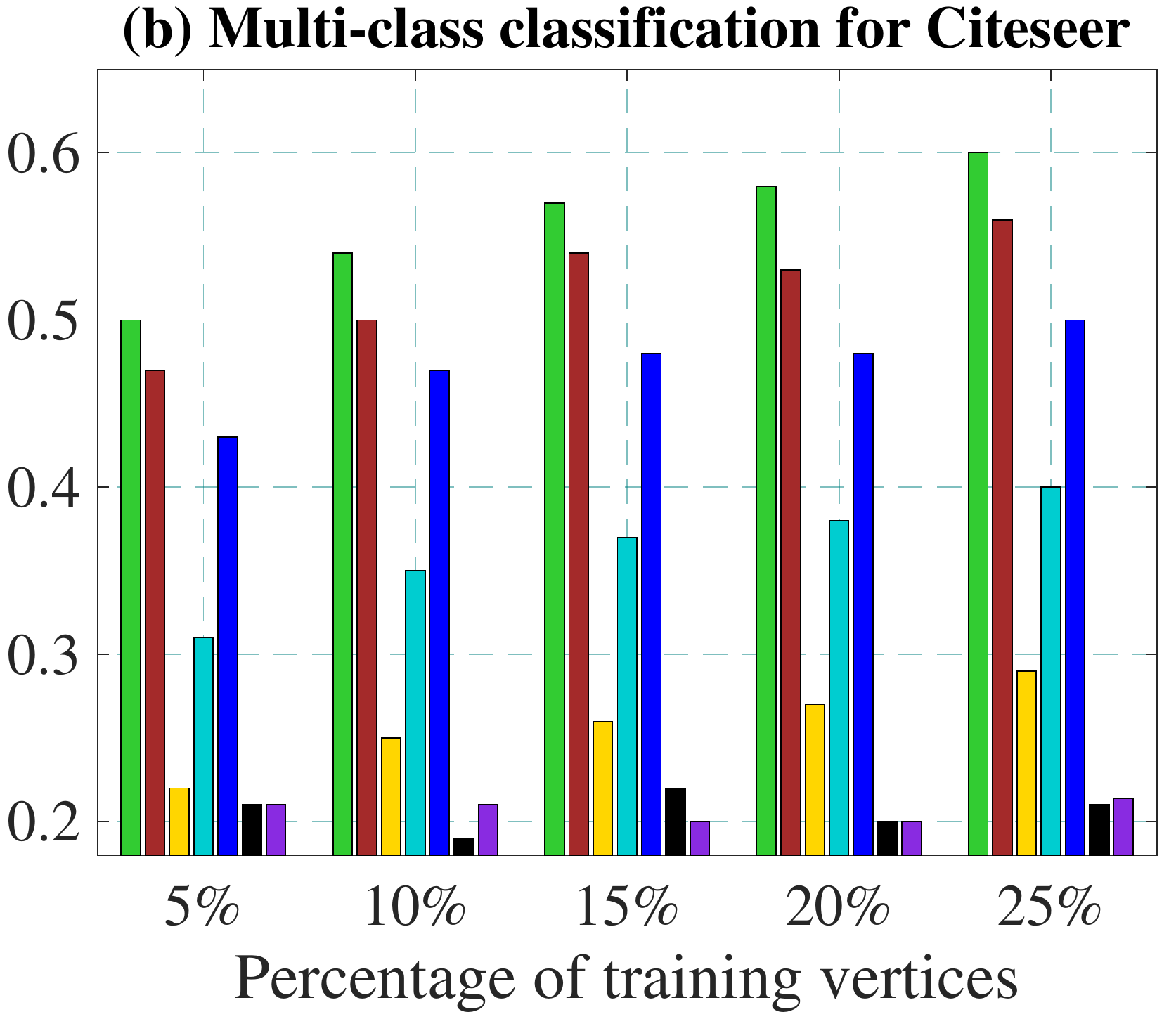}
    \includegraphics[width=0.32\linewidth,height=4.2cm]{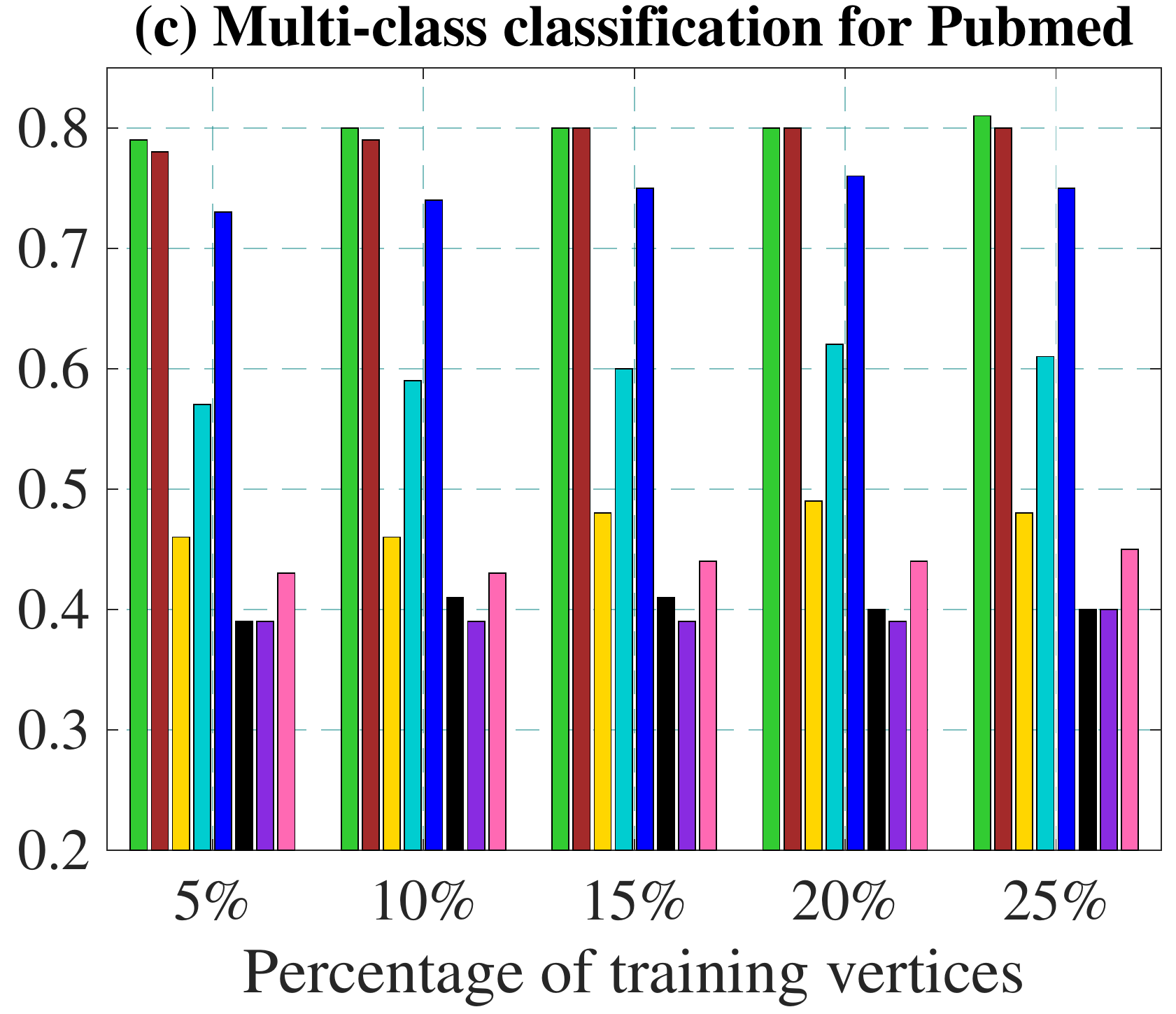}
    \includegraphics[width=0.32\linewidth,height=4.2cm]{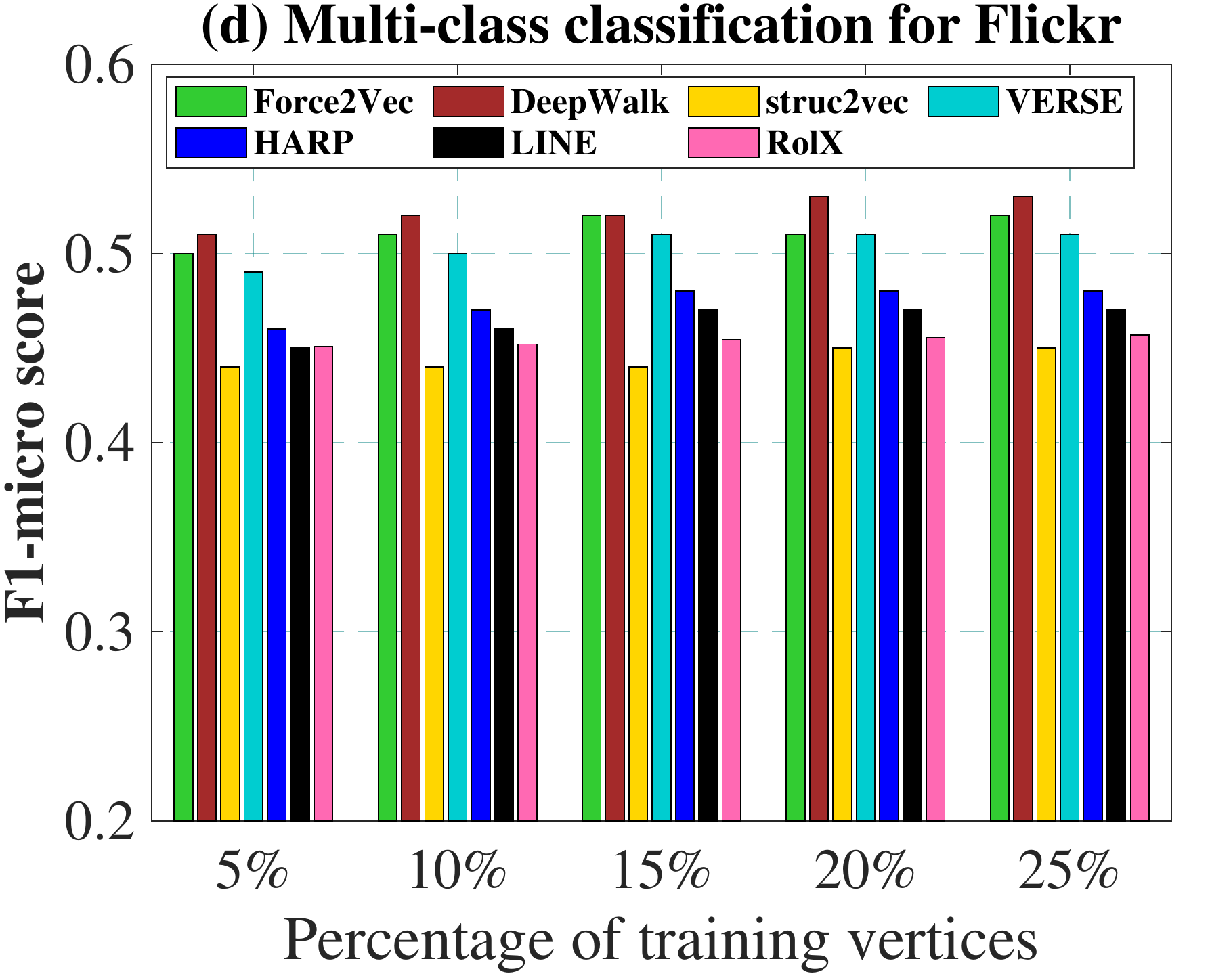}
    \includegraphics[width=0.32\linewidth,height=4.2cm]{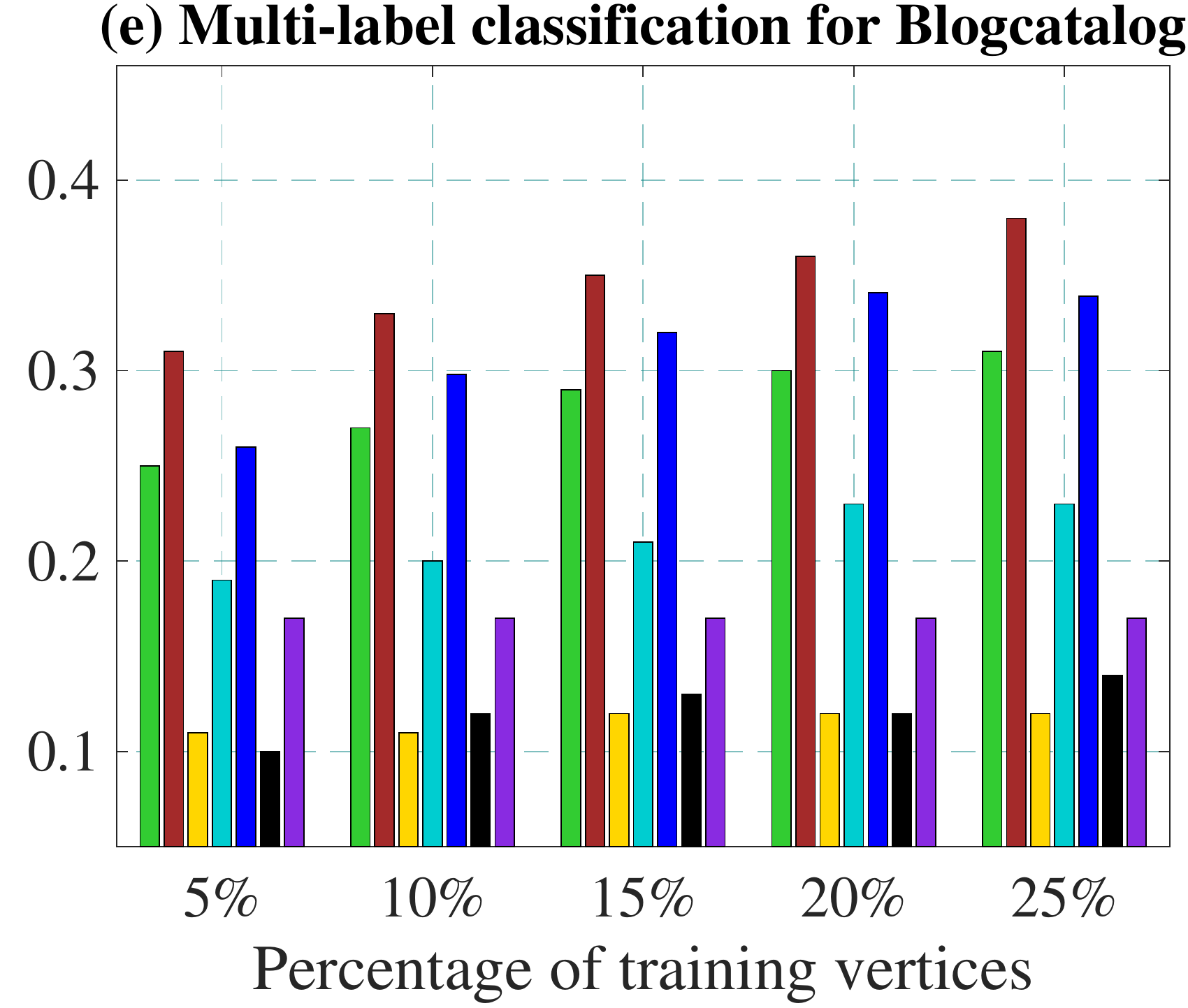}
    \includegraphics[width=0.32\linewidth,height=4.2cm]{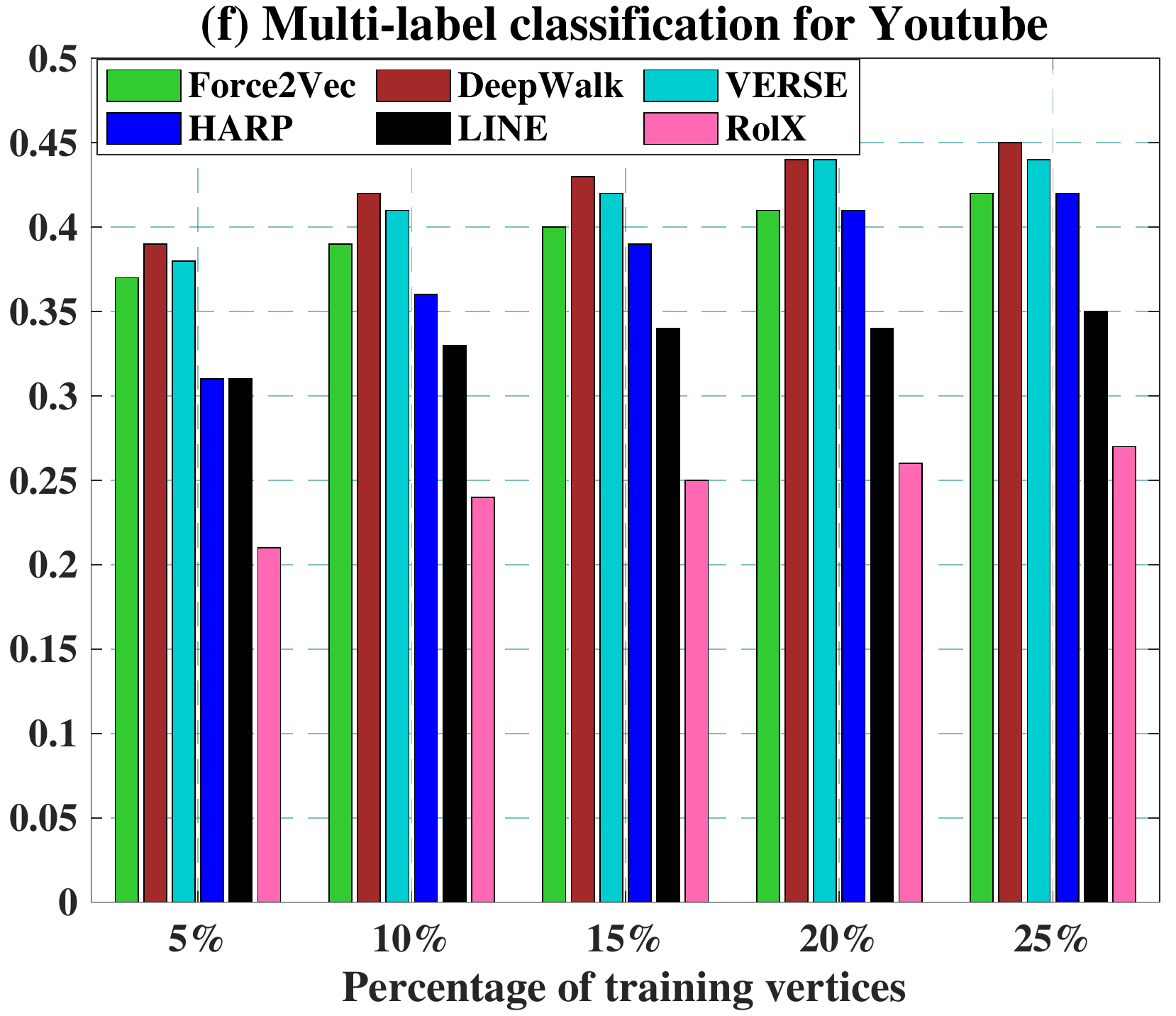}
    \caption{Results of F1-micro scores for node classification tasks for (a) Cora, (b) Citeseer, (c) Pubmed, (d) Flickr, (e) BlogCatalog, and (f) Youtube datasets.}
    \label{fig:abcdefmultilabel}
\end{figure*}

\subsubsection{Node Classification}
Node classification is one of the most celebrated applications of graph representation learning. 
In this task, labels of some vertices in the test set are predicted based on the structure of the graph and other information available for training data~\cite{goyal2018graph}. 
For a homogeneous graph where each vertex belongs to a unique class, node classification is formulated as a multi-class classification problem. 
By contrast, node classification on heterogeneous graphs where a vertex could belong to more than one class is formulated as a multi-label classification task.
In both cases, the label of a vertex can be inferred by using its embedding. 
For the unsupervised methods, node embeddings are generated based on the graph structure and node features. 
A classifier such as a logistic regression model is trained using a fraction of vertices with known labels and then the trained classifier is used to predict labels of other nodes whose labels are not known. 
For semi-supervised methods such as GNNs, the classification problem is incorporated in the loss function and solved in an end-to-end training of GNNs. 

\begin{figure*}[!htb]
    \centering
    \includegraphics[width=0.36\linewidth,height=4.2cm]{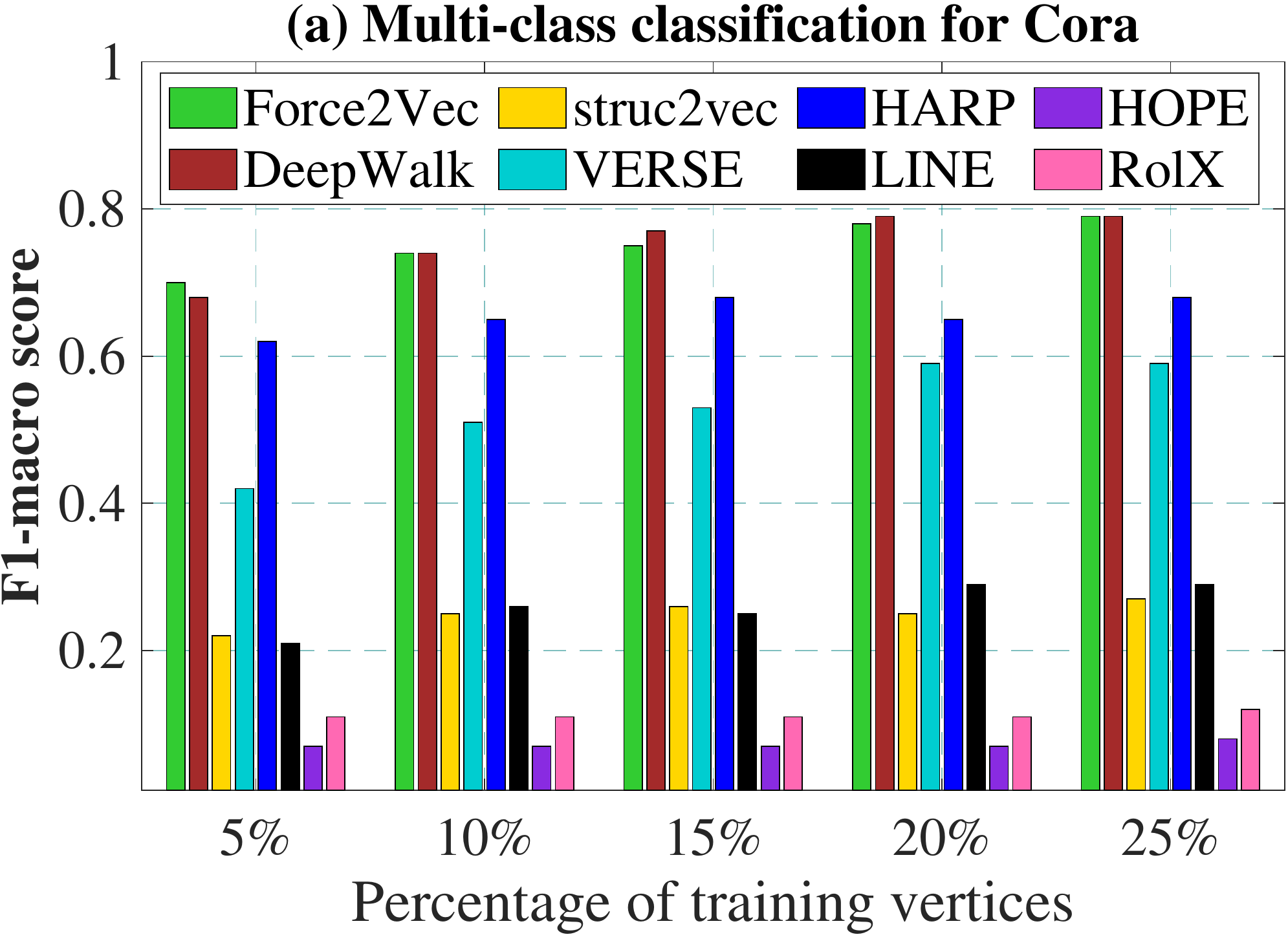}
    \includegraphics[width=0.28\linewidth,height=4.2cm]{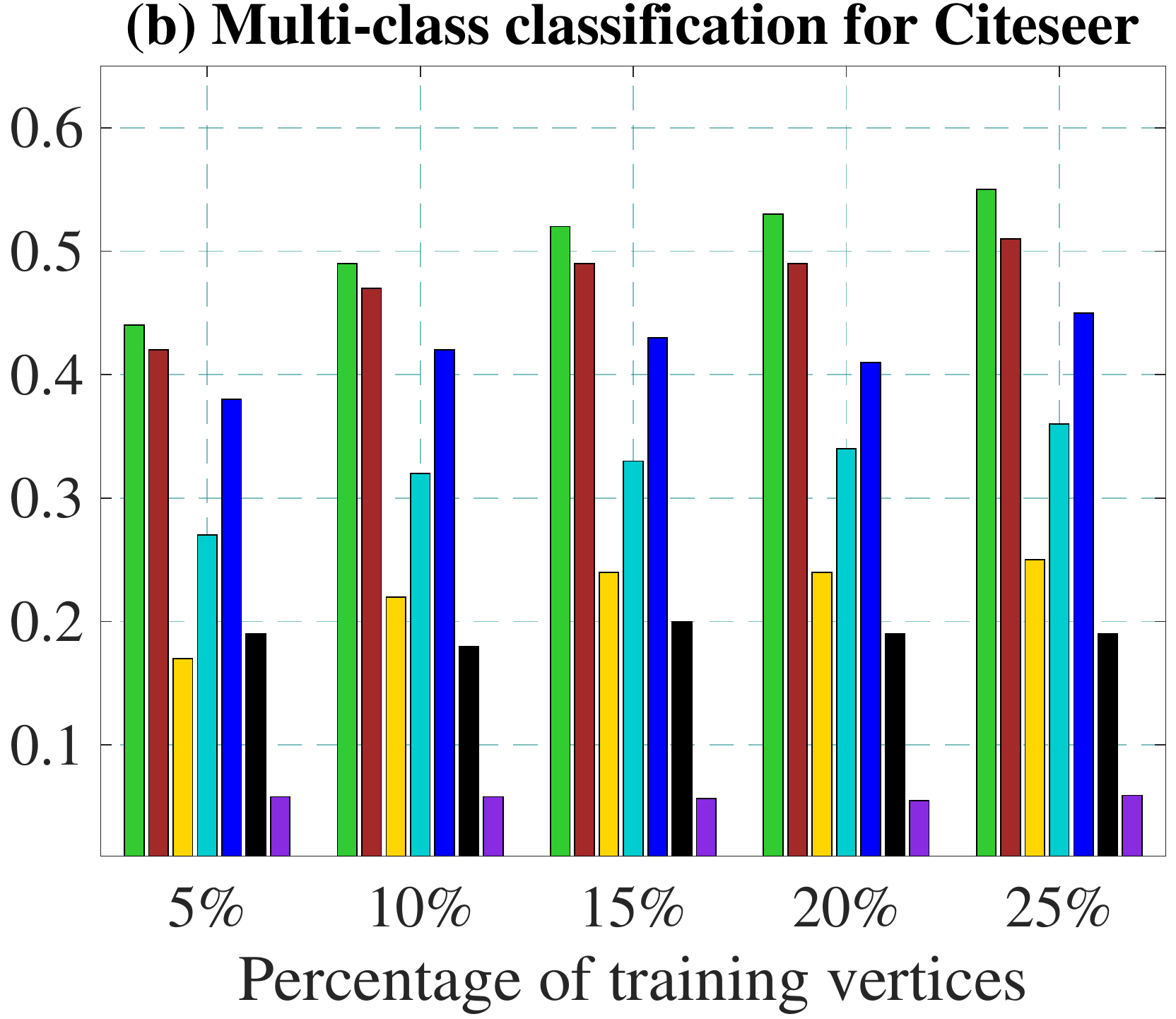}
    \includegraphics[width=0.28\linewidth,height=4.2cm]{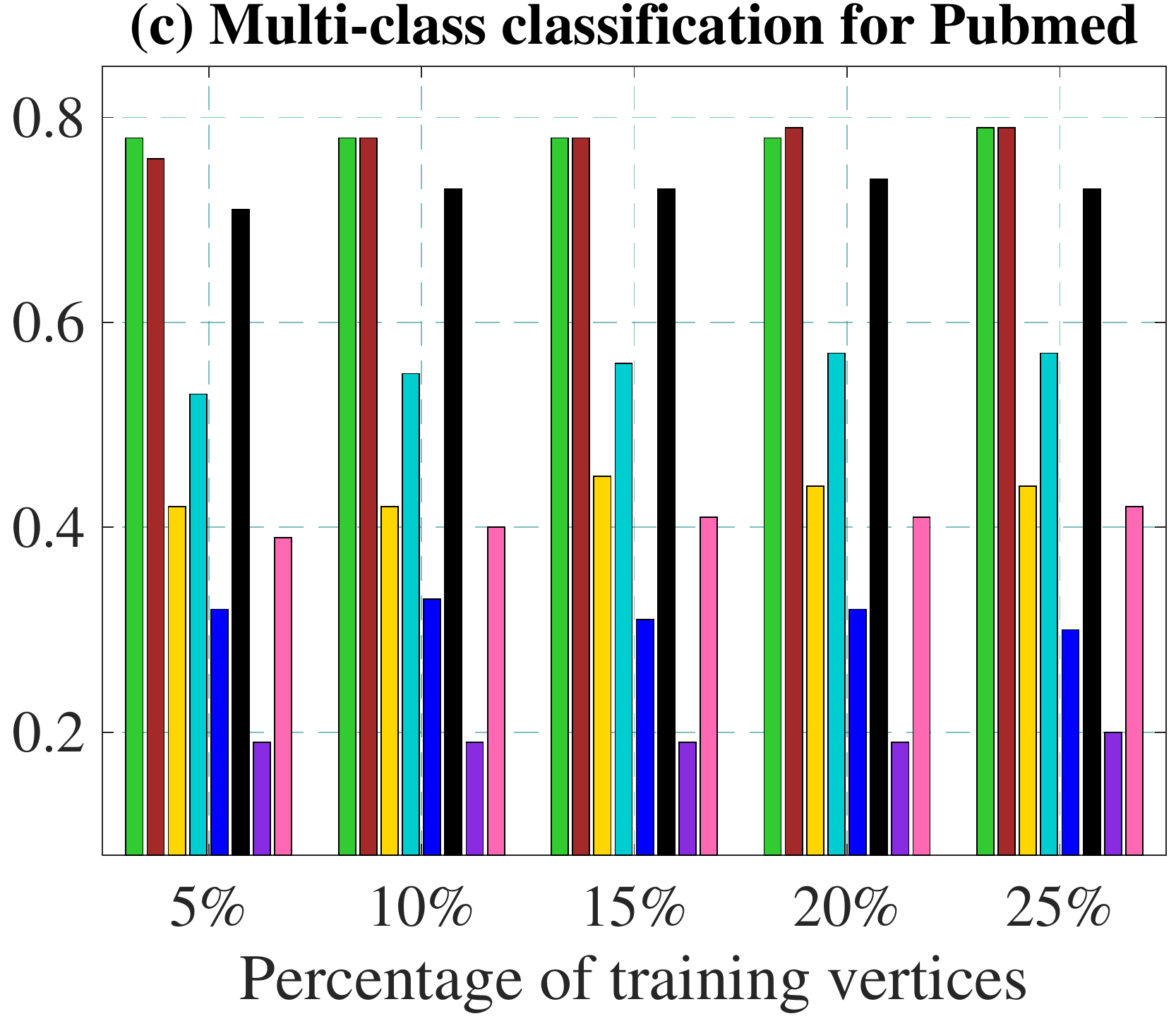}
    \caption{Results of F1-macro scores on multi-label classification task for (a) Cora, (b) Citeseer, (c) Pubmed}
    \label{fig:abcdefmacroscore}
\end{figure*}

\textbf{Unsupervised Methods.} We report the results of F1-micro scores for six different datasets in Fig \ref{fig:abcdefmultilabel}, where the $x$-axis denotes the percentage of vertices used to train the logistic regression model. The remaining vertices are used for testing when computing the F1-micro scores shown on the $y$-axis in Fig \ref{fig:abcdefmultilabel}. Generally, the increase of training data also boosts the F1-micro score for all methods. 
For a given training-testing split, no single method performs the best for all the datasets. Force2Vec performs better for Cora, Citeseer, and Pubmed, whereas DeepWalk performs better for Flickr, Blogcatalog and Youtube. 
Due to the high memory requirement, we could not run HOPE on larger graphs. 
We also report F1-macro scores for different datasets in Figs. \ref{fig:abcdefmacroscore} (a), (b), and (c) which show results similar to Figs. \ref{fig:abcdefmultilabel} (a), (b), and (c).

\begin{figure*}[!htb]
    \centering
    \includegraphics[width=0.34\linewidth,height=4cm]{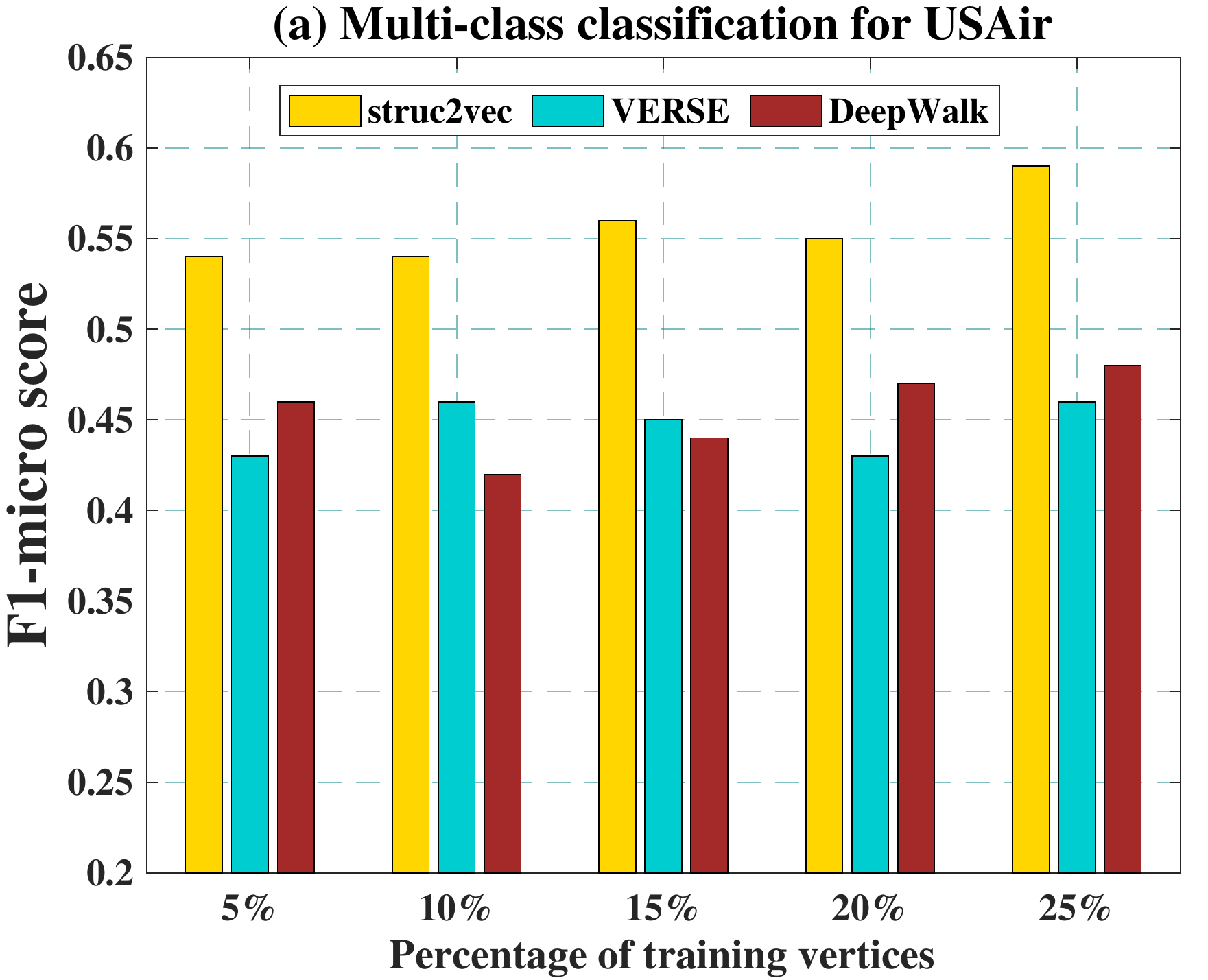}
    \includegraphics[width=0.3\linewidth,height=4cm]{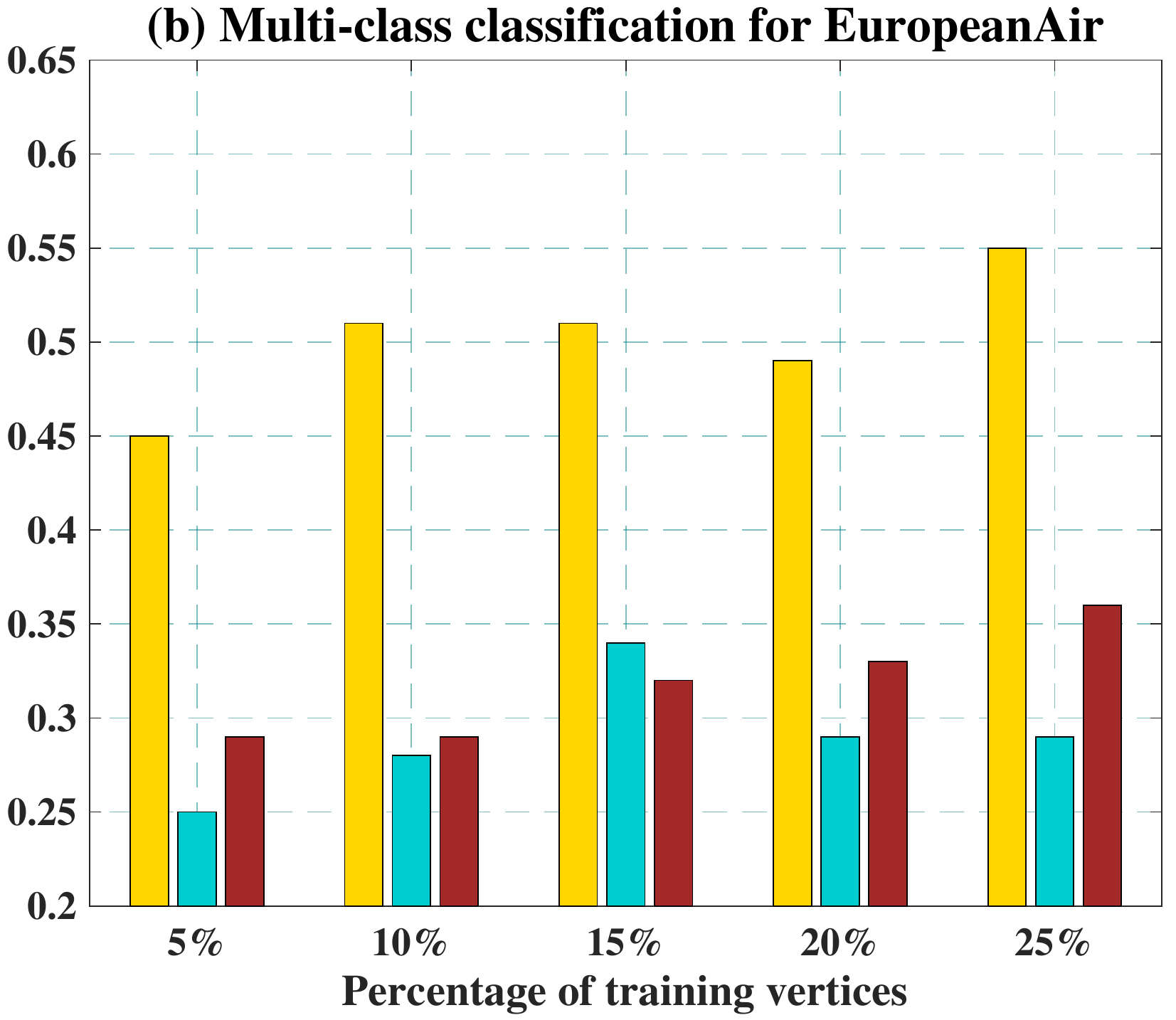}
    \includegraphics[width=0.3\linewidth,height=4cm]{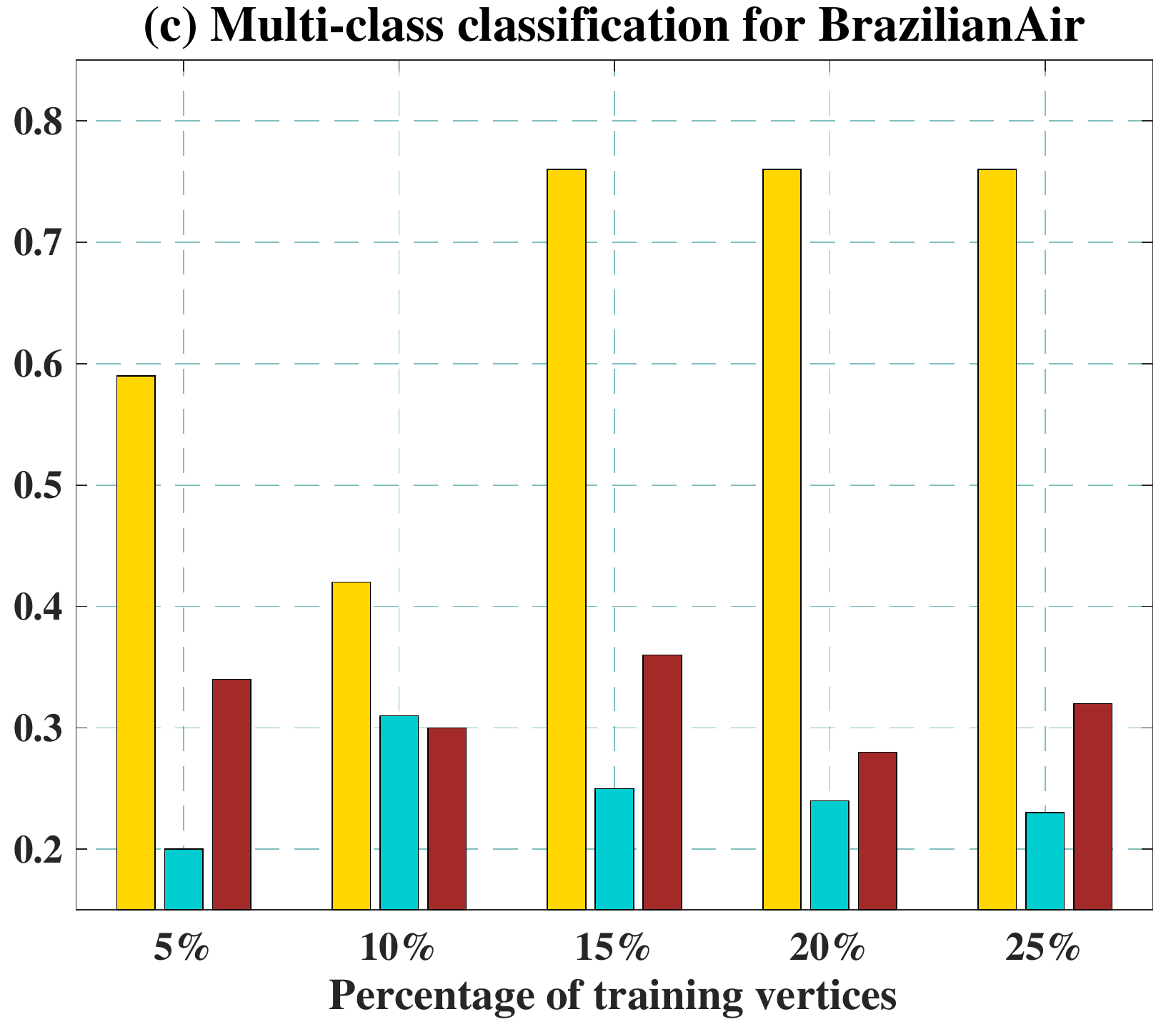}
    \caption{Results of F1-micro scores on multi-label classification task for (a) USA airport, (b) European airport, and (c) Brazilian airport datasets.}
    \label{fig:abcmultilabelair}
\end{figure*}
\begin{figure*}[!htb]
    \centering
    \includegraphics[width=0.47\linewidth,height=4.5cm]{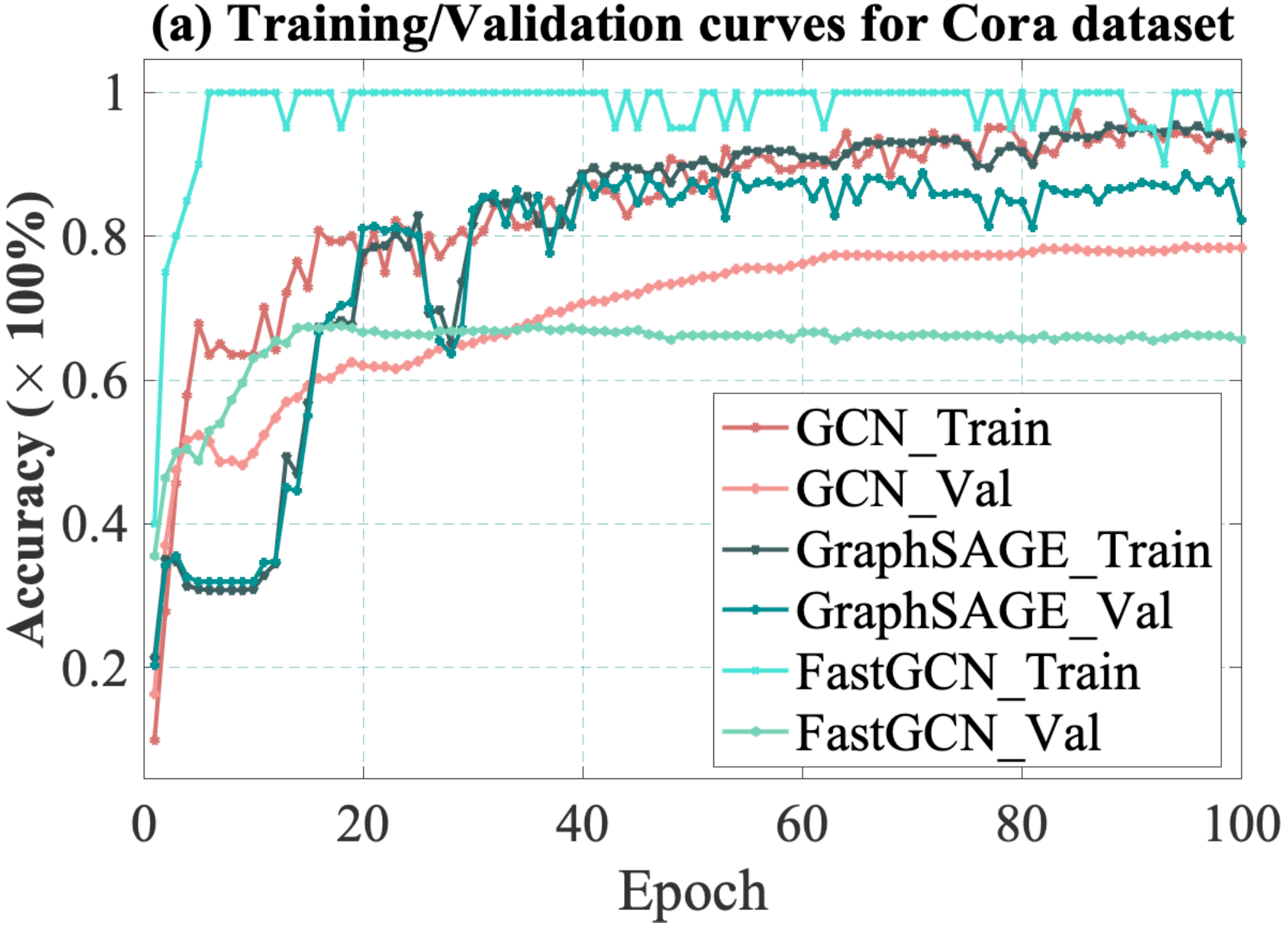}
    \includegraphics[width=0.47\linewidth,height=4.5cm]{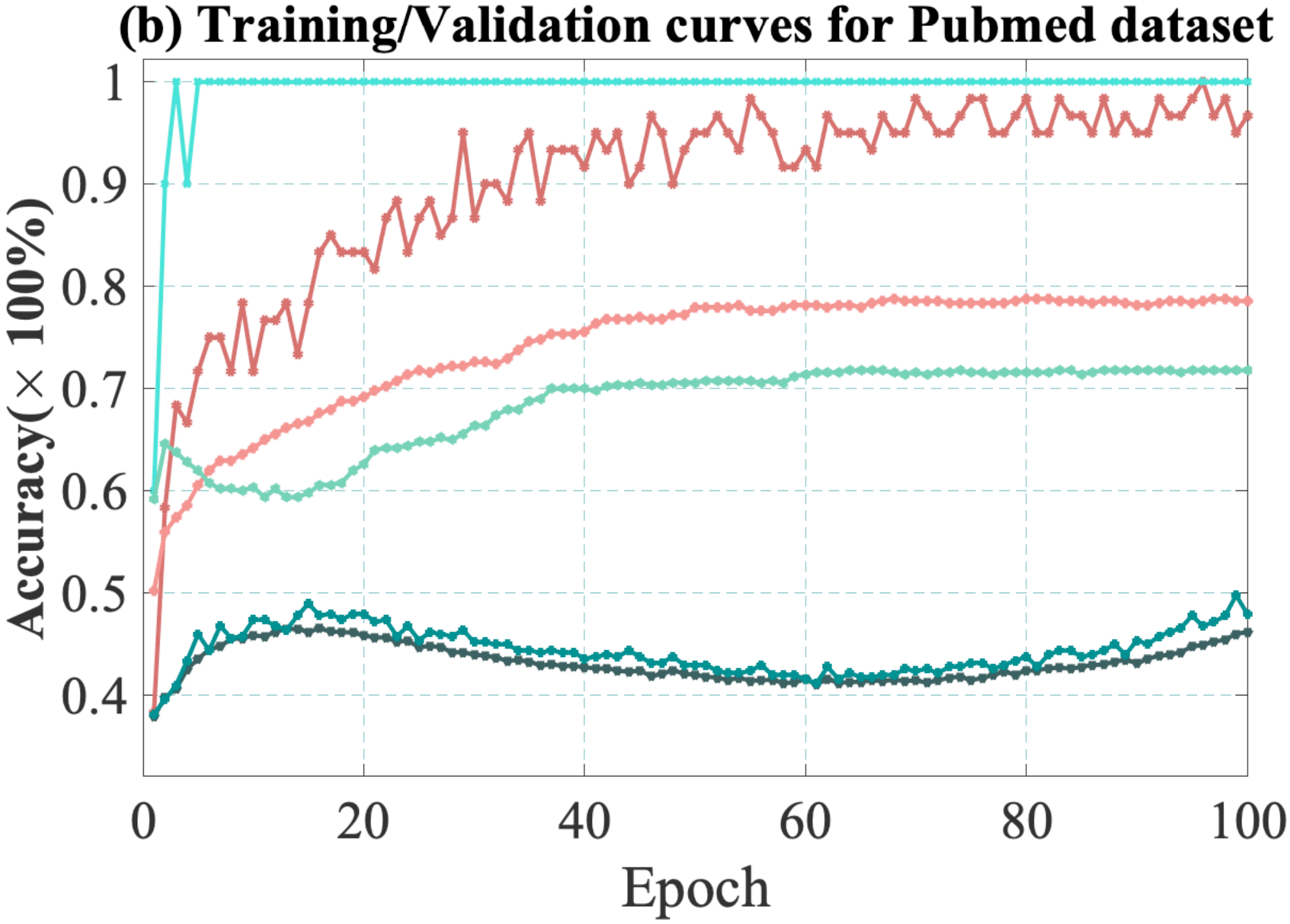}
    
    \vspace{-0.3cm}
    \caption{Training and validation accuracy for GNN methods on (a) Cora and (b) Pubmed datasets.}
    \label{fig:abgcntrainvalidation}
\end{figure*}

Graphs shown in Figs. \ref{fig:abcdefmultilabel} and  \ref{fig:abcdefmacroscore} 
exhibit strong homophilic relations (nodes with same labels are more likely to be connected by edges). 
It is well understood that most graph embedding and GNN methods perform reasonably well on  homophilic graphs, but do not perform well for heterophilic graphs.
There are a few methods such as struc2vec that are especially designed for this type of graphs.
struc2vec discovers structural equivalence instead of node proximity and thus performs well for heterophilic graphs. 
We collect three airport datasets from \cite{ribeiro2017struc2vec} to conduct experiments on structural equivalence under the same settings used in previous classification experiments and report the results in Fig. \ref{fig:abcmultilabelair}. 
We observe that struc2vec indeed outperforms DeepWalk and VERSE in the node classification task for all these airport datasets.

\textbf{Graph Neural Networks.} Fig. \ref{fig:abgcntrainvalidation} shows the classification accuracy (F1-micro scores) for three GNN methods using Cora and Pubmed datasets. For this experiment, we divide all vertices in a graph into three sets: (1) training set, (2) validation set, and (3) testing set. For both graphs, we choose 1000 random vertices without replacement in the testing set, 500 random vertices without replacement in the validation dataset, and 20 vertices per-class in the training set. In Fig. \ref{fig:abgcntrainvalidation} (a), we can observe that the \graphsage{} model has robust training and validation in F1-micro scores for the Cora dataset. On the other hand, \emph{FastGCN} achieves higher values in F1-micro score, whereas it shows poor validation performance. 
On test samples of the Cora dataset, \emph{GCN}, \graphsage{} and \emph{FastGCN} achieve F1-micro scores of 0.81, 0.85 and 0.72, respectively. In Fig. \ref{fig:abgcntrainvalidation} (b), we report the results for the Pubmed dataset. We oberserve that \graphsage{} performs poorly in both training and validation cases. \emph{FastGCN} achieves a high F1-micro score in the training set, as usual, whereas it shows poor performance in the validation set but not as bad as \graphsage{}. \emph{GCN} performs better in the validation set. In test cases of the Pubmed dataset, \emph{GCN}, \graphsage{} and \emph{FastGCN} achieve F1-micro scores of 0.78, 0.44 and 0.72, respectively. Notice that \graphsage{} or \emph{GCN} achieves higher or similar accuracy compared to other types of methods for the Cora dataset, but using the same experimental setting, they could not reach the same level of accuracy as DeepWalk or Force2Vec for the Pubmed dataset.

\begin{figure*}[!htb]
    \centering
    \includegraphics[width=0.47\linewidth,height=4.5cm]{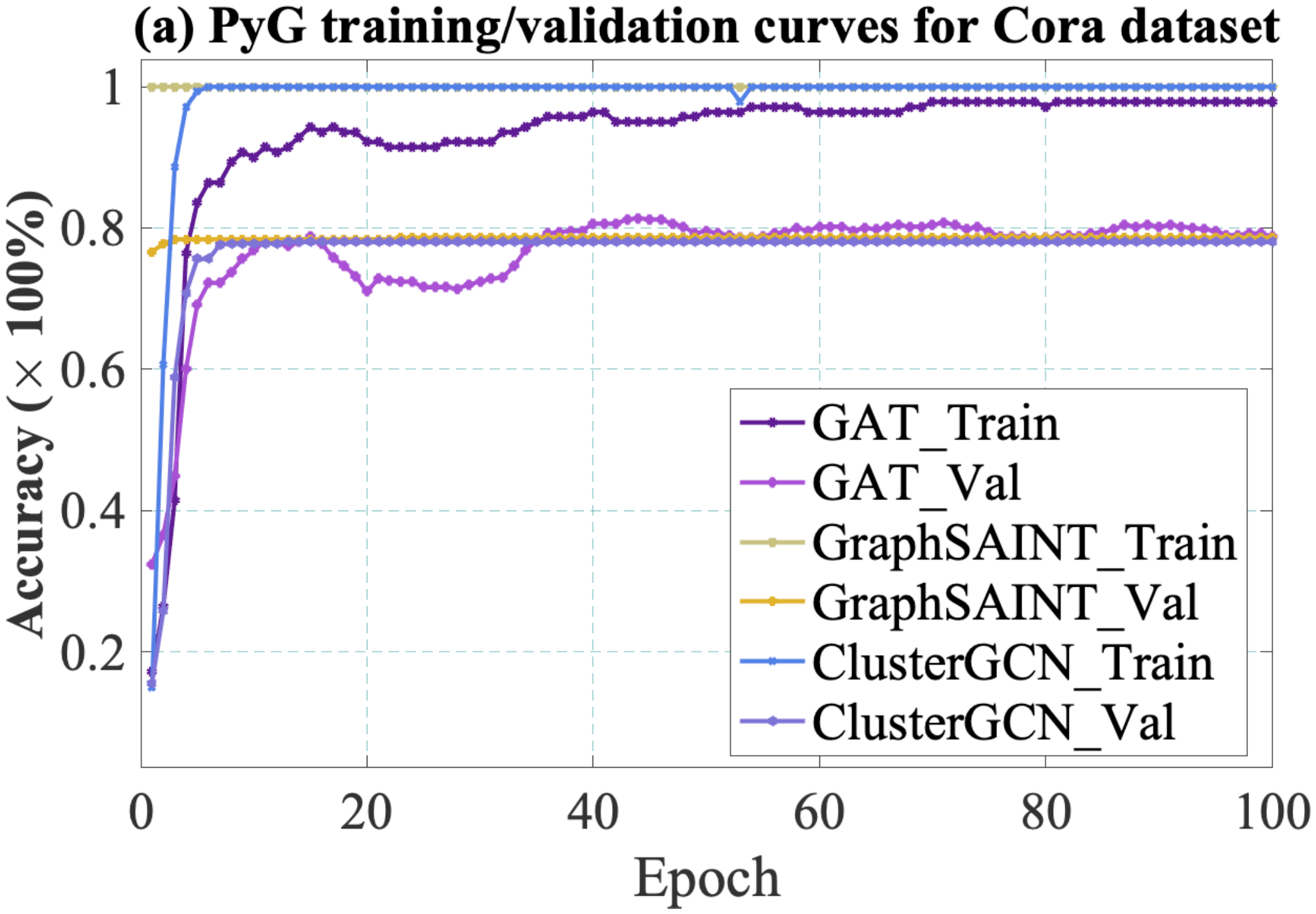}
    \includegraphics[width=0.47\linewidth,height=4.5cm]{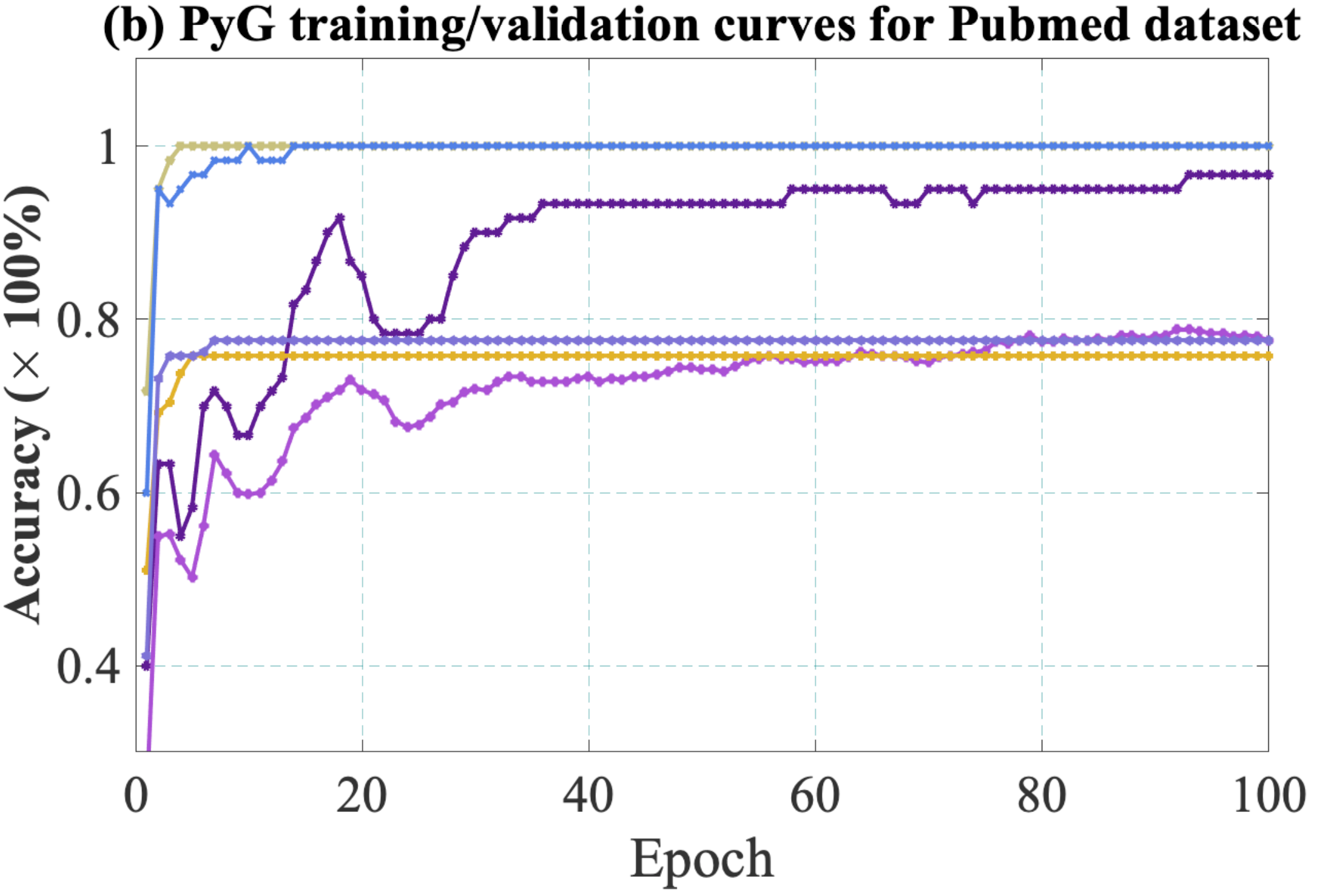}
    
    \vspace{-0.3cm}
    \caption{Training and validation accuracy for GNN methods in PyG framework using (a) Cora and (b) Pubmed datasets.}
    \label{fig:pyg_gcntrainvalidation}
\end{figure*}

We also conduct experiments using the GAT, ClusterGCN and GraphSAINT methods available in the PyG framework. We report the resutls in Figs. \ref{fig:pyg_gcntrainvalidation} (a), and (b) for Cora and Pubmed datasets, respectively. We observe that the ClusterGCN and GraphSAINT method show better training accuracy over GAT method; however, their validation accuracy is slightly less than the GAT method for Cora dataset. For the Pubmed dataset, ClusterGCN shows better validation curve than other methods.

Nowadays, graph neural network has become a popular strategy for the graph mining. In the recent years, GNN methods are mainly proposed for node classification or graph classification tasks; however, other graph learning tasks such as link prediction, clustering, and visualization have been heavily studied using the unsupervised methods. Thus, without loss of generality, we focus on discussing unsupervised methods for these tasks in the rest of the experimental analyses.

\subsubsection{Link Predictions}
\label{sec:linkpred}
\begin{figure}[!ht]
    \centering
    \includegraphics[width=0.85\linewidth,height=3cm]{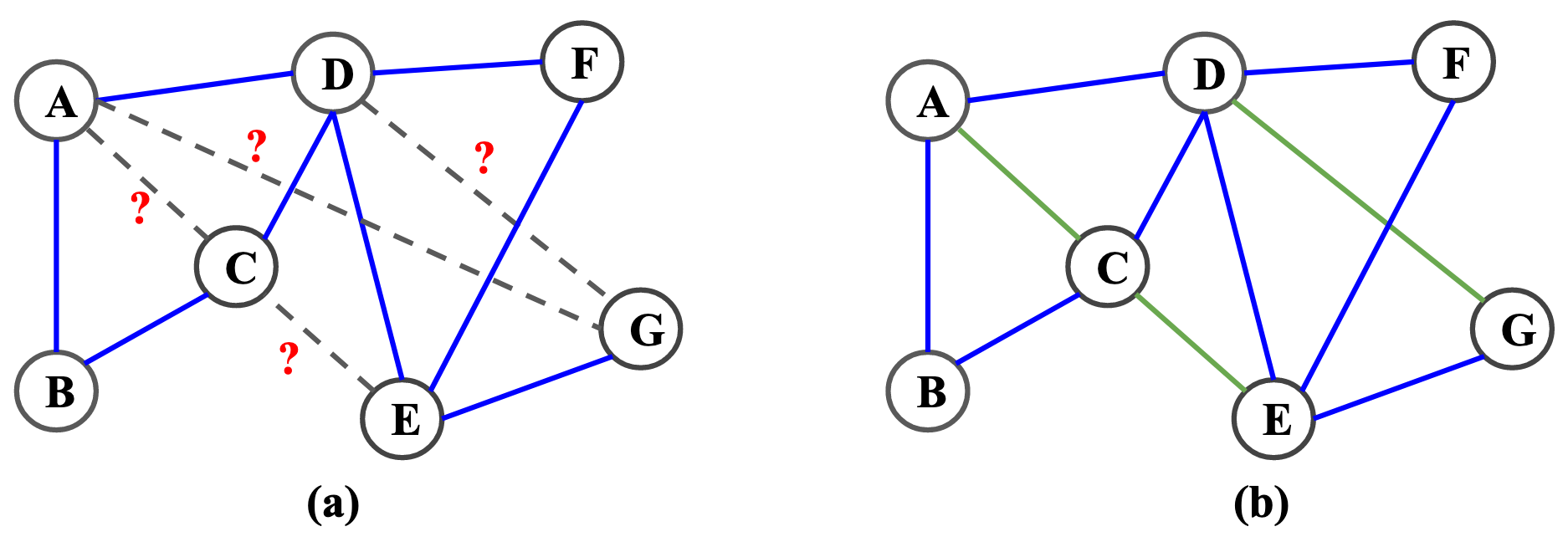}
    \caption{(a) An example graph having some missing links to be predicted (shown in dotted color with question marks). (b) Graph with predicted missing links shown in green colored.}
    \label{fig:linkpred}
\end{figure}

Link prediction itself is a broadened area of research in social network analysis~\cite{al2011survey}. It has many applications in social and biological networks, such as the recommendation of new friends in Facebook \cite{symeonidis2014link} and drug-disease association prediction in biomedical networks \cite{yue2019graph}. For link prediction, a good embedding method should capture local information from the network and preserve it in the embedding space. Thus, we can easily apply graph embedding to predict possible links in a graph. We show an example in Fig. \ref{fig:linkpred} (a), where blue colored edges represent existing links in the graph when the embedding is generated and dotted grey lines with question marks indicate links to be predicted. For the sake of simplicity, we can assume that there is a high probability to have a link between one vertex to another if one's neighbors are connected to another vertex. In Fig. \ref{fig:linkpred} (b), we show predicted links with green colored lines. For such link prediction tasks, we create a dataset from the embedding where positive samples are constructed from pairs of vertices with existing edges in the graph and negative samples are constructed from pairs of vertices having no edges. Thus, it becomes a binary classification problem which is mostly tackled by logistic regression~\cite{grover2016node2vec,tsitsulin2018verse}. When we take a pair of vertices, we can choose the Hadamard \cite{grover2016node2vec} vector operator to construct a new vector from embeddings of two vertices. This has been found effective in the literature, whereas some other vector operators exist, such as weighted L1, weighted L2, average\footnote{Several studies have found that average operator does not perform well compared to other operators~\cite{grover2016node2vec,tsitsulin2018verse}.}, etc. Finally, we choose 50\% of the edges to train the model and the rest of the edges is used for testing.

\begin{table}[!htb]
    \centering
    \begin{tabular}{c|c|c}
        Operator & Notation & Definition \\ \hline
        Hadamard & $\boxdot$ & $z_i\boxdot z_j= z_i \textasteriskcentered z_j$ \\
        Weighted-L1 & $\parallel.\parallel_{1}$ & $\parallel z_i . z_j\parallel_{1} =  |z_i - z_j|$ \\
        Weighted-L2 & $\parallel.\parallel_{2}$ & $\parallel z_i . z_j\parallel_{2} =  |z_i - z_j|^2$ \\
    \end{tabular}
    \caption{Vector operators used for link prediction. These notational definitions are borrowed from \emph{node2vec}~\cite{grover2016node2vec}.}
    \label{tab:linknotations}
\end{table}

\begin{figure*}[!htb]
    \centering
    \includegraphics[width=0.32\linewidth,height=4cm]{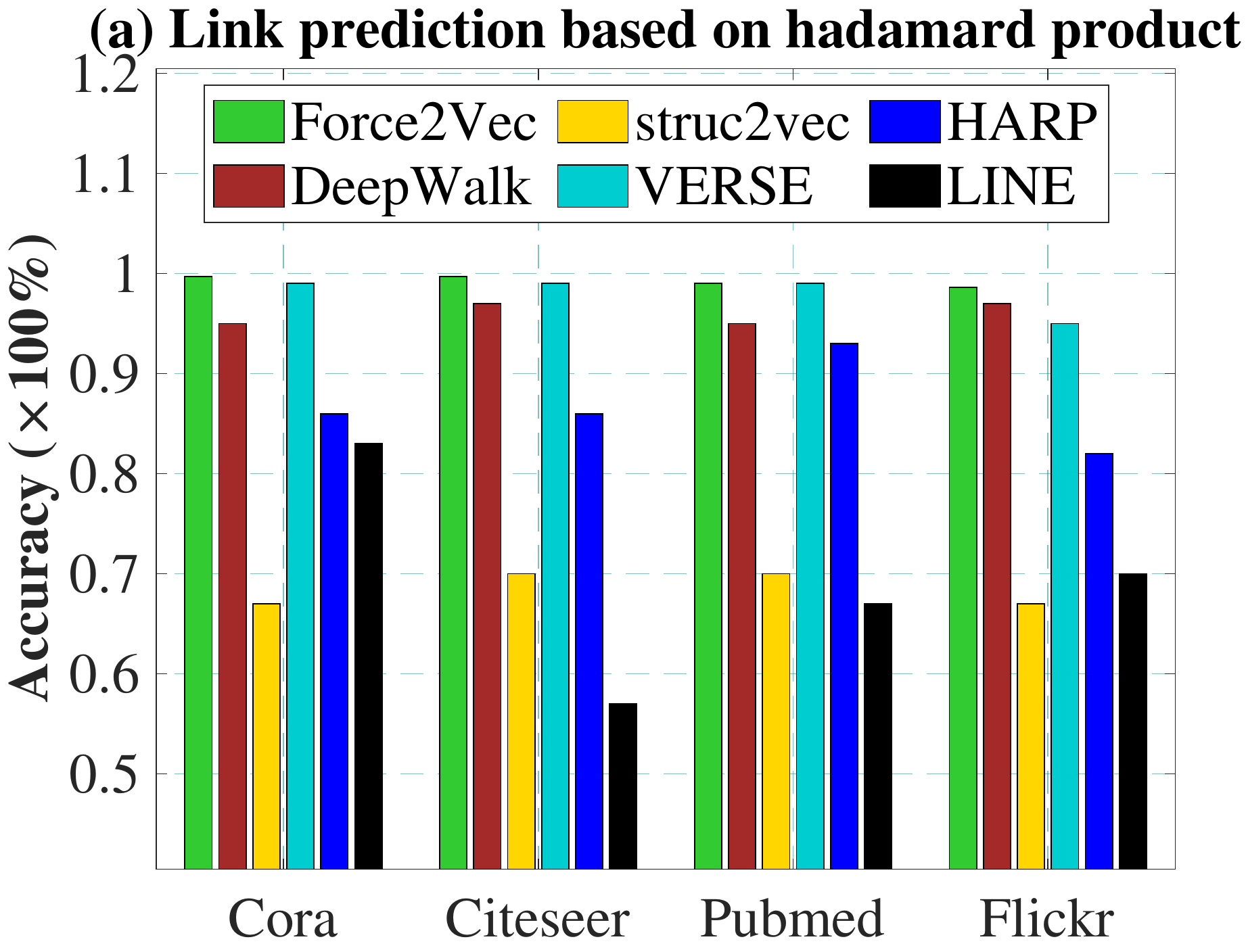}
    \includegraphics[width=0.32\linewidth,height=4cm]{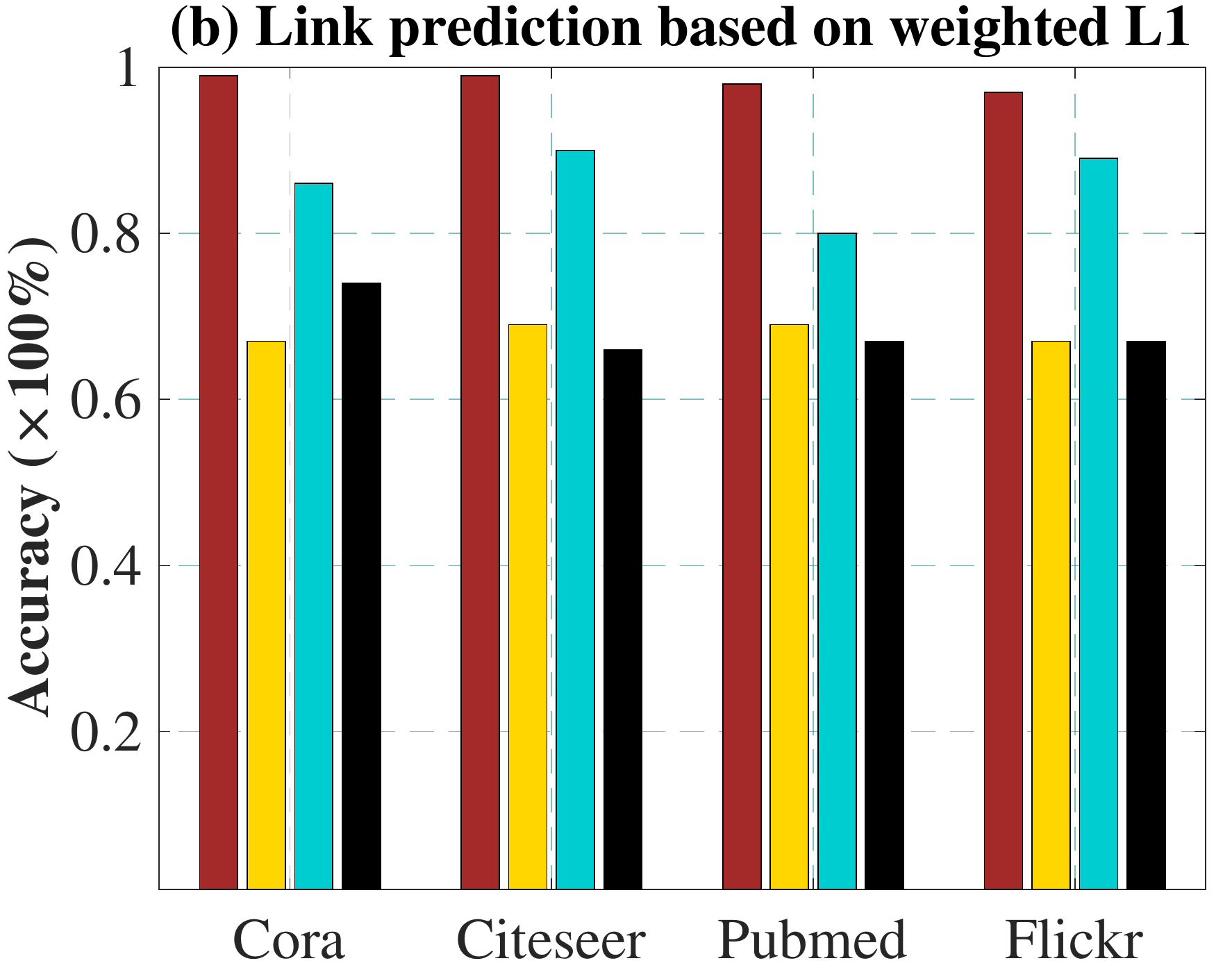}
    \includegraphics[width=0.32\linewidth,height=4cm]{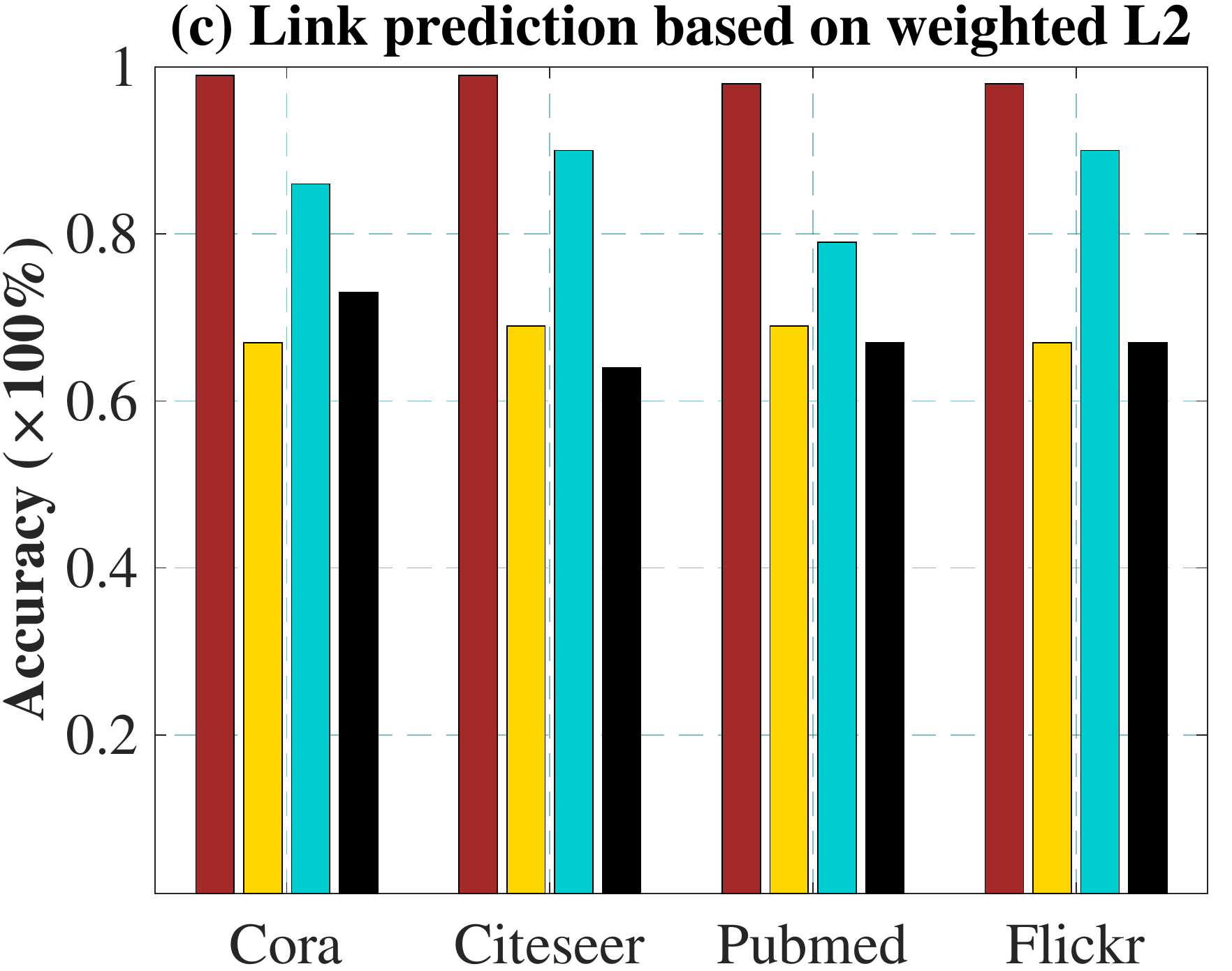}
    \caption{(a) Link prediction based on Hadamard vector operator. (b) Link prediction based on weighted-L1 based vector operator. (c) Link prediction based on weighted-L2 based vector operator.}
    \label{fig:abclinkprediction}
\end{figure*}

We show notational definitions of three vector operators in Table \ref{tab:linknotations}. All these operators have been used previously for link prediction tasks \cite{grover2016node2vec}. We show the experimental results of unsupervised methods on this task in Figs. \ref{fig:abclinkprediction} (a), \ref{fig:abclinkprediction} (b), and \ref{fig:abclinkprediction} (c). As link prediction is a binary classification problem, we only report accuracy. We can see while using the Hadamard operator, Force2Vec and VERSE show competitive performance and outperform other methods. They achieve almost 99\% accuracy in all datasets. For weighted L1 and L2 operators, DeepWalk performs better than other methods. For all cases, struc2vec is the worst performer as this tool is mainly designed to capture structural equivalence in the network. We also conducted experiments for link prediction task using HOPE and RolX. We observe that HOPE achieves accuracy of 79.8\% and 80\% for the Cora and the Pubmed datasets, respectively, using the Hadamard vector operator. On the other hand, RolX achieves accuracy of 79.7\% and 74.8\% for the Cora and the Pubmed datasets, respectively, using the Hadamard vector operator. These values of accuracy are lower than that of Force2Vec, VERSE and DeepWalk.

\subsubsection{Clustering}
The clustering of vertices is an important task in graph mining where common/similar vertices tend to form a cluster. High quality embedding can be helpful to detect a community in large scale social networks. A good graph clustering has a higher number of intra-cluster edges and a lower number of inter-cluster edges.  Generally, the Louvain algorithm \cite{blondel2008fast} is widely used to find clusters in a graph which focuses on maximizing \emph{modularity}. However, we can not apply it on embedding as we do not have any structural information about the graph. Instead, we apply the $k$-means\footnote{\url{https://en.wikipedia.org/wiki/K-means_clustering}} algorithm which can effectively detect clusters in the embedding space of the graph. The common practice is to set a value for $k$ in a range and find the clusters that show the highest \emph{modularity} score \cite{tsitsulin2018verse}.

\begin{figure}[!htb]
    \centering
    \includegraphics[width=0.55\linewidth,height=4cm]{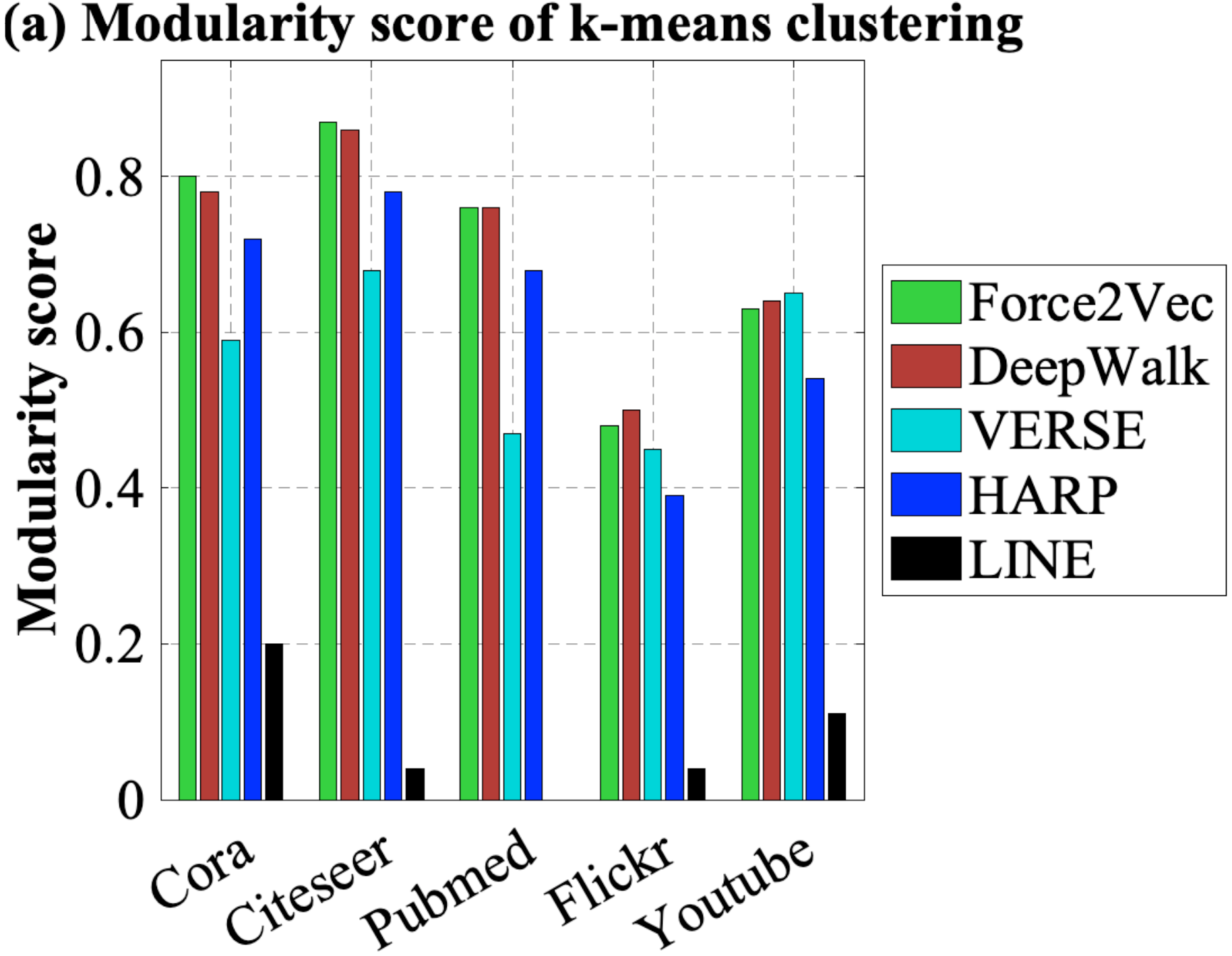}
    \includegraphics[width=0.435\linewidth,height=4cm]{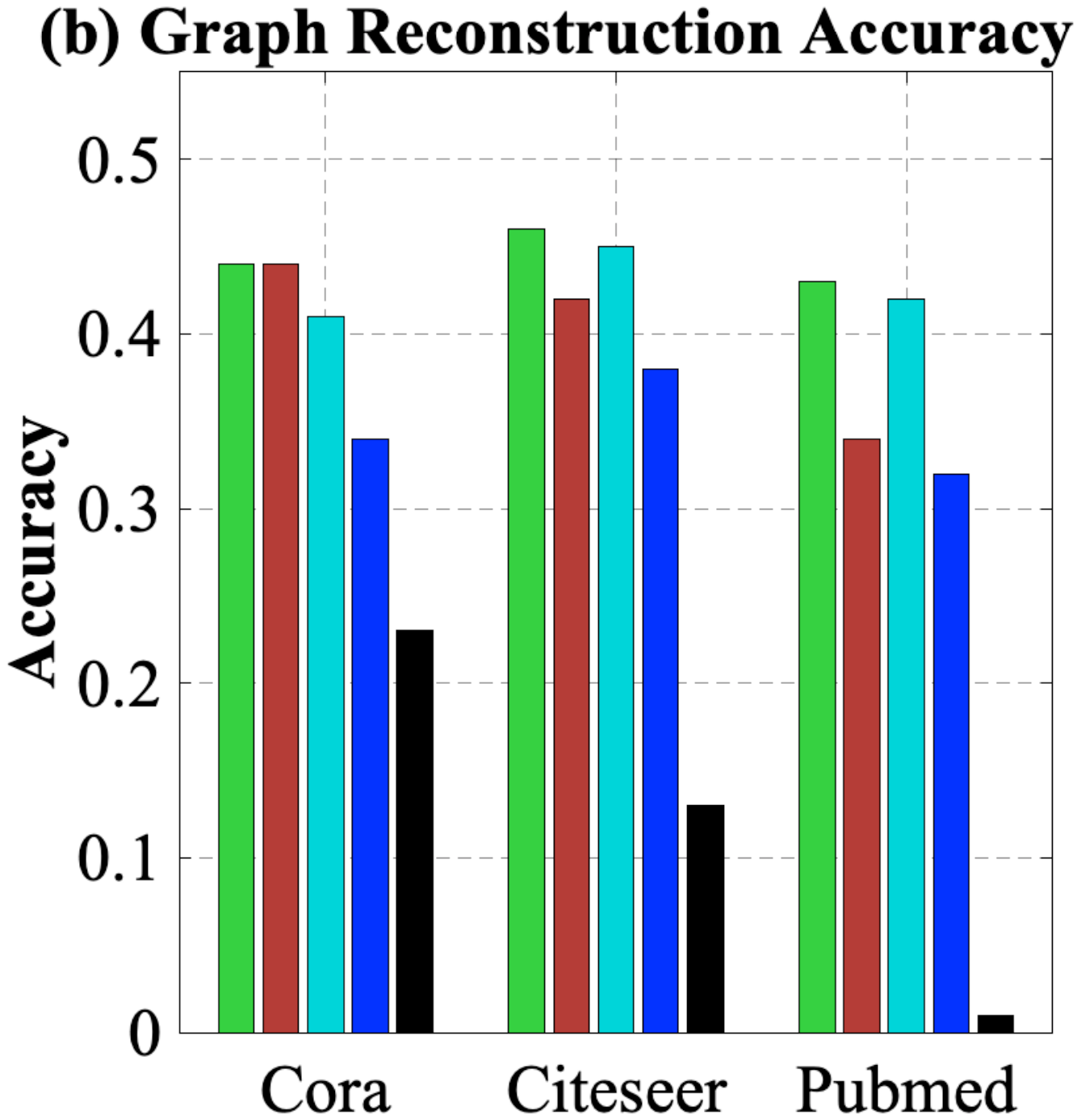}
    \caption{(a) Modularity scores of different methods for different benchmark datasets. (b) Results of graph reconstruction accuracy for different methods.}
    \label{fig:abclusterreconstruction}
\end{figure}

\begin{figure*}[!htb]
    \centering
    \begin{subfigure}{.23\textwidth}
  \centering
  \fbox{\includegraphics[width=0.95\linewidth,height=3.5cm]{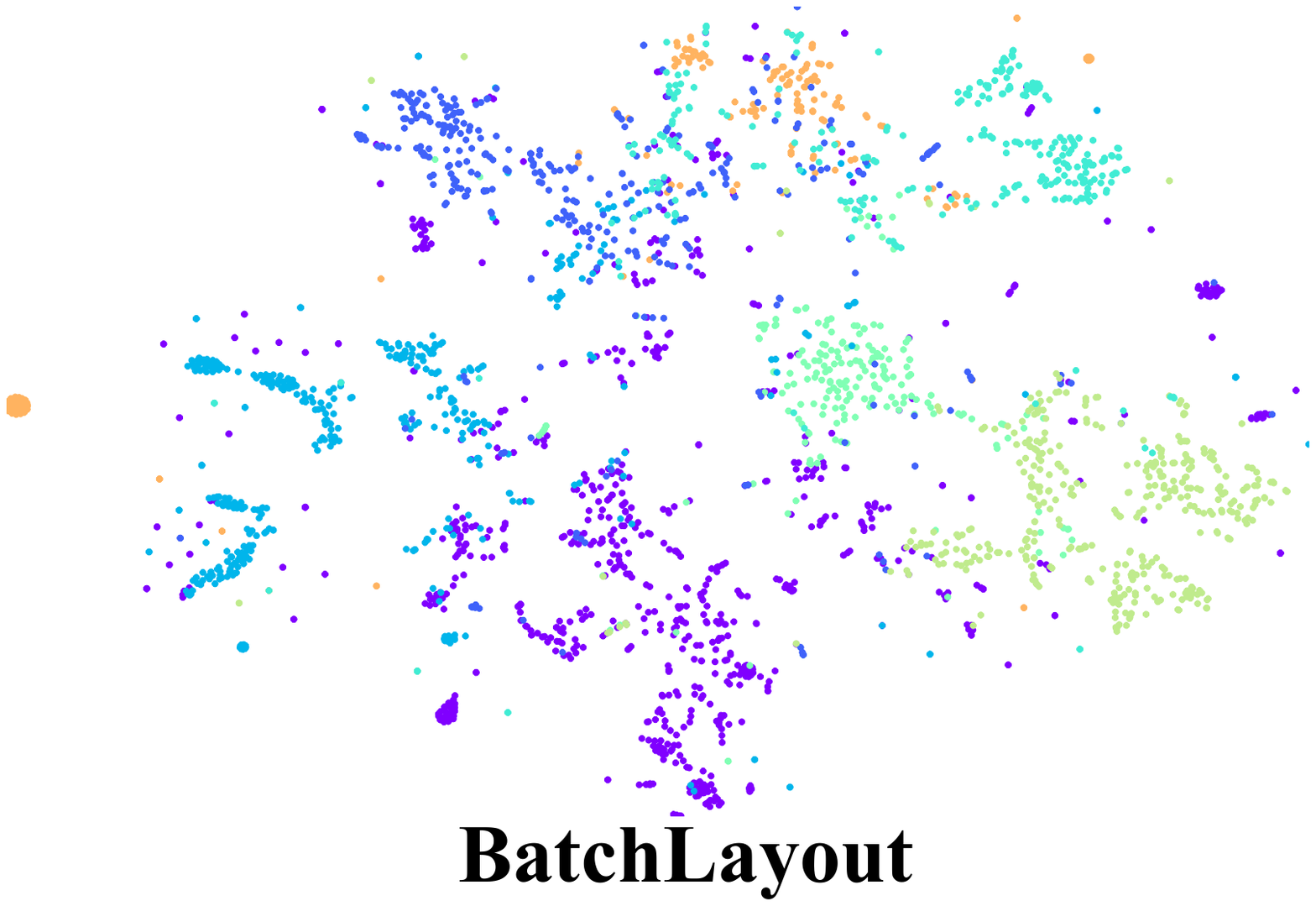}} 
  \caption{Force2Vec}
  \label{fig:coraf2v_tsne}
\end{subfigure}
\begin{subfigure}{.23\textwidth}
  \centering
   \fbox{\includegraphics[width=0.95\linewidth,height=3.5cm]{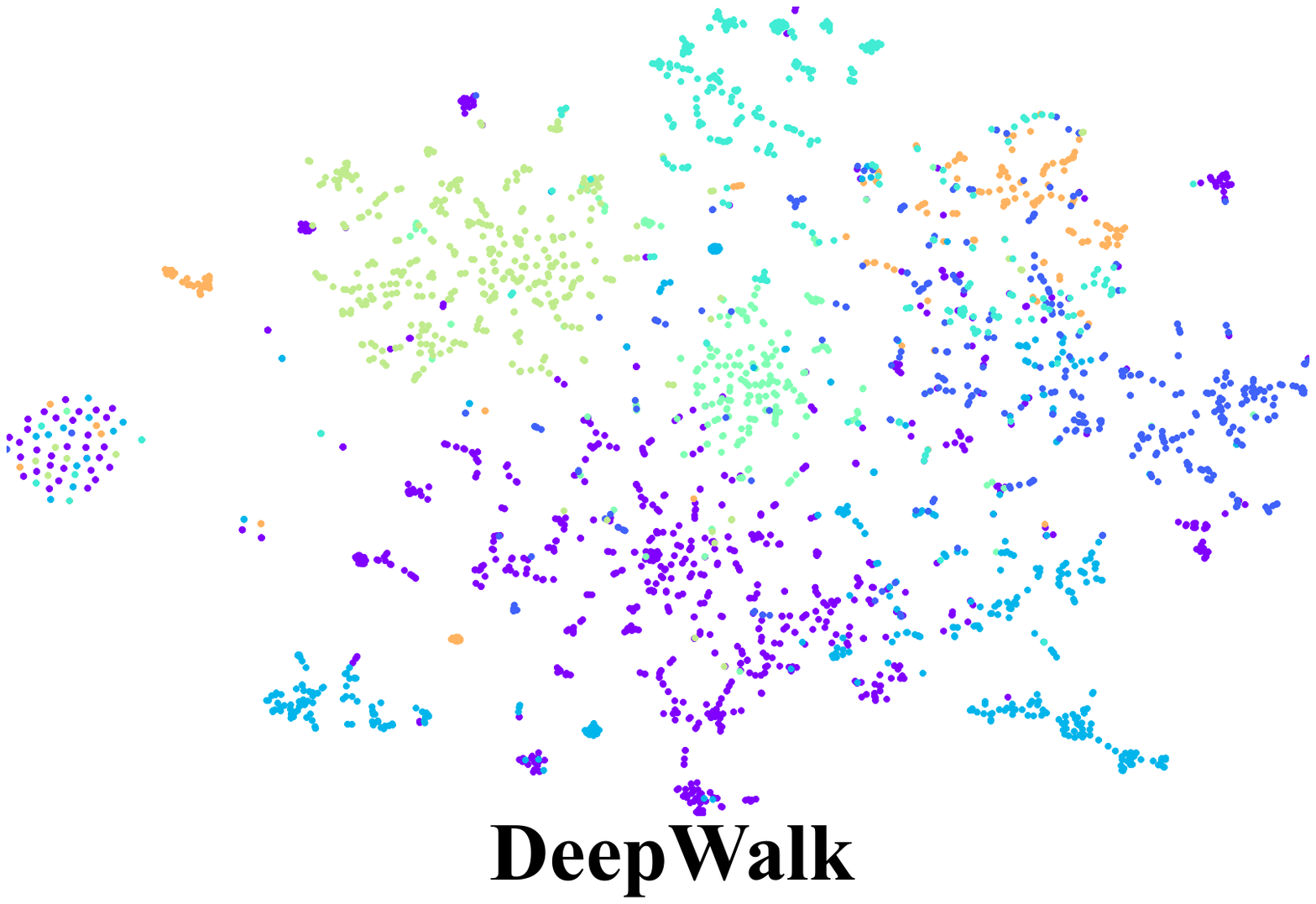}}
  \caption{DeepWalk}
  \label{fig:coradeepwalk_tsne}
\end{subfigure}
   \begin{subfigure}{.23\textwidth}
  \centering
    \fbox{\includegraphics[width=0.95\linewidth,height=3.5cm]{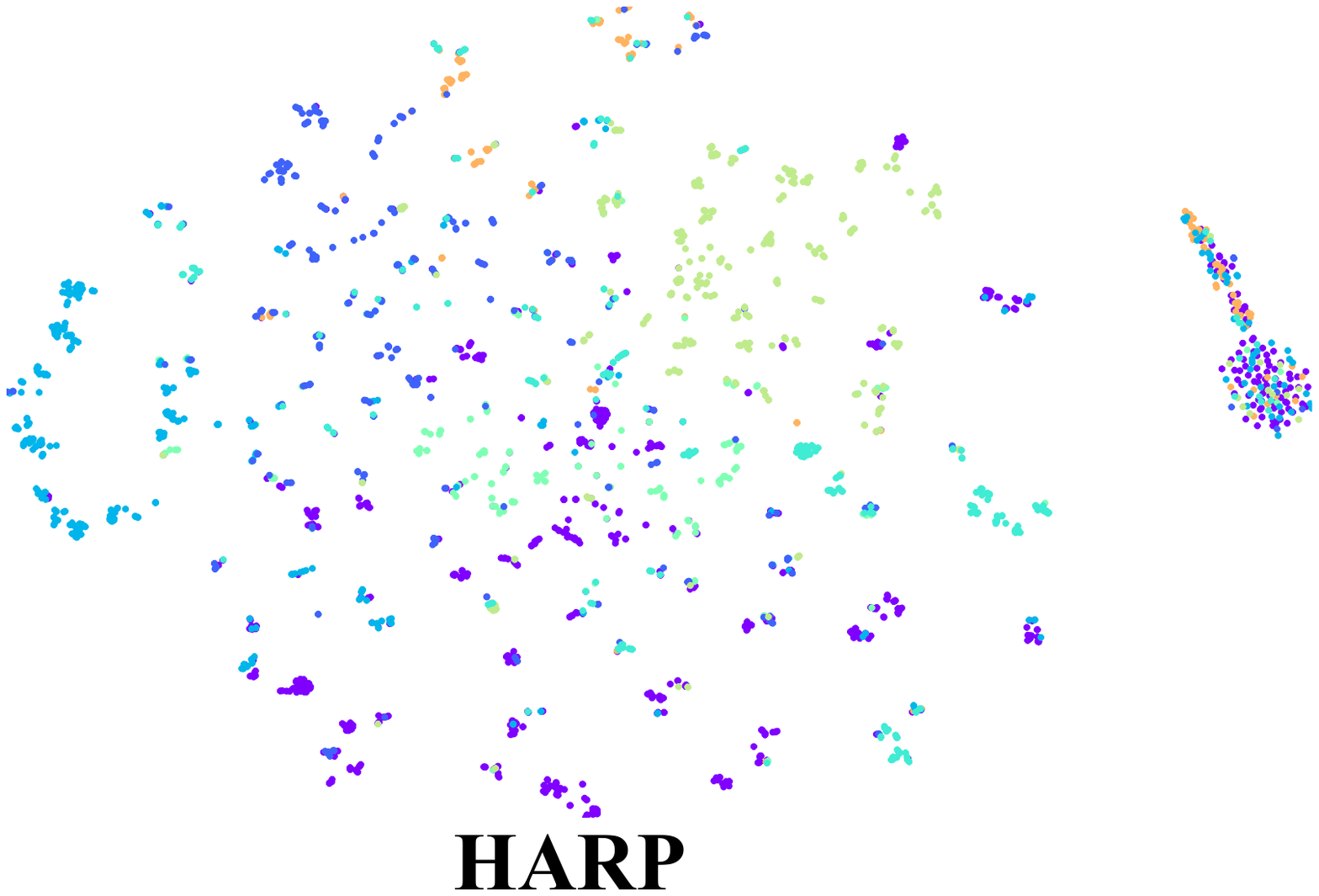}}
  \caption{HARP}
  \label{fig:coraharp_tsne}
\end{subfigure}
\begin{subfigure}{.23\textwidth}
  \centering
    \fbox{\includegraphics[width=0.95\linewidth,height=3.5cm]{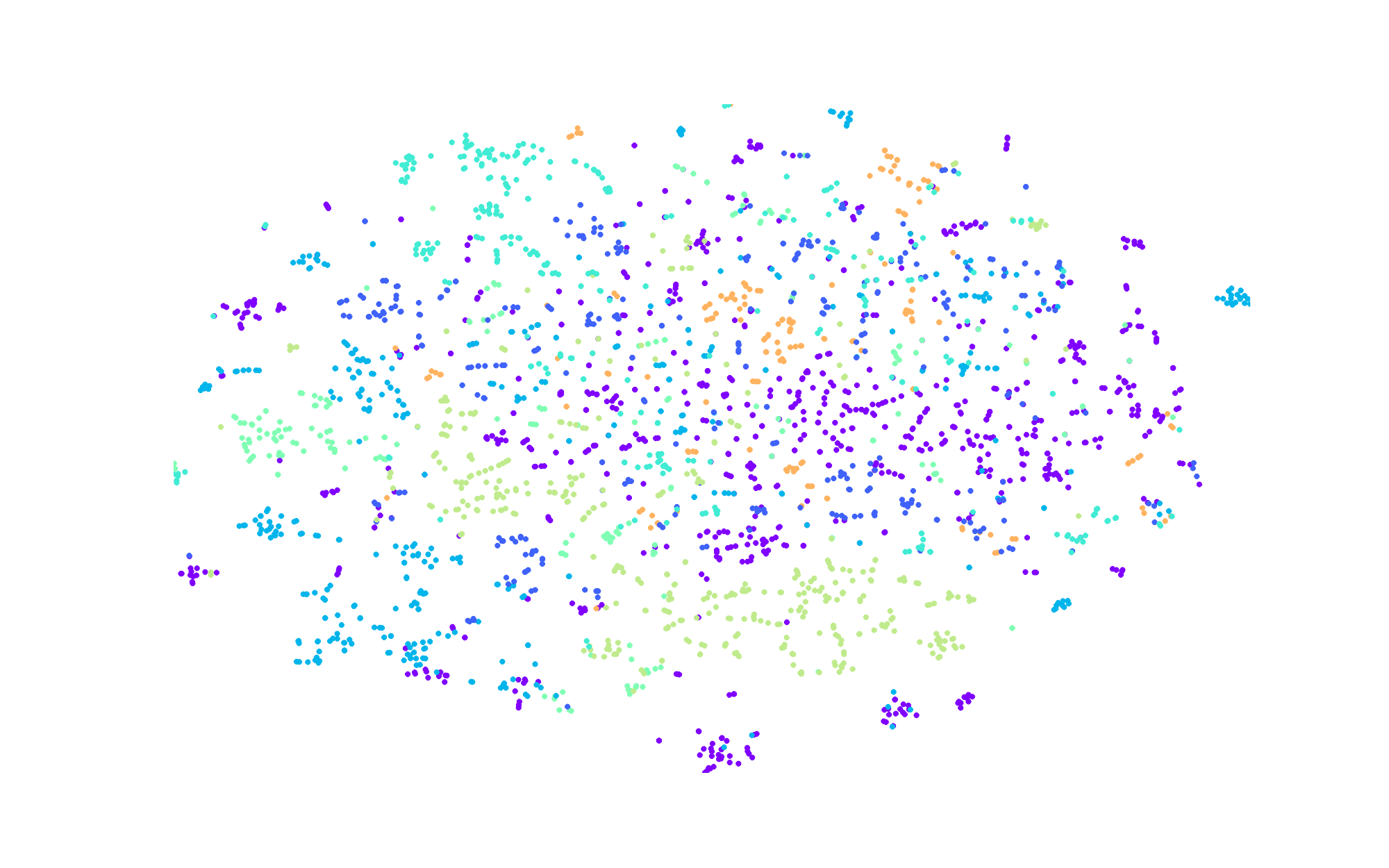}}
  \caption{VERSE}
  \label{fig:coraverse_tsne}
\end{subfigure}
    \caption{2D visualization of Cora dataset applying t-SNE tool on embeddings generated by (a) Force2Vec, (b) DeepWalk, (c) HARP, and (d) VERSE. Different colors represents different class labels of vertices.}
    \label{fig:coravis}
\end{figure*}
\begin{figure*}[!htb]
    \centering
    \begin{subfigure}{.23\textwidth}
  \centering
  \fbox{\includegraphics[width=0.95\linewidth,height=3.5cm]{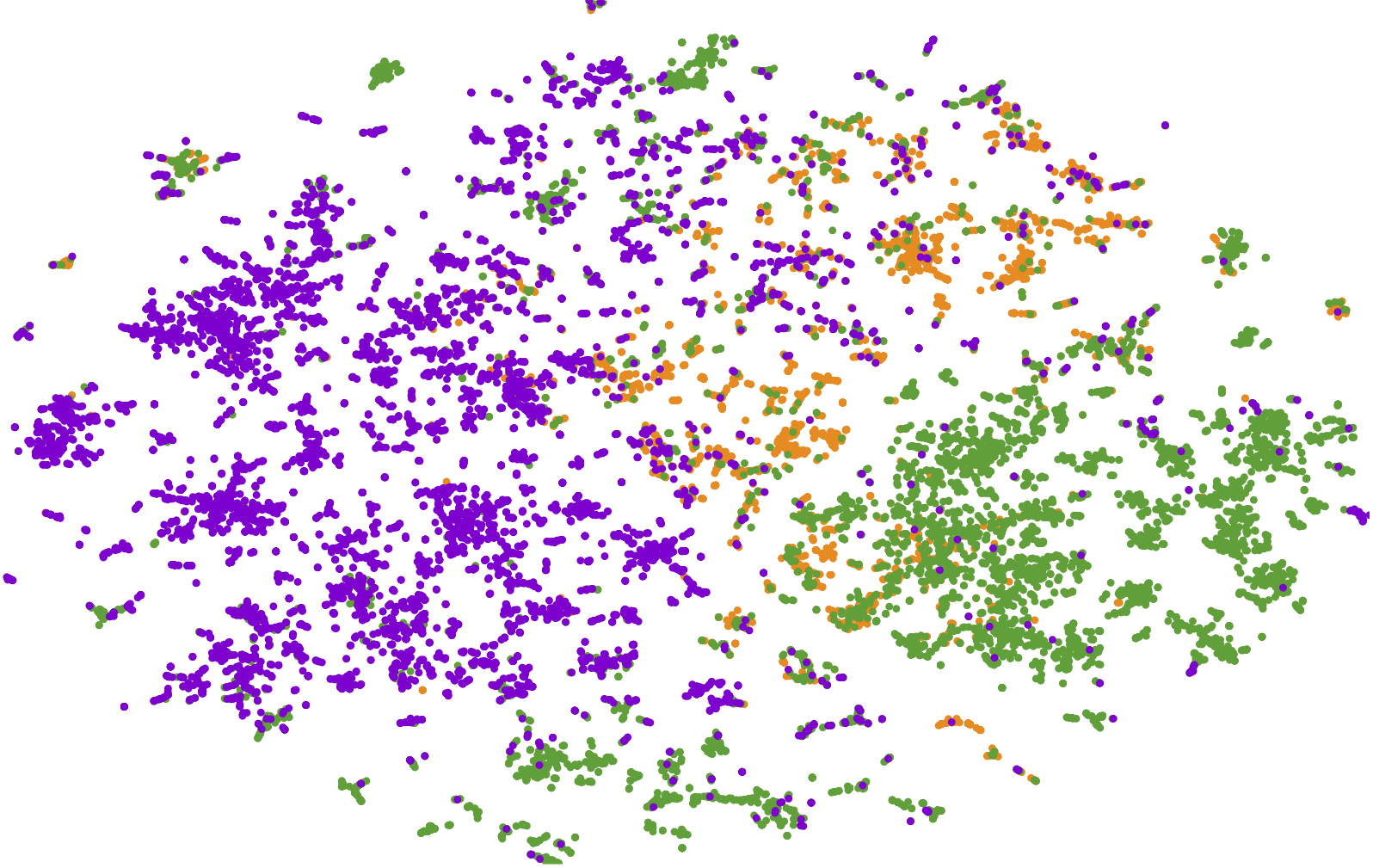}} 
  \caption{Force2Vec}
  \label{fig:f2v_tsne}
\end{subfigure}
\begin{subfigure}{.23\textwidth}
  \centering
   \fbox{\includegraphics[width=0.95\linewidth,height=3.5cm]{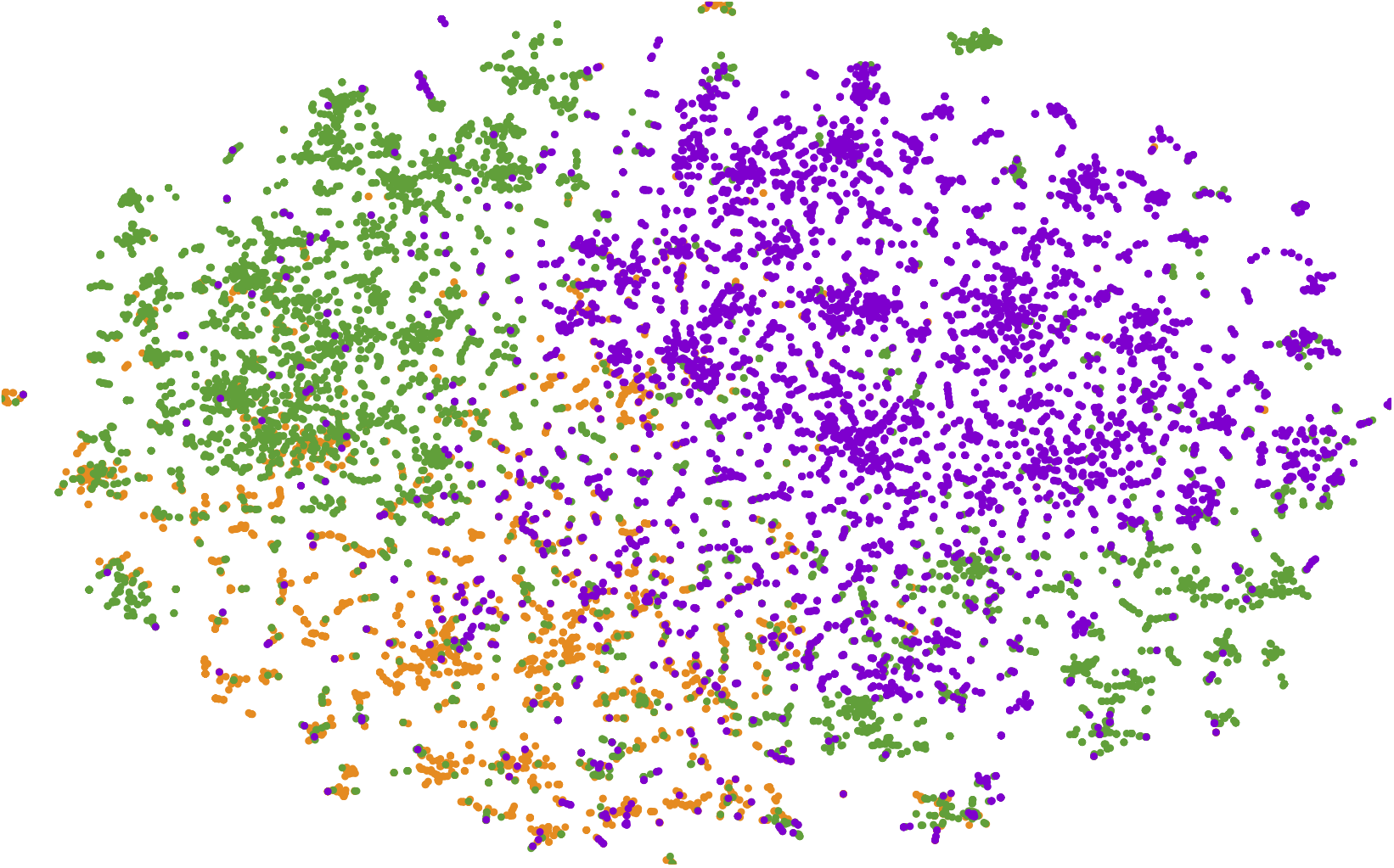}}
  \caption{DeepWalk}
  \label{fig:deepwalk_tsne}
\end{subfigure}
   \begin{subfigure}{.23\textwidth}
  \centering
    \fbox{\includegraphics[width=0.95\linewidth,height=3.5cm]{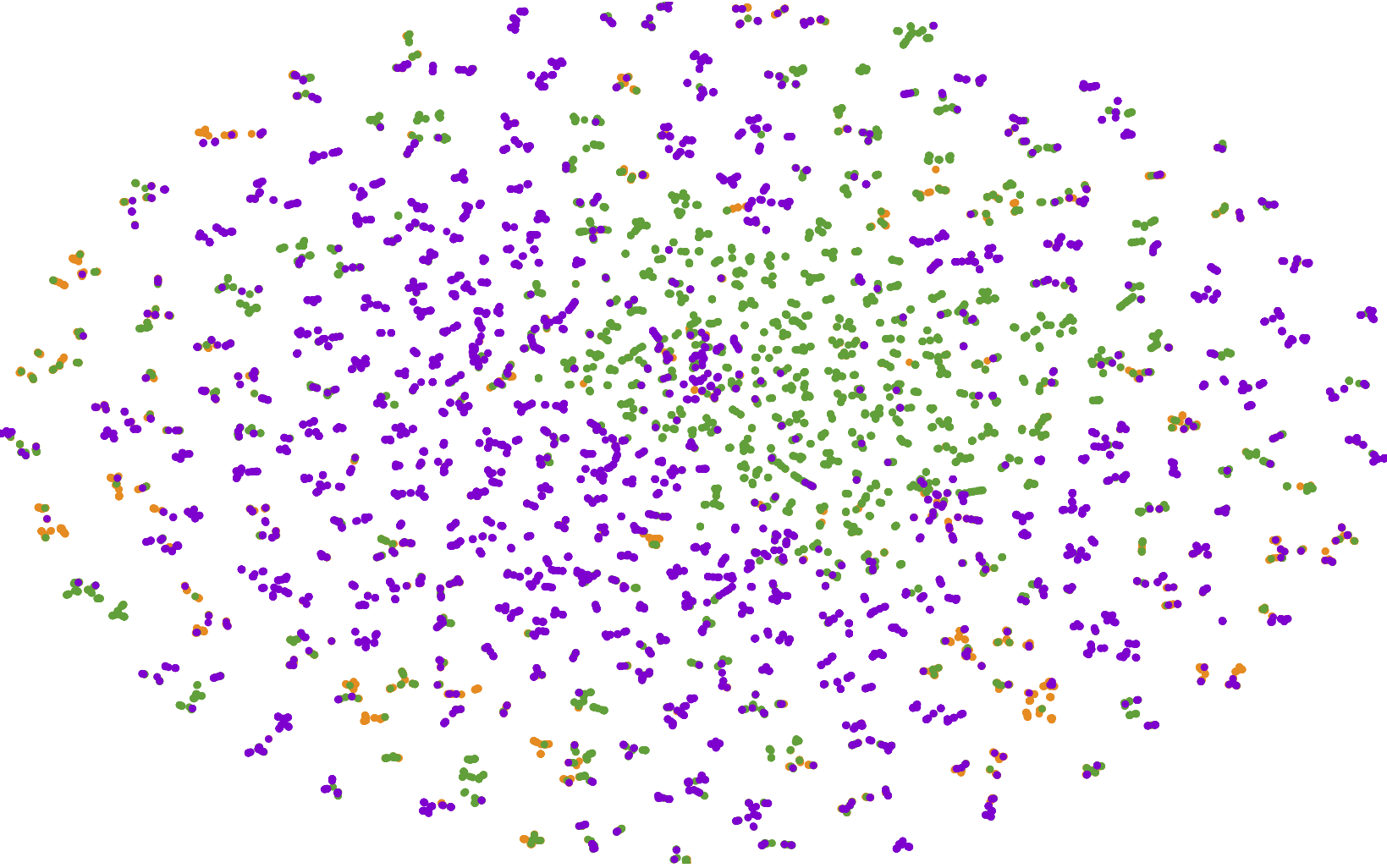}}
  \caption{HARP}
  \label{fig:harp_tsne}
\end{subfigure}
\begin{subfigure}{.23\textwidth}
  \centering
    \fbox{\includegraphics[width=0.95\linewidth,height=3.5cm]{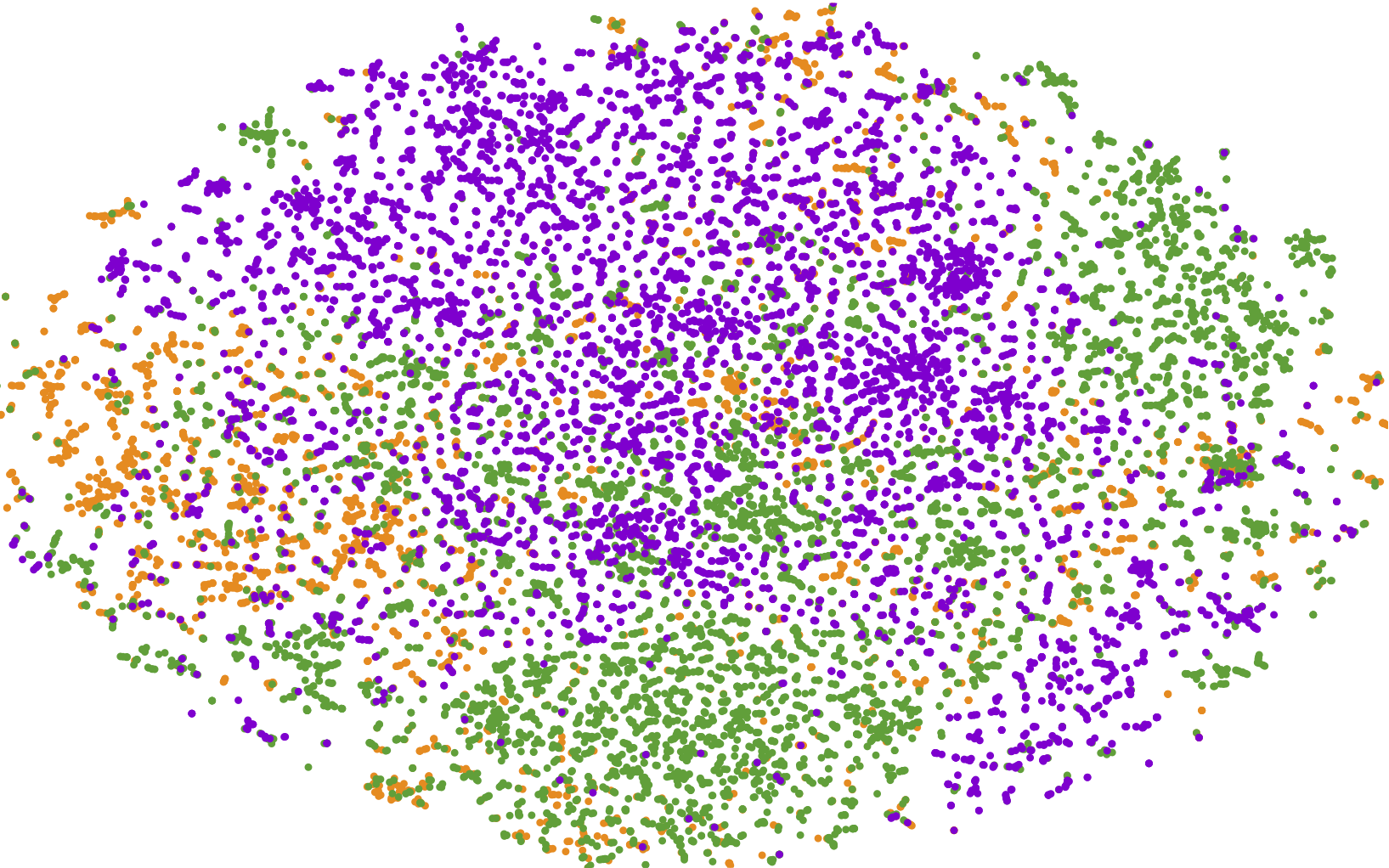}}
  \caption{VERSE}
  \label{fig:verse_tsne}
\end{subfigure}
    \caption{2D visualization of Pubmed dataset applying t-SNE tool on embeddings generated by (a) Force2Vec, (b) DeepWalk, (c) HARP, and (d) VERSE. Different colors represents different class labels of vertices.}
    \label{fig:pubmedvis}
\end{figure*}

We report modularity scores of Force2Vec, DeepWalk, VERSE, HARP and LINE across different datasets in Fig. \ref{fig:abclusterreconstruction} (a). We consider Louvain algorithm as the baseline method. It has the modularity scores of 0.81, 0.88, 0.73, and 0.49 for Cora, Citeseer, Pubmed and Flickr datasets, respectively. We observe that Force2Vec achieves higher modulartiy scores for Cora, Citeseer, and Pubmed datasets. \deepwalk{} achieves higher modularity scores than other methods for Flickr dataset and VERSE shows better modularity score for the Youtube dataset. Notably these results are very competitive to the baseline method. For large  graphs, such as Flickr and Youtube, the modularity scores of Force2Vec, VERSE and DeepWalk are comparative. As the LINE from authors' repository shows poor performance as usual, we run another implementation\footnote{\url{https://github.com/shenweichen/GraphEmbedding}}. This version of LINE shows better performance compared to the previous version. In particular, this version of LINE achieves the modularity scores of 0.61,	0.51, and 0.51 for the Cora, Citeseer, and Pubmed datasets, respectively. Note that we do not show the results of other methods such as struc2vec or HOPE due to their poor performance in graph clustering task using the $k$-means algorithm.

\subsubsection{Graph Reconstruction}
The graph reconstruction task is related to the construction of a graph from the generated embedding such that it can create edges as of the original graph with high accuracy. If a method generates a high quality embedding, then it can accurately reconstruct the original graph. However, it requires $\frac{n(n-1)}{2}$ comparisons to find the nearest neighbors for a graph of $n$ vertices which is highly expensive for large graphs. Thus, we select 1000 vertices randomly to reconstruct a sub-graph and check the accuracy of finding neighborhoods. Generally, the degree of each vertex is used to find the nearest neighborhoods and these are compared with the neighbors of that vertex in the original graph. We compute the accuracy to explore how many vertices can accurately capture their neighborhoods based on their consine similarity\footnote{\url{https://en.wikipedia.org/wiki/Cosine_similarity}} scores.

We report the results of the graph reconstruction accuracy in Fig. \ref{fig:abclusterreconstruction} (b). We observe that both Force2Vec and VERSE show similar performance on the graph re-construction task. Specifically, Force2Vec (sigmoid) achieves a higher reconstruction accuracy than DeepWalk on the Citeseer and Pubmed datasets. LINE and HARP, on the other hand, show very poor reconstruction ability on the Pubmed dataset. Notice that this task has an inherent relation with link prediction. Here, we find the top-most degree number of neighborhoods based on their cosine similarity and the methods that perform well for the link prediction task, as reported in Section \ref{sec:linkpred}, also show competitive performance for graph reconstruction task.

\subsection{Visualization}
Graph visualization is an enriched research field which has diversified applications in data mining and bioinformatics \cite{de2003visual}. The common practice of graph visualization is to generate two dimensional coordinates for each vertex of the graph and then plot in the Cartesian coordinate system connecting all the edges. A plethora of studies in graph drawing community have focused on different types of network visualization \cite{xu2007graphscape,herman2000graph,rahman2020batchlayout}. However, when we generate an embedding, we basically create a  $d$-dimensional projection of a graph where graph visualization methods are not directly applicable. Thus, generic data visualization technique such as t-SNE \cite{maaten2008visualizing} or UMAP~\cite{mcinnes2018umap} are used to visualize the graphs in 2D using its embedding. This type of visualization can help to identify clusters or latent characteristics of the graph. To conduct this experiment, we use t-SNE by setting several of its hyper-parameters to default values and generate two dimensional coordinates for visualization in the Cartesian space.

In Fig. \ref{fig:coravis}, we show two dimensional visualizations for the Cora dataset applying t-SNE on 128 dimensional embeddings generated by different methods. We color the vertices of distinct class with distinct colors i.e., vertices of the same class have the same color. We observe that the visualization of the Force2Vec method preservers more local and global clustering structure. The DeepWalk method, on the other hand, also shows local clustering information but some vertices from different classes form cluster in the left side. The VERSE method also shows local clustering, but they are high in numbers and have scattered placement. HARP method shows similar visualization to the VERSE method. We also show two dimensional visualizations for the Pubmed dataset in Fig. \ref{fig:pubmedvis}. We observe similar results that the Force2Vec and DeepWalk methods show better visualizations than other methods. Force2Vec ($t$-distribution) can gain more information to separate clusters and has better local and global clustering information.

\subsection{Parameter Sensitivity}
\begin{figure}[!ht]
    \centering
    \includegraphics[width=0.49\linewidth,height=4.0cm]{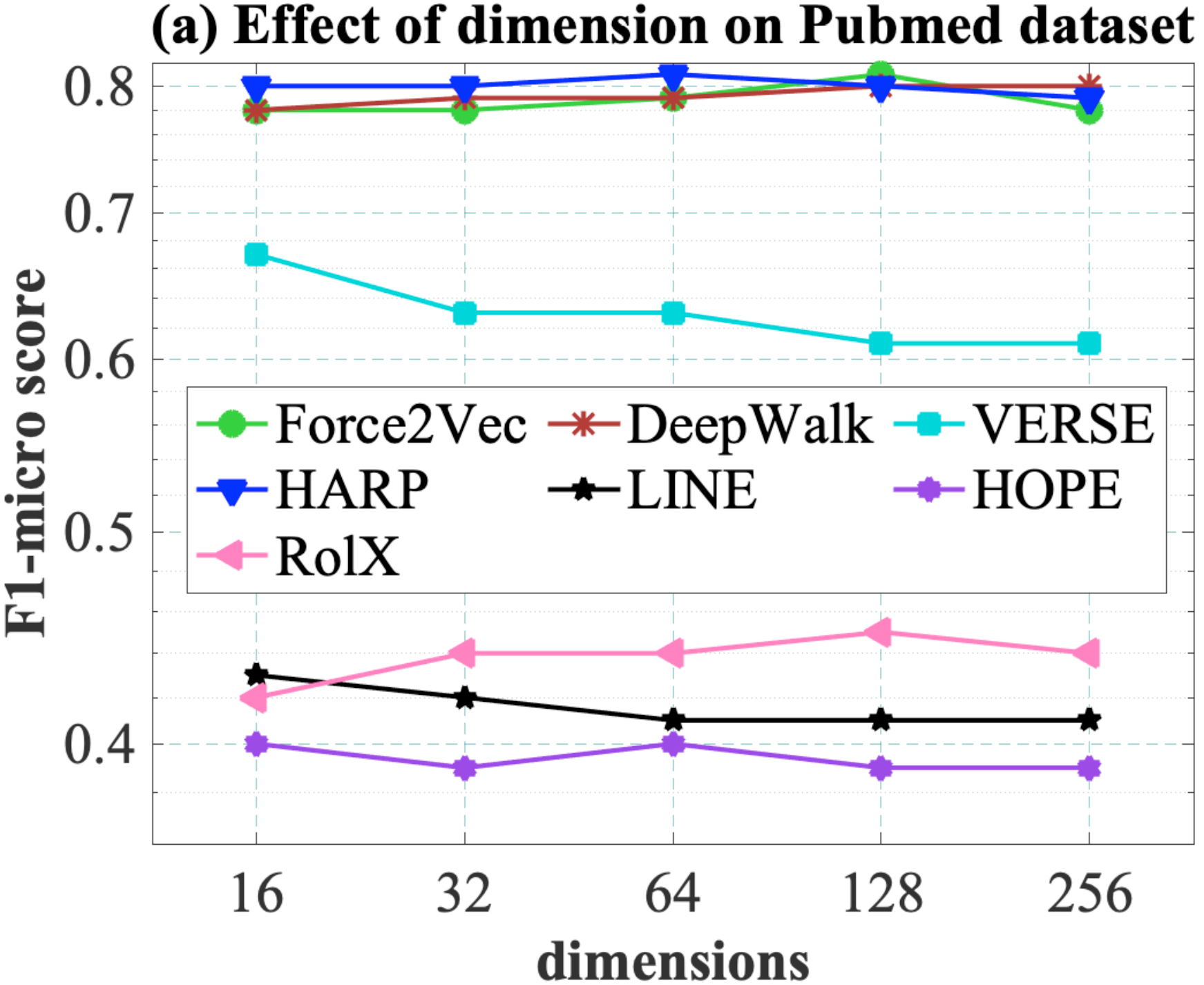}
    \includegraphics[width=0.49\linewidth,height=4.0cm]{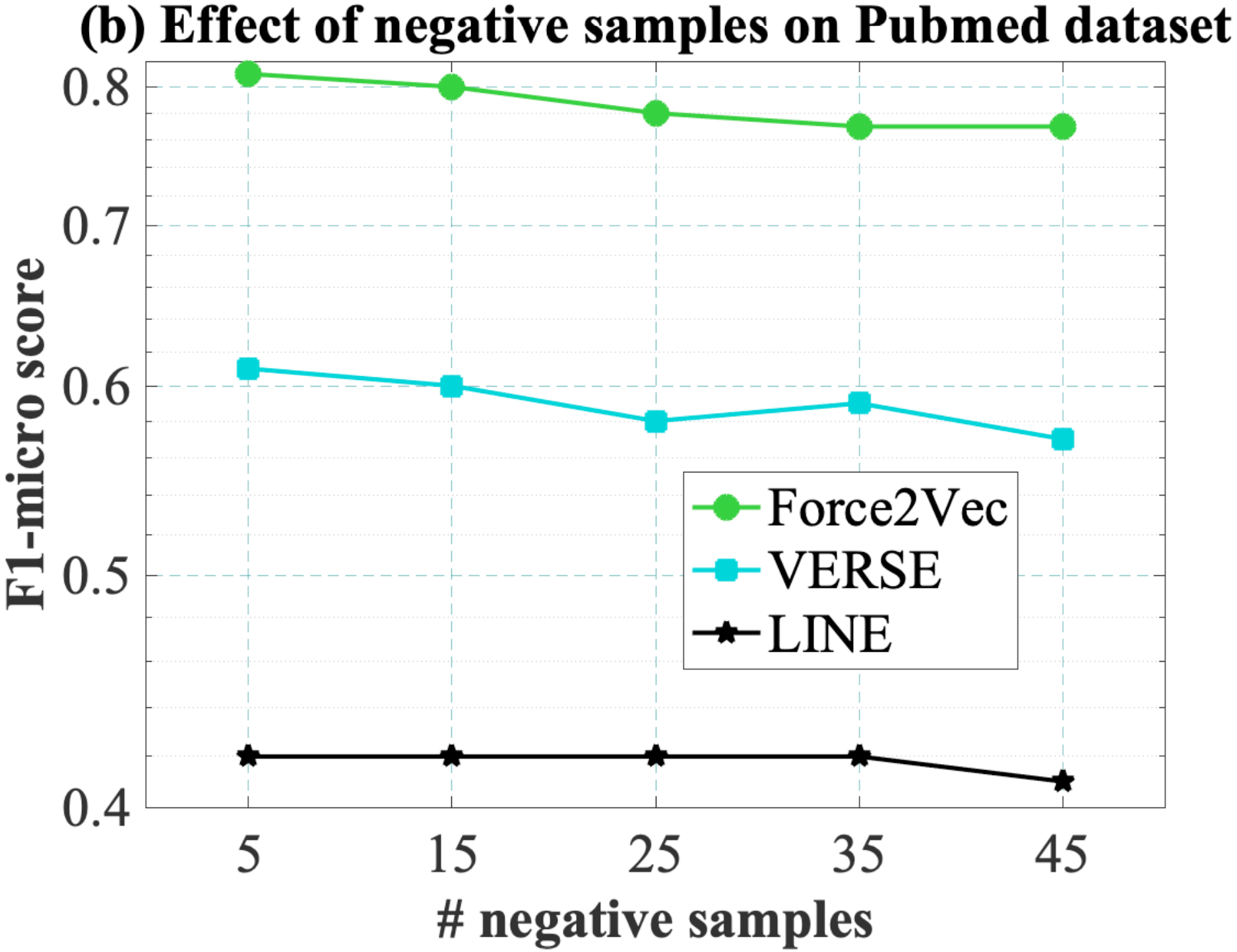}
    \caption{(a) Effect on classification score for Pubmed dataset using various dimensional embedding. (b) Effect of different number of negative samples on Cora and Pubmed datasets.}
    \label{fig:abcdimension}
\end{figure}

\subsubsection{Effect of Dimensions} Some previous studies have shown that the performance on the prediction task may vary if we choose different values for hyper-parameters~\cite{perozzi2014deepwalk,grover2016node2vec,tsitsulin2018verse,rahman2020training}. For example, after reaching a certain value for dimensionality, the accuracy of prediction starts to drop when we increase it further. Most of the previous studies suggest using dimensional embedding. To summarize the results, we conduct experiments varying the dimensions of the output embedding for some shallow network-based methods. We set different parameters as described in Section \ref{sec:expsetup} and take 20\% of the dataset to train the logistic regression model while the rest of the samples in the dataset are used for the classification. We report the results of the F1-micro scores for the Pubmed dataset in Figs. \ref{fig:abcdimension} (a). We observe that Force2Vec, DeepWalk, and HARP perform better than other methods for various dimensional embedding. We also notice that, for lower dimensions, the F1-micro scores are not that much less compared to higher dimensions. In fact, the VERSE tool shows better performance for 16-dimensional embedding for the Pubmed dataset. RolX shows high sensitivity for different dimensions. It shows the lowest performance for 16-dimensional embedding. Then, with the increase of dimension, the F1-micro score also increases until 128-dimension. Then, it falls a little for 256-dimensional embedding. The LINE method shows similar sensitivity to the VERSE method though its F1-micro scores are lower than the VERSE.

\subsubsection{Effect of Various Negative Samples}
Noise-contrastive estimation~\cite{gutmann2010noise} is a popular technique used by most of the shallow graph embedding models~\cite{tsitsulin2018verse,zhang2018sine,rozemberczki2019gemsec}. Using this technique, a subset of vertices are randomly selected from a uniform distribution as negative samples which are used alongside positive samples (i.e., $k$-hop neighbors) to optimize the objective function. However, randomly generated negative samples may hurt the optimization function due to the selection of false negative samples and this become vital if the number of negative samples is very large. \citeauthorandyear{armandpour2019robust} have made efforts to analyze this issue theoretically and proposed robust negative sampling techniques for graph embedding problem. NSCaching~\cite{zhang2019nscaching} is another interesting work that generates efficient negative samples for knowledge graphs. In Fig. \ref{fig:abcdimension} (b), we empirically show the effect on performance measures varying the number of negative samples. To conduct experiment, we choose Force2Vec, VERSE and LINE methods as they support random negative sampling approach. We take 25\% of the vertices in the training set and report the F1-micro score for the rest of the testing dataset. We observe that performance score drops when we use more than 5 negative samples for the Pubmed dataset and continues to decrease for more negative samples. VERSE method is more sensitive to higher negative samples than others as its performance score significantly drops after using more than 15 negative samples for the Pubmed dataset. Note that the size of this dataset is relatively small as it has only around 19K vertices. For smaller graphs than Pubmed, the number of randomly selected negative samples will show more sensitivity than bigger graphs than the Pubmed dataset. The reason is that for bigger graphs, the probability of selecting false negative is lower compared to the smaller graphs. Thus, care must be taken to choose an effective number of negative samples rather than random selection.

\begin{figure}[!htb]
    \centering
    \includegraphics[width=0.49\linewidth,height=4.0cm]{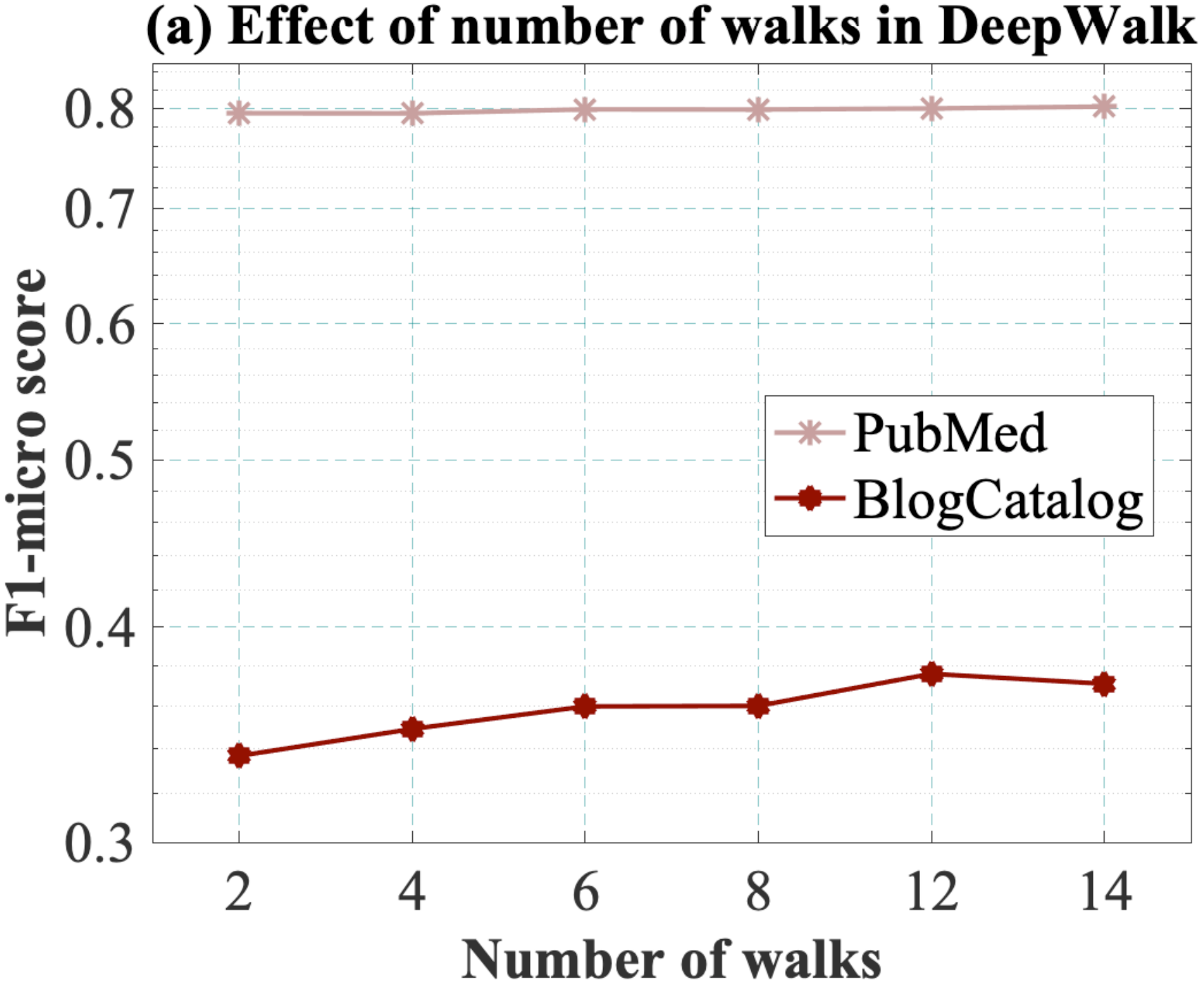}
    \includegraphics[width=0.49\linewidth,height=4.0cm]{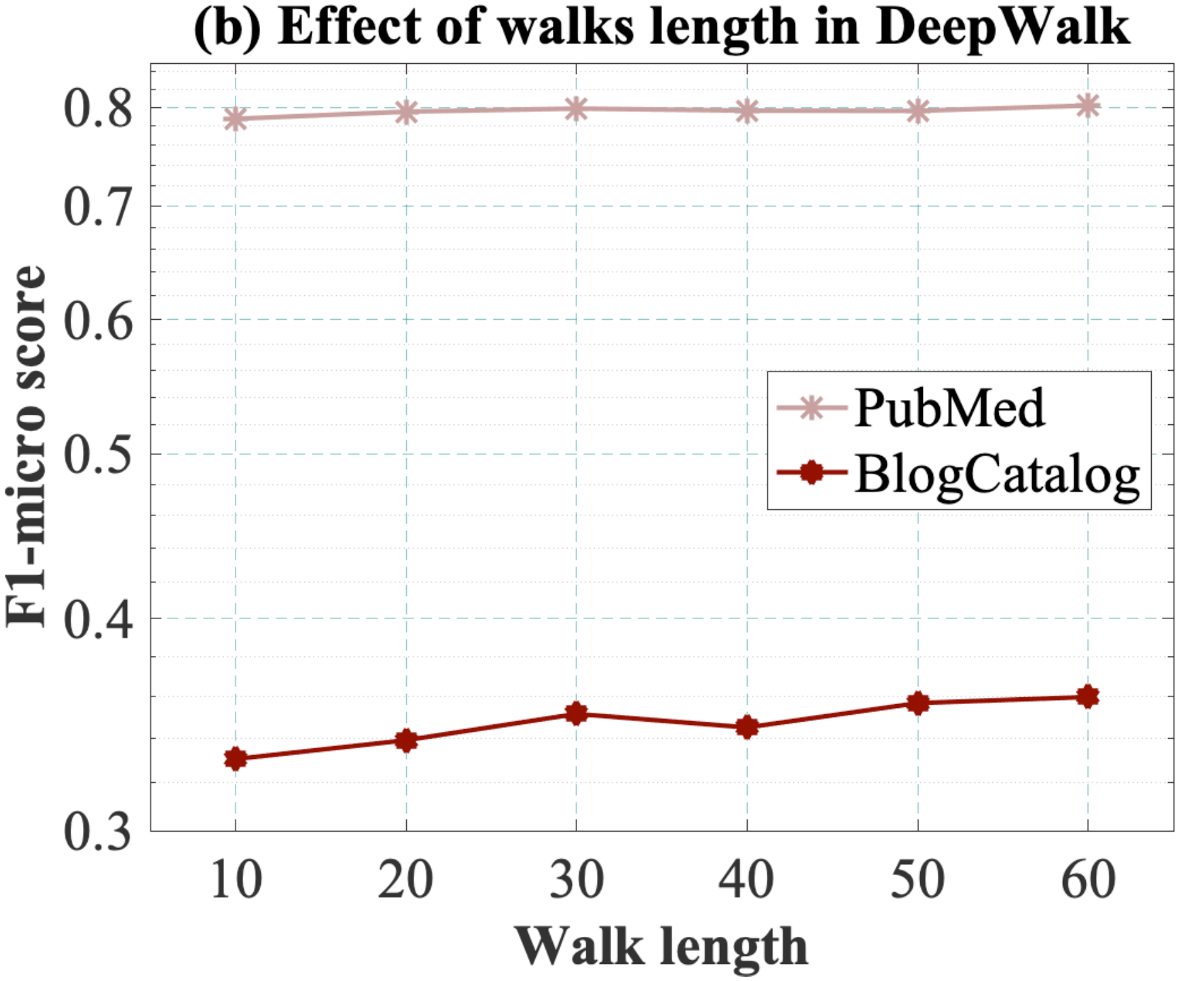}
    \caption{(a) Effect of the number of walks and (b) Effect of the walk length in DeepWalk for PubMed and BlogCatalog datasets.}
    \label{fig:deepwalkparameters}
\end{figure}

\subsubsection{Effect of the Number of Walks and Walk-length}
The random-walk based method such as DeepWalk, shows sensitivity on the number of walks as well as the length of the walk. We analyze these aspects using DeepWalk and report the results on Figs. \ref{fig:deepwalkparameters} (a) and (b), respectively for Pubmed and Blogcatalog graphs. In Fig. \ref{fig:deepwalkparameters} (a), we observe that DeepWalk shows less sensitivity for the number of walks on the Pubmed dataset for a fixed walk-length of 40 since this is a homogeneous network. However, for the heterogenous network Blogcatalog, it shows reasonable sensitivity i.e., F1-micro score increases with the increase of the number of walks. We also see similar sensitivity for different walk-length for a fixed 10 number of walks on Fig. \ref{fig:deepwalkparameters} i.e., Blogcatalog shows higher sensitivity for different length of walks since this is a heterogeneous network.

\section{Discussions and Limitations}
\label{sec:limitations}
\begin{table*}[!htb]
\centering
\begin{tabular}{|c|c|c|c|c|c|c|c|c|}
\hline
{\color[HTML]{000000} \textbf{Characteristics}}                                                              &  \textbf{RolX} &  \textbf{HOPE} &  \textbf{DeepWalk} &  \textbf{struc2vec} &  \textbf{LINE} & \textbf{HARP} & \textbf{VERSE} & \textbf{Force2Vec} \\ \hline
{\color[HTML]{FE0000} \begin{tabular}[c]{@{}c@{}}Memory \\ consumption\end{tabular}} & M                                    & VH                                   & H                                         & M                                        & H                                     & H                                    & L                                   & L                                                                                \\ \hline
{\color[HTML]{FE0000} Runtime}                                                       & H                                    & L                                    & VH                                        & VH                                       & H$^*$                                     & H                                   & L                                   & L                                                                               \\ \hline
{\color[HTML]{FE0000}\begin{tabular}[c]{@{}c@{}}Number of \\ Hyper-parameters\end{tabular}}                & L                                    & L                                    & H                                         & H                                        & M                                     & H                                    & M                                   & M                                                                              \\ \hline
{\color[HTML]{036400} Scalability}                                                   &          L                            & L                                   & L                                         & L                                        & H                                     & M                                    & H                                   & VH                                                                               \\ \hline
{\color[HTML]{036400} Robustness}                                                    & M                                    & L                                    & H                                         & L                                        & L                                     & H                                    & H                                   & H                                                                             \\ \hline
{\color[HTML]{036400} Performance}                                                   & L                                    & L                                    & H                                       & M/H                                        & M                                     & H                                    & H                                   & H                                                                              \\ \hline

\end{tabular}
\caption{Summary of graph embedding methods except GNN methods based on different characteristics. We score each tool from L, M, H, and VH which represent Low, Medium, High, and Very High, respectively. Note that L represents better score for memory consumption, running time and number of hyper-parameters whereas VH represents better score for scalability, robustness and performance.}
\label{tab:discussion}
\end{table*}

Graph representation learning is a challenging problem as it is hard to capture different latent characteristics of the original graph in the embedding space. If the output embedding can not capture the intrinsic properties well, then it performs poorly on several prediction tasks. A single tool can not generalize the embedding well for all prediction tasks. For example, struc2vec performs well for structural equivalence prediction whereas DeepWalk performs well for homophily prediction. In addition, we have shown in our experimental analyses that some methods may consume a high amount of memory but run fast, whereas some methods may take less memory but run slow. Some methods can generate a high quality embedding but have higher runtime and consume more memory. We have summarized the results of some embedding methods except GNN models in Table \ref{tab:discussion}, based on several characteristics. We have evaluated all of the tested methods in this survey scoring from L, M, H, and VH denoting Low, Medium, High, and Very High values, respectively. Note that we want low values (L) for memory consumption, running time and number of hyper-parameters. On the other hand, we want very high values (VH) representing a better score for scalability, robustness and performance. We put `M/H' for struc2vec in performance as it performs well for networks having structural equivalence properties, however shows moderate performance for networks having homophily characteristics. H* represents the variable running time for the LINE tool as it has a high running time for small networks but takes comparatively less time for large networks with respect to other methods. From our analyses, we can presume that an optimal embedding tool will have less runtime and memory consumption cost; furthermore, it will generate a high quality embedding that will show superior performance in various prediction tasks. In reality, there is always a trade-off between our expectation and available resources. From our analyses, we observe that the existing methods face the following challenges and limitations.

\begin{itemize}
    \item \textbf{Runtime, Memory Cost and Scalability:} Runtime and memory requirement are two bottlenecks of random-walk based embedding methods, though they show superior performance, e.g., DeepWalk achieves superior performance for most of the prediction tasks, though it consumes significant amount of memory and runs very slow. Besides these, most of the real-world social networks are dynamic in nature i.e., they are changing (in size) rapidly. For example., Facebook has around 2.5 billion users and this number is increasing day by day. This large graph makes matrix factorization methods obsolete. Thus, scalability is an import issue for future research in developing graph embedding methods. 
    \item \textbf{Bottleneck in Optimization:} Some embedding methods support multi-core implementations in shared memory architecture and employ asynchronous SGD to optimize the objective function. It makes the output embedding non-deterministic which is not expected. It also incurs the \emph{false sharing} problem, which is an unexpected phenomenon in shared memory programming due to poor optimization~\cite{bolosky1993false}. Thus, efforts can be made for new embedding methods so that the optimization process becomes deterministic and free of unexpected bottlenecks~\cite{rahman2021fusedmm}. 
    
    \item \textbf{Visualization:} The current methods embed graphs in vector space. Thus, they need another dimensionality reduction tool (e.g., t-SNE or UMAP) to be embedded in euclidean space for visualization. We have seen that the embedding of shallow networks or matrix factorization based methods can be used for several prediction tasks. A future direction can be to explore whether we can generate an embedding which can be directly used for visualization. Another interesting idea for dynamic graph visualization might be to use graph convolutional neural networks. The current visualization techniques mostly visualize static graph. If new vertices or edges arrive in the network, then we need to run the visualization method again. So, if we can learn weights for a graph using an inductive approach, as of graph convolutional neural networks, we can easily generate coordinates for new nodes which will facilitate visualization of dynamic graphs.
    
    \item \textbf{Interpretability of Representation:} Considering the irregular connections and structures of graphs, it is often found hard to generalize a model so that it can characterize different types of graphs well. Most of the methods assume that neighboring vertices will have a similar representation. This strategy works well for link prediction or even multi-class classification problem. If we go beyond that e.g., multi-label classification, random-walk or graph convolutional networks based methods become viable approaches. However, there are a few works on the interpretability aspects of graph embedding. In this direction, there is scope to explore more on the applicability of different complex network motifs and their theoretical aspects.
\end{itemize}

In addition to the above challenges, the selection of effective negative samples from a graph can be a future work. We have seen that false negative in randomly selected negative sampling approach plays an important role in optimization. Thus, special care must be taken while selecting negative samples from a graph. An illustrative theoretical work can give new insight about effective negative sampling in graph embedding domain.

\section{Conclusions}
\label{sec:conclusions}
In this paper, we have formally discuss the graph representation learning problem and several existing methodologies to solve it effectively. We have shown a new taxonomy of existing methods that are basically deduced from its evolution over time. We have also discussed some benchmark datasets that are commonly used to compare experimental results. We have conducted an extensive set of experiments to show comparative results of different methods for different prediction tasks such as link prediction, node classification, graph reconstruction, and clustering. Our empirical results show that some embedding methods can achieve better accuracy in prediction tasks but they may have high runtime and memory cost whereas some methods can perform reasonably well with moderate runtime and memory. Thus this paper provides a guideline to choose an embedding method focusing on a trade-off between better performance and computing resources. We have analyzed the sensitivity on hyper-parameters, e.g., a higher value of embedding dimension and the number of negative samples may not be important for making effective predictions. Finally, we have provided a list of challenges that can be considered as future directions. We hope that this paper shows a new insight into the evolution of graph embedding methods and their comparative analysis which will help future researchers to incorporate some challenging issues in the field.

\balance
\bibliographystyle{ACM-Reference-Format}
\bibliography{main}


\end{document}


\setcounter{page}{1}
\title{A Comprehensive Analytical Survey on Unsupervised and Semi-Supervised Graph Representation Learning Methods\\ (Supplementary File)}
\author{Md. Khaledur Rahman, and Ariful Azad}
\maketitle

\begin{table}[!htb]
\centering
\caption{Parameter settings for various graph embedding methods.}
\begin{tabular}{|c|p{8.5cm}|}
\hline
\textbf{Method} & \textbf{Values of parameters} \\ \hline
RolX & bin size = $5$, and others are default values. \\ \hline
HOPE & all are default values. \\ \hline
DeepWalk & walk length = $80$, num. of walks per vertex = $10$ \\ \hline
LINE & order = $2$, negative samples = $5$, number of samples = $10000$, rho = $0.025$, and others are default values \\ \hline
struc2vec & num. of walks = 20, walk length = 80, window size = 5, num. of layers = 6 and use all optimization options \\ \hline
VERSE & num. of negative samples = $5$ \\ \hline
HARP & \emph{line} model and window size = $2$ \\ \hline
Force2Vec & $t$-distribution for classification/clustering, and sigmoid version for link-prediction, learning rate=0.02 \\ \hline
GNNs & unless otherwise mentioned, we mostly used default values for all parameters. \\ \hline
\end{tabular}
\label{tab:parameters}
\end{table}